\documentclass[10pt,twocolumn,letterpaper]{article}
\usepackage[pagenumbers]{cvpr} % To force page numbers, e.g. for an arXiv version

\usepackage{cuted}
\usepackage{amsthm,mathtools}
\usepackage{multirow,array}
\usepackage{placeins}
\usepackage{algorithm,algpseudocode}
\usepackage{tikz}
\usetikzlibrary{arrows.meta}
\usepackage{xr-hyper}
\makeatletter\newcommand{\sref}[1]{\@ifundefined{sref@#1}{\textbf{??}}{\csname sref@#1\endcsname}}\makeatother
\expandafter\def\csname sref@supp:notation\endcsname{A}
\expandafter\def\csname sref@supp:proofs\endcsname{B}
\expandafter\def\csname sref@supp:densities\endcsname{B.5}
\expandafter\def\csname sref@supp:circle\endcsname{B.6}
\expandafter\def\csname sref@supp:method\endcsname{C}
\expandafter\def\csname sref@supp:estimation\endcsname{C.1}
\expandafter\def\csname sref@supp:hmm\endcsname{C.2}
\expandafter\def\csname sref@supp:impl\endcsname{C.3}
\expandafter\def\csname sref@supp:training\endcsname{C.4}
\expandafter\def\csname sref@supp:frameworks-impl\endcsname{C.5}
\expandafter\def\csname sref@supp:cost\endcsname{C.6}
\expandafter\def\csname sref@supp:eval\endcsname{D}
\expandafter\def\csname sref@supp:protocol\endcsname{D.1}
\expandafter\def\csname sref@supp:ourevaluation\endcsname{D.2}
\expandafter\def\csname sref@supp:evaldetails\endcsname{D.3}
\expandafter\def\csname sref@supp:kernels\endcsname{D.4}
\expandafter\def\csname sref@supp:data\endcsname{E}
\expandafter\def\csname sref@supp:results\endcsname{F}
\expandafter\def\csname sref@supp:res-q1\endcsname{F.1}
\expandafter\def\csname sref@supp:res-q2\endcsname{F.2}
\expandafter\def\csname sref@supp:frameworks\endcsname{F.3}
\expandafter\def\csname sref@supp:res-q4\endcsname{F.4}
\expandafter\def\csname sref@supp:res-q5\endcsname{F.5}

\usepackage[pagebackref,breaklinks,colorlinks,allcolors=cvprblue]{hyperref}

\newcommand{\M}{\mathcal{M}}
\newcommand{\Exp}{\operatorname{Exp}}
\newcommand{\Log}{\operatorname{Log}}
\newcommand{\vol}{\operatorname{vol}}
\newcommand{\Lip}{\operatorname{Lip}}
\newcommand{\E}{\mathbb{E}}
\newcommand{\R}{\mathbb{R}}
\newcommand{\SOthree}{\mathrm{SO}(3)}
\newcommand{\best}[1]{{\boldmath\textbf{#1}}}
\newcommand{\pmcell}[2]{\begin{tabular}[c]{@{}c@{}}$#1$\\[-1pt]$\pm#2$\end{tabular}}
\newtheorem{proposition}{Proposition}
\newtheorem{assumption}{Assumption}

\newcommand{\mref}[1]{\cref{#1}}

\definecolor{cvprblue}{rgb}{0.21,0.49,0.74}
\usepackage[pagebackref,breaklinks,colorlinks,allcolors=cvprblue]{hyperref}

\def\confName{CVPR}
\def\confYear{2027}

\title{Start Where the Data Are:\\Statistical Priors for One-Step Generation on Manifolds}

\author{
Yuyao Wang$^{1}$ \qquad Haoliang Sun$^{2\dagger}$ \qquad Xiankai Lu$^{2}$  \qquad
Xinxin Zhang$^{2}$ \qquad Yongshun Gong$^{2\dagger}$ \\[0.5em]
$^{1}$Department of Mathematics and Statistics, Boston University \\
$^{2}$School of Software, Shandong University \\
{\tt\small yuyaow@bu.edu, \{haolsun, ysgong\}@sdu.edu.cn}
}

\begin{document}
\maketitle
{\let\thefootnote\relax\footnotetext{$^{\dagger}$Corresponding authors.}}

\begin{abstract}

Fast generation on manifolds is important for scientific applications such as protein design, yet existing one-step generators often produce spurious samples in low-density gaps between data modes. Our analysis highlights the mismatch between uniform noise and concentrated data, motivating a statistical prior that captures the data’s coarse structure. We first estimate this prior by fitting mixtures of tangent-space Gaussians through geometry-aware moment matching. However, independent noise–data pairing can still connect different modes and weaken the benefit of the fitted prior. We therefore introduce posterior coupling, which samples a latent component conditioned on each data point and draws its paired noise from that component. To extend this approach to long torsion-angle chains, where full-covariance mixtures become costly, we use a hidden Markov prior to model sequential dependence with shared parameters. The resulting framework plugs into existing one-step training pipelines while retaining the generator architecture, training objective, and single network evaluation at inference. Experiments on spheres, tori, and SO(3) demonstrate substantial improvements across multiple generators, reducing distributional error by up to 125× and the protein Ramachandran outlier rate from 12.1\% to 0.55\%. Code is available at \url{https://github.com/olivia3395/statprior-onestep}.

\iffalse
Scientific data on manifolds, such as protein dihedral angles and earthquake locations, occupy a small part of the
manifold, and applications such as protein design need many samples from them. One-step generators offer fast sampling
on manifolds, but often produce samples outside the regions supported by real data, such as sterically forbidden protein
conformations. We trace this failure to the starting point: existing methods start from uniform noise over the manifold,
forcing a single map to concentrate probability mass into the small regions occupied by data. We show that this mismatch
is intrinsic: an exact, invertible map from prior to data must contract log-volume, on average, by the entropy gap
between them, and uniform noise imposes the largest contraction of any prior. Our solution is to start closer to the
data, with a lightweight prior fitted to the training data: a mixture of tangent-space Gaussians or, for chains of
torsion angles, a hidden Markov model. A posterior coupling preserves this prior while keeping transport paths short,
without changing the generator architecture, training objective or sampler. Across three one-step generators and nine
real datasets, our approach reduces one-step error in all 45 generator--dataset pairs, by up to 125$\times$. On
proteins, Ramachandran outliers drop from 12.1\% to 0.55\%, usable backbone fragments rise from 34\% to 93\%, and
rotamer outliers fall from 43\% to 3.9\%.
\fi

\end{abstract}

\section{Introduction}
\label{sec:intro}

\begin{figure}
\centering
\includegraphics[width=\linewidth]{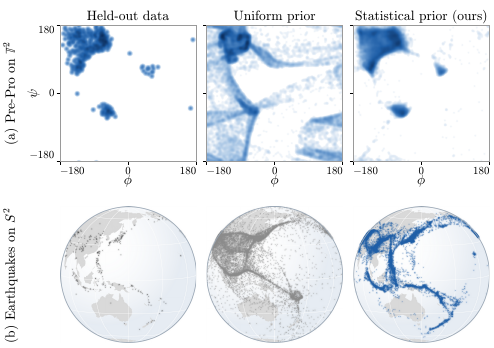}
\captionsetup{hypcap=false}
\caption{\textbf{Reducing spurious samples in low-density regions.} Held-out data (left), samples from a one-step
baseline using a uniform prior and independent pairing (middle), and samples from our method using a fitted
statistical prior and posterior coupling (right). Both generators use the same architecture and loss formulation. The
baseline places excess probability mass between conformational modes and away from the observed seismic belts, while
our method reduces these artifacts.}
\label{fig:teaser}
\captionsetup{hypcap=true}
\vspace{-1em}
\end{figure}

% \begin{figure}
% \centering

% % 上方：(a) Pre-Pro
% \begin{minipage}{\linewidth}
% \centering
% \includegraphics[
%     width=\linewidth,
%     trim=0 467.6 0 5,
%     clip
% ]{figs/fig_rama.pdf}\\[-1pt]
% \includegraphics[
%     width=\linewidth,
%     trim=0 4 0 338,
%     clip
% ]{figs/fig_rama.pdf}\\[1pt]
% {\small (a) Pre-Pro backbone dihedral angles on $\mathbb{T}^2$.}
% \end{minipage}

% \par\vspace{6pt}

% % 下方：(b) Earthquake
% \begin{minipage}{\linewidth}
% \centering
% \includegraphics[
%     width=\linewidth,
%     trim=0 0 0 163,
%     clip
% ]{figs/fig_globe.pdf}\\[1pt]
% {\small (b) Earthquake locations on $\mathbb{S}^2$.}
% \end{minipage}

% \par\vspace{2pt}

% \captionsetup{hypcap=false}
% \caption{
% \textbf{Reducing spurious samples in low-density regions.}
% Held-out data (left), samples from a one-step baseline using a uniform prior
% and independent pairing (middle), and samples from our method using a fitted
% statistical prior and posterior coupling (right).
% Both generators use the same architecture and loss formulation.
% The baseline places excess probability mass between conformational modes
% and away from the observed seismic belts, while our method reduces these artifacts.
% }
% \label{fig:teaser}
% \captionsetup{hypcap=true}
% \end{figure}

Fast generation of plausible samples supports scientific applications such as protein design \citep{watson2023novo}. Many scientific variables lie on manifolds, making their non-Euclidean geometry a challenge for generative modeling.  

Generative modeling on manifolds has advanced through Riemannian diffusion \citep{de2022riemannian,huang2022riemannian,lou2023scaling}, flow matching \citep{chen2024flow}, and normalizing flows \citep{mathieu2020riemannian,rezende2020normalizing,rozen2021moser,ben2022matching}. While diffusion and flow-matching methods typically require iterative sampling, recent generalized flow maps \citep{davis2026generalised} and Riemannian MeanFlows \citep{zhong2026riemannian,woo2026riemannian} enable generation with a single network evaluation. However, with uniform initialization, the one-step generators studied here often place excess probability mass in low-density gaps between data modes. In \cref{fig:teaser}, this appears as protein angles in sterically forbidden regions and earthquake samples dispersed away from the observed seismic belts.  

   We analyze this behavior through the mismatch between uniform noise and concentrated data. For exact smooth invertible transports on compact manifolds, the entropy gap determines the average log-volume change, with a uniform prior requiring the greatest average contraction for a fixed target (\cref{prop:jacobian}). Controlled experiments further show that one-step errors from uniform noise grow as the data modes become more concentrated.  

   To reduce this mismatch, our key idea is to first estimate a statistical prior on the manifold that captures the data's coarse multimodal structure, and then train a one-step generator to refine this approximation. Specifically, we represent the prior as a mixture of tangent-space Gaussians and fit it using posterior responsibilities and geometry-aware moment matching. The components are restricted to injectivity domains and mapped onto the manifold by the exponential map, with volume corrections ensuring valid densities and tractable component posteriors. However, independent pairing can still associate noise and data from different modes. We therefore sample a component from its posterior given each observation and draw the paired noise from that component. This posterior coupling encourages local transport and, for mixture priors at weight-update fixed points, preserves the sampling prior as the training-noise marginal. For long torsion-angle chains, where full-covariance joint mixtures become costly, we use a hidden Markov prior with shared transition and emission parameters to capture sequential dependence \citep{hamelryck2006sampling,boomsma2008generative}. The chain coupling samples a latent state sequence conditioned on the observed chain, followed by noise from the corresponding emissions. The resulting framework plugs into existing one-step training pipelines while retaining the generator architecture, loss formulation, and single network evaluation at inference. 

   Our contributions are: 
   \begin{itemize}[nosep,leftmargin=*] 
   \item We characterize the average log-volume change required by exact invertible transports and investigate prior--data mismatch through controlled experiments on one-step generation.  

   \item We develop geometry-aware statistical priors and posterior coupling, with a hidden Markov extension for torsion-angle chains, in a framework compatible with multiple one-step generators.  

   \item We demonstrate improvements across multiple generators on spheres, tori, and $\SOthree$, reducing distributional error by up to 125$\times$ and the Ramachandran outlier rate of generated protein residues from 12.1\% to 0.55\%. 
   \end{itemize}

\section{Related Work}
\label{sec:related}

\noindent\textbf{One-step generation on manifolds.} Flow maps learn the map from noise to data directly
\citep{boffi2024flow,frans2025one,geng2026mean,geng2026improved,zhang2026alphaflow}. On manifolds, generalized flow maps
\citep{davis2026generalised} and Riemannian MeanFlows \citep{zhong2026riemannian,woo2026riemannian} generate samples with a
single network evaluation. These methods change the training objective and all start from uniform noise. We change only the prior and the pairing, so our method applies to all of them. 

% We keep the
% objective and change the prior and the pairing, so our method applies to each of them.

\noindent\textbf{Priors.} Data-dependent and mixture priors have been proposed for Euclidean diffusion and flow models
\citep{lee2021priorgrad,jia2024structured,luo2026flow,rubbi2026shortest}, and \citet{salmona2022can} show that a push-forward
of a unimodal latent needs a large Lipschitz constant to fit multimodal data. We give an exact counterpart on compact
manifolds and build priors with tractable component posteriors on spheres, tori and $\SOthree$, following directional
statistics and TorusDBN \citep{mardia2009directional,hamelryck2006sampling,boomsma2008generative}.

\noindent\textbf{Couplings.} Minibatch optimal transport \citep{tong2023improving,pooladian2023multisample} and data-dependent
couplings \citep{albergo2023stochastic} pair noise with nearby data. Our posterior coupling needs no transport solver and
keeps the training noise distributed as the prior used for sampling. Further related work is discussed in
Supp.~\ref{supp:related}.

\section{Prior and Coupling Analysis}
\label{sec:why}

To understand the artifacts in \cref{fig:teaser}, we examine two design choices:
the initial distribution and the coupling used to form noise--data training pairs.

\noindent\textbf{Volume contraction under uniform initialization.}
Uniform noise distributes mass over the manifold, whereas the target density
is concentrated around its modes.
The following identity quantifies the average volume change required by an
exact transport.

\begin{proposition}[Average log-volume change]\label{prop:jacobian}
Let $p_0$ and $p_1$ be densities with finite entropies relative to the volume
measure of a compact Riemannian manifold $\mathcal M$.
Suppose that a $C^1$ diffeomorphism $T:\mathcal M\to\mathcal M$ satisfies
$T_\#p_0=p_1$, and let $J_T(x)>0$ denote its Riemannian volume Jacobian.
Then
\begin{equation}
  \E_{x\sim p_0}\bigl[\log J_T(x)\bigr]=H(p_1)-H(p_0).
  \label{eq:entropy}
\end{equation}
Consequently, for a fixed target $p_1$,
\[
  -\E_{x\sim p_0}\bigl[\log J_T(x)\bigr]
  = H(p_0)-H(p_1)
  \leq \mathrm{KL}(p_1\,\|\,\mathcal U),
\]
where $\mathcal U$ is the uniform density on $\mathcal M$, with equality when $p_0=\mathcal U$.
\end{proposition}
Thus, uniform initialization imposes the largest net average log-volume contraction for a fixed target. The proof and further analysis are provided in Supp.~\sref{supp:proofs}. We examine the relevance of this requirement to learned one-step maps empirically: controlled experiments show that errors from uniform initialization increase as the target modes become more concentrated (\cref{sec:q1}). These findings motivate estimating a statistical prior that captures the data's coarse modal structure before learning the remaining transport.

\noindent\textbf{Independent pairing under a fitted prior.} A fitted prior determines where noise is sampled, but does not determine how it is paired with data. Under independent pairing, $\pi_{\mathrm{ind}}=p_0\otimes p_1$, noise near one mode can be paired with observations from another. This distinction persists even if $p_0=p_1$: independent pairs can still cross modes, whereas the identity coupling pairs each point with itself. Matching the marginal distributions therefore does not by itself ensure local training pairs. Our ablation shows the practical importance of the coupling: on Pre-Pro, fitting the prior with independent pairing reduces the finest-scale error from 58 to 52 noise-floor standard deviations, while adding posterior coupling lowers it to 6.5 (\cref{tab:ablation}).  

Together, these observations motivate a statistical prior that captures the coarse data structure and a coupling that encourages local pairs while maintaining consistency between the training-noise marginal and the sampling prior. We develop these two components in \cref{sec:method}.

\begin{figure*}[t]
\vspace{-0.6em}
\centering
\includegraphics[width=\textwidth]{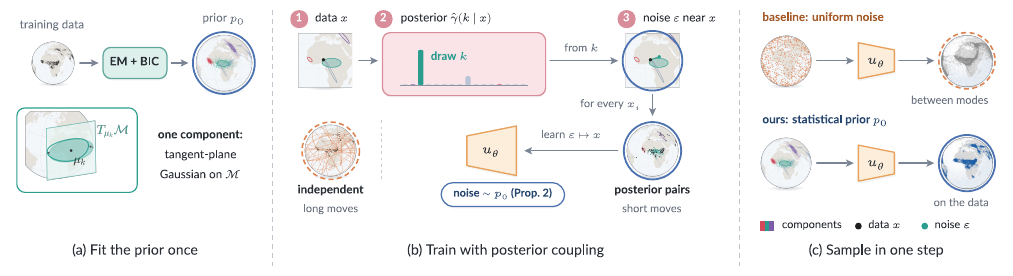}
\vspace{-2.3em}
\caption{\textbf{Overview.} (a) A mixture of tangent-space Gaussians is fitted once to the training data; colors mark
mixture components. (b) Each data point $x$ draws a component $k$ from its posterior, then noise from that component, so
the pairs are short, unlike independent pairing, and the noise still follows $p_0$; only $u_\theta$ is trained. (c) With
the same one-step network, uniform noise yields samples between modes, while the statistical prior yields samples on
the data.}
\label{fig:framework}
\vspace{-0.8em}
\end{figure*}

% \begin{figure*}[t]
% \vspace{-0.6em}
% \centering
% \includegraphics[width=\textwidth]{figs/fig_overview_v16.pdf}
% \caption{\textbf{Overview.} (a) A mixture of tangent-space Gaussians is fitted once to the training data; colors mark
% mixture components. (b) Each data point $x$ draws a component from its posterior, then noise from that component, so the
% pairs are short and the noise still follows $p_0$. (c) Sampling starts from $p_0$, with one network evaluation.}
% \label{fig:framework}
% \vspace{-0.4em}
% \end{figure*}

% \begin{figure*}[t]
% \vspace{-0.6em}
% \centering
% \includegraphics[width=\textwidth]{figs/fig_overview_v15.pdf}
% \caption{\textbf{Overview}, fitted to the Fire data. (a) A mixture of tangent-space Gaussians, fitted once; colors mark
% three of its $K=16$ components. (b) Each $x$ draws a component from its posterior, then noise from that component: the
% pairs shown move $11^\circ$ on average ($95^\circ$ if independent), and the noise still follows $p_0$. (c) Sampling
% starts from $p_0$, with one network evaluation.}
% \label{fig:framework}
% \vspace{-0.4em}
% \end{figure*}

\section{Methodology}
\label{sec:method}

We construct a statistical prior that captures the data's coarse modal
structure and use the fitted model's latent components to form
noise--data training pairs. The one-step generator then learns the
remaining transport from this initialization.
We introduce geometry-aware mixture priors and posterior coupling in
\cref{sec:prior,sec:coupling}, extend both constructions to torsion-angle
chains in \cref{sec:hmm}, and describe their integration with one-step
generators in \cref{sec:plugin}.

\subsection{Geometry-Aware Mixture Priors}
\label{sec:prior}

Given training observations $x_1,\ldots,x_n\in\M$, we fit a mixture
$\sum_{k=1}^{K}w_k q(x;\mu_k,\Sigma_k)$, where the positive weights
$w_k$ sum to one.
The components represent coarse modal structure and provide the
latent assignments used for posterior coupling.
Each component must define a valid manifold density, support sampling,
and permit evaluation of its posterior probability.

\noindent\textbf{Manifold-valued components.}
We construct each component by mapping a truncated tangent-space
Gaussian through the exponential map.
For $\mu\in\M$, an orthonormal basis $B_\mu$ of $T_\mu\M$, and
$\Sigma\succ0$, draw $v\sim\mathcal N(0,\Sigma)$ conditioned on
$v\in\mathcal D$, where $\mathcal D$ is an injectivity domain, and set
$x=\Exp_\mu(B_\mu v)$.
Away from the cut locus, the density relative to Riemannian volume is
\begin{equation}
  q(x;\mu,\Sigma)
  =\frac{\mathcal N(v;0,\Sigma)}{Z(\Sigma)\,j(v)},
  \qquad
  v=B_\mu^\top\Log_\mu x,
  \label{eq:density}
\end{equation}
where $Z(\Sigma)=\Pr_{v\sim\mathcal N(0,\Sigma)}(v\in\mathcal D)$
and $j(v)$ is the volume Jacobian of the exponential map.
On the unit sphere $\mathbb S^2$ and on $\SOthree$ with the
rotation-angle metric, $\mathcal D$ is the ball of radius $\pi$;
on the flat torus $\mathbb T^d$, it is $(-\pi,\pi]^d$.
Writing $r=\|v\|$, the respective Jacobians are $\sin r/r$,
$2(1-\cos r)/r^2$, and $1$, with their values at zero defined by
continuity.
For density evaluation, we approximate $Z(\Sigma)$ by Monte Carlo.
Component sampling uses rejection in tangent coordinates and does not
require evaluating this normalizing constant.

\noindent\textbf{Geometry-aware fitting.}
We combine EM responsibility updates \citep{dempster1977maximum}
with approximate geometric moment updates.
Given the current parameters, one iteration computes
\begin{equation}
\begin{gathered}
  \gamma_{ik}
  =\frac{w_k q(x_i;\mu_k,\Sigma_k)}
  {\sum_{\ell=1}^{K}w_\ell q(x_i;\mu_\ell,\Sigma_\ell)},
  \qquad
  w_k\leftarrow\frac1n\sum_{i=1}^{n}\gamma_{ik},
  \\[2pt]
  \mu_k\leftarrow
  \Exp_{\mu_k}\!\left(
    \frac{\sum_i\gamma_{ik}\Log_{\mu_k}x_i}
         {\sum_i\gamma_{ik}}
  \right),
  \\[2pt]
  \Sigma_k\leftarrow
  \frac{\sum_i\gamma_{ik}v_{ik}v_{ik}^{\top}}
       {\sum_i\gamma_{ik}},
  \qquad
  v_{ik}=B_{\mu_k}^{\top}\Log_{\mu_k}x_i .
\end{gathered}
\label{eq:em}
\end{equation}
With responsibilities held fixed, the mean update is a Riemannian
gradient step for the weighted Fr\'echet objective.
We repeat this step three times before computing the tangent coordinates
and covariance update at the resulting mean.
The moment updates approximate the component parameters rather than
solve an exact maximum-likelihood M-step for the truncated model.
We select $K$ using BIC \citep{schwarz1978estimating};
fitting details are provided in Supp.~\sref{supp:estimation}.

\noindent\textbf{Sampling prior.}
With fitted parameters $(\hat w_k,\hat\mu_k,\hat\Sigma_k)$, we define the sampling prior
\begin{equation}
  p_0(x)
  =\sum_{k=1}^{K}\hat w_k
  q(x;\hat\mu_k,s\hat\Sigma_k),
  \label{eq:prior}
\end{equation}
where $s\geq1$ scales the covariance parameters of the tangent-space
Gaussians while retaining their centers and mixture weights.
We use $s=2$ to broaden the components around the fitted modes.
On the manifolds considered, each component is positive almost
everywhere, so the prior imposes no hard support restriction.

\subsection{Posterior Coupling}
\label{sec:coupling}

We use the mixture's latent components to encourage local training pairs.
During training, an observation $x$ is sampled uniformly from
$\{x_i\}_{i=1}^{n}$.
We then sample $k\sim\hat\gamma(\cdot\mid x)$ and draw
$\varepsilon\sim q(\cdot;\hat\mu_k,s\hat\Sigma_k)$, where the
component posterior and conditional noise density are
\begin{equation}
\begin{aligned}
  \hat\gamma(k\mid x)
  &=
  \frac{\hat w_k q(x;\hat\mu_k,\hat\Sigma_k)}
  {\sum_{\ell=1}^{K}\hat w_\ell
    q(x;\hat\mu_\ell,\hat\Sigma_\ell)},
  \\[2pt]
  \rho(\varepsilon\mid x)
  &=
  \sum_{k=1}^{K}\hat\gamma(k\mid x)
  q(\varepsilon;\hat\mu_k,s\hat\Sigma_k).
\end{aligned}
\label{eq:coupling}
\end{equation}
Together with uniform sampling from the training set, this conditional
density specifies the training coupling.
The posterior uses the fitted covariances $\hat\Sigma_k$, while the
paired noise uses the scaled covariances $s\hat\Sigma_k$.
The resulting noise marginal admits the following characterization.

\begin{proposition}[Training-noise marginal]
\label{prop:marginal}
Let $\bar w_k=\frac1n\sum_{i=1}^{n}\hat\gamma(k\mid x_i)$.
Under the sampling procedure in \cref{eq:coupling}, the training-noise
marginal is
\[
  \bar p_0(\varepsilon)
  =\sum_{k=1}^{K}\bar w_k
  q(\varepsilon;\hat\mu_k,s\hat\Sigma_k),
\]
and satisfies
\[
  \mathrm{TV}(\bar p_0,p_0)
  \leq\frac12\sum_{k=1}^{K}|\bar w_k-\hat w_k|.
\]
In particular, $\bar p_0=p_0$ whenever $\bar w_k=\hat w_k$ for all
$k$, including at a fixed point of the mixture-weight update in
\cref{eq:em}.
\end{proposition}

The marginal guarantee depends on consistency between the averaged
posterior assignments and the fitted mixture weights.
Assigning each observation only to its most likely component can
change these effective weights.
Sharing a component encourages local pairing without imposing a
distance bound on individual pairs.
Optional within-component Hungarian matching \citep{kuhn1955hungarian}
can further reduce the total matching cost by permuting the sampled
noise; it retains the noise multiset and hence its aggregate marginal
while changing the joint pairing (Supp.~\sref{supp:impl}).

\subsection{Extension to Torsion-Angle Chains}
\label{sec:hmm}

For a chain of $L$ residues with angles
$\theta=(\theta_1,\ldots,\theta_L)\in(\mathbb T^2)^L$,
a full-covariance mixture requires $O(L^2)$ covariance parameters
per component.
We use a hidden Markov model \citep{rabiner1989tutorial},
following TorusDBN \citep{hamelryck2006sampling,boomsma2008generative},
to model sequential dependence with shared parameters.
A sequence of latent states $z=(z_1,\ldots,z_L)$ replaces the single
mixture component.

\begin{figure}[t]
\centering
\begin{tikzpicture}[>={Latex[length=1.6mm]},
  st/.style={circle,draw,minimum size=7.2mm,inner sep=0pt,font=\small},
  ob/.style={circle,draw,fill=black!12,minimum size=7.2mm,inner sep=0pt,font=\small},
  nz/.style={circle,draw=cvprblue,thick,text=cvprblue,minimum size=7.2mm,inner sep=0pt,font=\small}]
  \node[st] (z1) at (0,0) {$z_1$};
  \node[st] (z2) at (2.0,0) {$z_2$};
  \node (zd) at (3.7,0) {$\cdots$};
  \node[st] (zL) at (5.4,0) {$z_L$};
  \foreach \n/\x/\l in {1/0/1, 2/2.0/2, L/5.4/L} {
    \node[ob] (t\n) at (\x-0.5,-1.35) {$\theta_{\l}$};
    \node[nz] (e\n) at (\x+0.5,-1.35) {$\varepsilon_{\l}$};
    \draw[->] (z\n) -- (t\n);
    \draw[->,cvprblue,dashed] (z\n) -- (e\n);
  }
  \draw[->] (z1) -- node[above,font=\scriptsize]{$A$} (z2);
  \draw[->] (z2) -- (zd);
  \draw[->] (zd) -- (zL);
  \node[font=\scriptsize,anchor=east] at (-0.55,0) {$\nu$};
  \draw[->] (-0.55,0) -- (z1);
\end{tikzpicture}
% \caption{\textbf{Hidden Markov prior for a chain of $L$ residues.} Latent states $z_j$ follow a Markov chain with initial
% distribution $\nu$ and transition matrix $A$. Each state emits the angles $\theta_j$ of its residue (shaded) with the fitted covariance and, in the coupling, the noise $\varepsilon_j$ (blue) with the covariance scaled by $s$.}
\caption{\textbf{Hidden Markov prior for a chain of $L$ residues.} Latent states $z_j$ form a Markov chain ($\nu$, $A$);
each emits its residue's angles $\theta_j$ (shaded) and, in the coupling, the noise $\varepsilon_j$ (blue).}
\vspace{-1.4em}
\label{fig:hmm}
\end{figure}
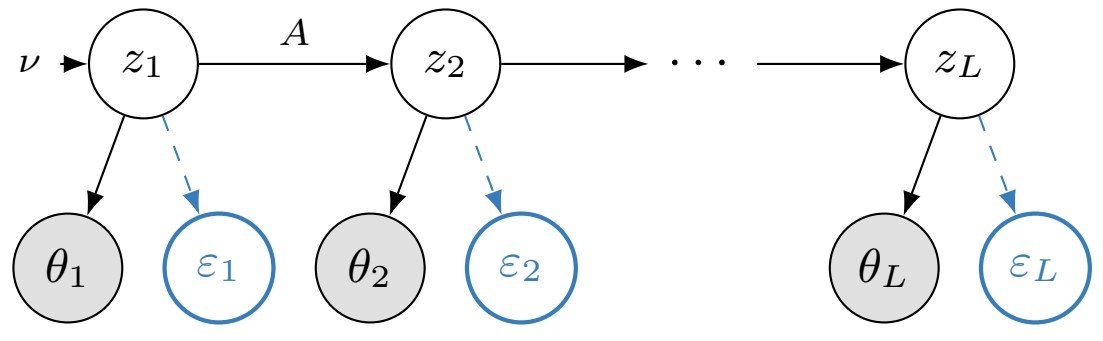

\noindent\textbf{Structured prior.}
%Each residue has a state $z_j\in\{1,\ldots,S\}$, and the joint model is
Each residue has a state $z_j\in\{1,\ldots,S\}$, and the joint
distribution factorizes along the chain (\cref{fig:hmm}) is
\begin{equation}
  p(z,\theta)
  =
  \nu_{z_1}
  \prod_{j=2}^{L}A_{z_{j-1}z_j}
  \prod_{j=1}^{L}q(\theta_j;\mu_{z_j},\Sigma_{z_j}),
  \label{eq:hmm}
\end{equation}
where $\nu$ is the initial state distribution, $A$ is the
$S\times S$ transition matrix, and each emission is a component
from \cref{eq:density} on $\mathbb T^2$.
The transition matrix and emission parameters are shared across
positions, with a common initial distribution across chains.
For fixed $S$, the parameter count is independent of $L$.
We fit the model using Baum--Welch-style iterations with
forward--backward inference and select $S$ by BIC.

After fitting, the sampling prior $p_0$ is obtained by marginalizing
the states in \cref{eq:hmm} and scaling each fitted emission covariance
by $s$.
We sample from this prior ancestrally, using the component sampler
for each emission, and evaluate its density with the forward algorithm.

\noindent\textbf{Posterior path coupling.}
For an observed chain $\theta$, we draw a complete state path
$z\sim\hat p(\cdot\mid\theta)$ using forward filtering and backward
sampling under the fitted model with unscaled emission covariances.
Conditional on this path, we independently draw
$\varepsilon_j\sim q(\cdot;\hat\mu_{z_j},s\hat\Sigma_{z_j})$
at each position.
Sampling the path jointly coordinates the noise across the chain
through the latent-state dependence.
For a dense transition matrix, both likelihood evaluation and
posterior-path sampling take $O(LS^2)$ time
(Supp.~\sref{supp:hmm}).

For training chains $\theta^{(1)},\ldots,\theta^{(n)}$ sampled
uniformly, the resulting noise marginal is
\begin{equation}
  \bar p_0(\varepsilon)
  =
  \frac1n\sum_{i=1}^{n}\sum_z
  \hat p(z\mid\theta^{(i)})
  \times
  \prod_{j=1}^{L} q(\varepsilon_j;\hat\mu_{z_j},s\hat\Sigma_{z_j}).
\end{equation}
It equals $p_0$ if the averaged posterior probability of every path
equals its ancestral probability
$\hat\nu_{z_1}\prod_{j=2}^{L}\hat A_{z_{j-1}z_j}$.
A Baum--Welch fixed point does not generally enforce this condition,
so the exact marginal guarantee in \cref{prop:marginal} does not
automatically extend to the HMM.

\subsection{Integration with One-Step Generators}
\label{sec:plugin}

The construction applies to one-step generators whose training
procedures support changes to both the source distribution and the
noise--data coupling.
We fit the statistical model once on the training split and hold its
parameters fixed during generator training.
Training uses posterior-coupled pairs while retaining the base
architecture and loss formulation.
At inference, noise is drawn from $p_0$ and passed to the base
one-step sampler, using a single network evaluation.

\Cref{alg:main} summarizes the mixture-prior procedure.
For torsion-angle chains, posterior state-path sampling replaces
component sampling during training, and inference uses ancestral
sampling from the HMM prior.

\begin{algorithm}[t]
\caption{One-step generation with a mixture prior.}
\label{alg:main}
\small
\begin{algorithmic}[1]
\Statex \textbf{Prior fitting}
\State Fit $(\hat w_k,\hat\mu_k,\hat\Sigma_k)_{k=1}^{K}$ on the
training split using \cref{eq:em}; select $K$ by BIC
\State Define $p_0$ using \cref{eq:prior}
\Statex \textbf{Generator training}
\For{each minibatch $\{x_i\}$}
  \State Sample $k_i\sim\hat\gamma(\cdot\mid x_i)$ for each $x_i$
  \State Draw
  $\varepsilon_i\sim q(\cdot;\hat\mu_{k_i},s\hat\Sigma_{k_i})$
  \State Update the generator using its base loss on $(\varepsilon_i,x_i)$
\EndFor
\Statex \textbf{Sampling}
\State Draw $\varepsilon\sim p_0$
\State Apply the base one-step sampler using one network evaluation
\end{algorithmic}
\end{algorithm}

\section{Experiments}
\label{sec:experiments}

We evaluate on the sphere, the torus and $\SOthree$, with data from Earth science and structural biology and controlled
benchmarks (\cref{sec:setup}). \Cref{sec:q1} relates prior mismatch to generation error; \cref{sec:q3} compares
generators and sampling budgets; \cref{sec:q4} covers higher dimensions, protein chains and conditional generation;
\cref{sec:q5} ablates each component and tests sensitivity to the prior.

% We evaluate statistical priors and posterior coupling for one-step
% generation on the sphere, the torus, and $\SOthree$, using data from
% Earth science and structural biology together with controlled benchmarks
% (\cref{sec:setup}).
% Controlled experiments examine the relationship between prior mismatch
% and generation error (\cref{sec:q1}), followed by comparisons across
% generators and sampling budgets (\cref{sec:q3}).
% We further evaluate higher-dimensional distributions, protein chains,
% and conditional generation (\cref{sec:q4}), and use ablations and
% sensitivity analyses to assess component contributions and the effects
% of prior construction (\cref{sec:q5}).

\subsection{Experimental Settings}
\label{sec:setup}

% \noindent\textbf{Data.}
% We use 80/10/10 splits for training, validation, and testing
% (Supp.~\sref{supp:data}).
% AUTHOR CHECK: Specify the splitting unit for protein windows and
% fragments, including whether overlapping windows or fragments from
% the same protein can occur in different splits.

\noindent\textbf{Data.} All datasets are split 80/10/10 with a fixed seed (Supp.~\sref{supp:data}). Benchmarks follow
the splits of \citet{zhong2026riemannian} and \citet{davis2026generalised}; our protein windows and fragments are
non-overlapping, so no residue appears in two splits.

% \noindent\textbf{Data.} All datasets are split 80/10/10 into training, validation and test sets with a fixed seed
% (Supp.~\sref{supp:data}). Single residues and the Earth and $\SOthree$ benchmarks follow the splits of the released code
% of \citet{zhong2026riemannian} and \citet{davis2026generalised}; protein windows and fragments, which we introduce, are
% non-overlapping and split at random, so no residue appears in two splits.

\begin{itemize}[nosep,leftmargin=*]
\item Structural biology (\textbf{torus}): backbone angles
$(\phi,\psi)\in\mathbb T^2$ from the Top500 proteins
\citep{lovell2003structure}, grouped into four residue classes
(General, Glycine, Proline, Pre-Pro; 6910--138{,}208 residues);
side-chain angles $(\chi_1,\chi_2)\in\mathbb T^2$ from 14 residue types;
windows of 2, 3, and 5 residues ($\mathbb T^4$--$\mathbb T^{10}$);
backbone fragments of 10 and 20 residues
($\mathbb T^{20}$, $\mathbb T^{40}$); and RNA backbone torsion angles
on $\mathbb T^7$ \citep{murray2003rna}.
\item Earth science (\textbf{sphere}): volcanic eruptions, earthquakes, floods, and
wildfires on $S^2$ \citep{mathieu2020riemannian}
(827--12{,}809 events).
\item Controlled benchmarks ($\SOthree$): the four $\SOthree$ benchmarks of
\citet{zhong2026riemannian}; and synthetic distributions with known
densities, including von Mises--Fisher mixtures, rings, and arcs on
$S^2$, and bent Gaussian mixtures on $\mathbb T^d$ with $d\le50$.
\end{itemize}

\noindent\textbf{Generators and baselines.} We add the prior and the coupling to five one-step generators: the
Riemannian MeanFlow (RMF) of \citet{zhong2026riemannian}; the Eulerian, Lagrangian and semigroup formulations of the
Riemannian MeanFlow of \citet{woo2026riemannian} (RMF-E, RMF-L, RMF-S); and generalized flow maps (G-LSD)
\citep{davis2026generalised}, each with its original architecture, loss, sampler and training schedule. Baselines are
(i) the same generator with its uniform prior; (ii) the uniform prior with four sampling steps or five times longer
training; (iii) a kernel density estimate; and (iv) on $\SOthree$, the one-step results of RFM \citep{chen2024flow},
G-LSD, RMF and RMF-MT reported by \citet{zhong2026riemannian}.

\noindent\textbf{Metrics.} We report the unbiased MMD$^2$ \citep{gretton2012kernel} between generated and held-out
samples, divided by its noise floor, the standard deviation of the same statistic between disjoint sets of real data;
values $\le1$ lie within the noise floor. We use the RMF kernel of \citet{zhong2026riemannian} and finer-scale Laplace
kernels, which we prove characteristic \citep{sriperumbudur2011universality} (Supp.~\sref{supp:eval}). On synthetic data, we also report the log-likelihood lost relative to samples from the target.

\noindent\textbf{Protein geometry metrics.}
We evaluate torsion-angle outliers and a C$_\alpha$-based clash
criterion, motivated by standard structure-validation diagnostics
\citep{lovell2003structure,chen2010molprobity}.
A residue is a Ramachandran outlier if its $(\phi,\psi)$ angles lie
outside the 99.95\% density contour of its type; a side chain is a
rotamer outlier if its $(\chi_1,\chi_2)$ angles lie outside the
99.7\% contour \citep{shapovalov2011smoothed}.
A backbone fragment is counted as usable under our criteria if it
contains neither a Ramachandran outlier nor a C$_\alpha$ clash.
We define a clash as a pair of C$_\alpha$ atoms at least three residues
apart in the sequence and less than 3.0\,\AA{} apart in space.
For this evaluation, we sample eight-residue fragments for 5000 real
sequences and reconstruct their three-dimensional backbones.

\noindent\textbf{Implementation.}
The statistical model is fitted once per dataset using the training
split, with $K$ or the number of hidden Markov states selected by BIC
and covariance scale $s=2$.
Unless stated otherwise, experiments with the statistical prior also
use posterior coupling.
Main results do not use Hungarian matching.

\subsection{A Controlled Study of the Uniform Prior}
\label{sec:q1}

\begin{figure}[t]
\centering
\includegraphics[width=0.85\linewidth]{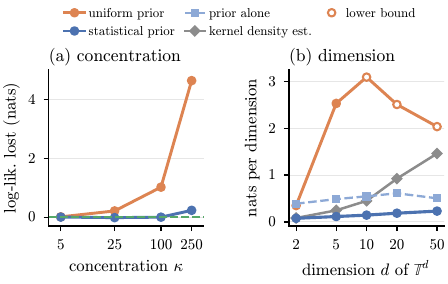}
% \caption{\textbf{Effects of concentration and dimension.}
% (a) Target log-density gap of generated samples as the concentration
% of von Mises--Fisher mixtures increases.
% (b) Gap per dimension on synthetic tori.
% Conditional generation and real torsion windows are evaluated in
% Supp.~\sref{supp:res-q4}.}
\caption{\textbf{Effects of concentration and dimension.} (a) Log-likelihood lost against the concentration of von
Mises--Fisher mixtures. (b) Loss per dimension on synthetic tori.}
\label{fig:trends}
\vspace{-1.4em}
\end{figure}

\begin{table}[t]
\centering\footnotesize\setlength{\tabcolsep}{3.2pt}
\caption{\textbf{How trained maps meet the demand} on representative datasets: required contraction (nats), folded prior
samples (\%) and displacement (rad). U, S: uniform, statistical prior. All nine datasets: Supp.~\sref{supp:res-q1}.}
\label{tab:geometry}
\begin{tabular*}{\linewidth}{@{\extracolsep{\fill}}l cc cc cc@{}}
\toprule
& \multicolumn{2}{c}{required} & \multicolumn{2}{c}{folded (\%)} & \multicolumn{2}{c}{displacement} \\
\cmidrule(lr){2-3}\cmidrule(lr){4-5}\cmidrule(lr){6-7}
Dataset & U & S & U & S & U & S \\
\midrule
Volcano & 4.25 & 0.04 & 33.6 & 28.9 & 0.66 & 0.05 \\
Fire & 3.10 & 0.46 & 22.3 & 8.2 & 0.76 & 0.03 \\
Proline & 3.35 & 0.41 & 44.9 & 0.3 & 1.76 & 0.09 \\
Pre-Pro & 2.34 & 0.35 & 37.8 & 1.2 & 1.53 & 0.10 \\
RNA ($\mathbb T^7$) & 16.84 & 1.63 & 51.0 & 13.0 & 3.92 & 0.30 \\
\bottomrule
\end{tabular*}
\end{table}

We test whether the contraction required by \cref{prop:jacobian} predicts the error of trained one-step generators and
how their maps deform space.

% We examine how one-step generation error and map deformation vary
% with prior--data mismatch.
% The entropy identity in \cref{prop:jacobian} quantifies the net average
% log-volume contraction required by an exact invertible transport.
% Here, controlled distributions and geometric diagnostics allow us to
% examine the empirical relationship between this quantity and the
% behavior of trained generators.

\noindent\textbf{Target concentration.}
We use mixtures of five von Mises--Fisher modes on $S^2$, varying only
their concentration $\kappa$ while keeping the remaining construction
fixed (\cref{fig:trends}a).
As $\kappa$ increases from 5 to 250, the target log-density gap of
uniform-prior samples increases from 0.01 to 4.6 nats.
The corresponding gap with our method remains below 0.23 nats.
Across the eight synthetic datasets, the uniform-prior gap is
positively associated with $\mathrm{KL}(p_1\,\|\,\mathcal U)$,
the entropy gap for uniform initialization
(rank correlation 0.90; Supp.~\sref{supp:res-q1}).

\noindent\textbf{Map geometry.}
The deformation statistics show a related trend
(\cref{tab:geometry}).
Using the fraction of sampled points with orientation-reversing
Jacobians as the folding diagnostic, uniform initialization gives
rates of 16--51\%, with higher rates associated with larger
contraction requirements (Spearman $\rho=0.85$).
The statistical prior reduces the estimated entropy gap by a factor
of 4.7 to 106.
With our method, the folding diagnostic falls to at most 1.2\% on the
protein datasets, and mean geodesic displacement is reduced by a
factor of 6 to 25 across the evaluated datasets.

\subsection{Improving Existing Generators}
\label{sec:q3}\label{sec:frameworks}\label{sec:q2}\label{sec:earth}

\begin{table*}[t]
\vspace{-1.2em}
\centering\small\setlength{\tabcolsep}{3.6pt}
\caption{\textbf{The statistical prior improves every one-step generator.} One-step error, uniform $\to$ statistical
prior, at the finest scale in noise-floor multiples ($\le1$: within the estimated noise floor). Prior alone: samples
drawn from the fitted prior without any network, compared with the uniform prior of RMF. RMF-E, RMF-L, RMF-S: the
Eulerian, Lagrangian and semigroup formulations of the Riemannian MeanFlow of~\cite{woo2026riemannian}. Bold: separated
by more than three standard deviations. Avg.\ improvement: geometric mean over the nine datasets of the ratio of the
uniform-prior error to the statistical-prior error, with $\le1$ counted as 1.}
% Each column has its own runs, so values compare within a column.
%\vspace{-.6em}
\label{tab:frameworks}
\begin{tabular*}{\textwidth}{@{\extracolsep{\fill}}l c cccc c@{}}
\toprule
 & Prior alone & \multicolumn{4}{c}{Riemannian MeanFlow} & Generalized flow map \\
\cmidrule(lr){3-6}\cmidrule(lr){7-7}
Dataset & (no network) & RMF~\cite{zhong2026riemannian} & RMF-E~\cite{woo2026riemannian} & RMF-L~\cite{woo2026riemannian} & RMF-S~\cite{woo2026riemannian} & G-LSD~\cite{davis2026generalised} \\
\midrule
Volcano & 1.2 & 4.6 $\to$ \textbf{$\le$1} & 4.8 $\to$ \textbf{$\le$1} & 4.1 $\to$ 1.1 & 4.6 $\to$ \textbf{$\le$1} & 1.3 $\to$ $\le$1 \\
Earthquake & 8.6 & 24 $\to$ \textbf{5.7} & 28.3 $\to$ \textbf{8.4} & 21.6 $\to$ \textbf{6.9} & 25.0 $\to$ \textbf{6.6} & 7.0 $\to$ 4.4 \\
Flood & 8.1 & 15 $\to$ \textbf{9.0} & 11.0 $\to$ \textbf{6.2} & 7.4 $\to$ 4.8 & 12.4 $\to$ \textbf{5.1} & 7.7 $\to$ 5.8 \\
Fire & 35 & 110 $\to$ \textbf{20} & 57.5 $\to$ \textbf{13.2} & 49.8 $\to$ \textbf{12.7} & 69.1 $\to$ \textbf{12.3} & 16.5 $\to$ \textbf{8.6} \\
\midrule
General & 149 & 442 $\to$ \textbf{13} & 289 $\to$ \textbf{46.0} & 434 $\to$ \textbf{5.7} & 403 $\to$ \textbf{3.2} & 49.1 $\to$ \textbf{3.5} \\
Glycine & 11 & 35 $\to$ \textbf{3.0} & 43.1 $\to$ \textbf{2.8} & 116 $\to$ \textbf{3.6} & 41.2 $\to$ \textbf{3.7} & 6.1 $\to$ \textbf{1.8} \\
Proline & 18 & 100 $\to$ \textbf{3.4} & 91.0 $\to$ \textbf{2.2} & 193 $\to$ \textbf{3.1} & 170 $\to$ \textbf{2.5} & 11.5 $\to$ \textbf{$\le$1} \\
Pre-Pro & 15 & 54 $\to$ \textbf{7.3} & 38.7 $\to$ \textbf{2.6} & 29.3 $\to$ \textbf{1.7} & 26.4 $\to$ \textbf{1.8} & 15.6 $\to$ \textbf{1.0} \\
RNA ($\mathbb T^7$) & 95 & 280 $\to$ \textbf{17} & 226 $\to$ \textbf{23.0} & 236 $\to$ \textbf{13.7} & 164 $\to$ \textbf{8.9} & 75.0 $\to$ \textbf{2.7} \\
\midrule
Avg.\; $\uparrow$ & 3.2$\times$ & 8.6$\times$ & 7.5$\times$ & 11.4$\times$ & 12.2$\times$ & 4.8$\times$ \\
\bottomrule
\end{tabular*}
%\vspace{-1em}
\end{table*}

\begin{table}[t]
\centering\footnotesize\setlength{\tabcolsep}{3pt}
\caption{\textbf{One step from the statistical prior against more steps from the uniform prior} (Riemannian MeanFlow of
\citet{zhong2026riemannian}, same network and training budget; MMD$^2\times10^3$ at the finest scale; lower is better). All eight datasets:
Supp.~\sref{supp:frameworks}.}
\label{tab:steps}
\begin{tabular*}{\linewidth}{@{\extracolsep{\fill}}lccc@{}}
\toprule
& \multicolumn{2}{c}{uniform prior} & statistical prior \\
\cmidrule(lr){2-3}\cmidrule(lr){4-4}
Dataset & 1 step & 4 steps & 1 step \\
\midrule
Fire & 5.37 & 5.19 & \best{0.98} \\
Earthquake & 2.37 & 1.69 & \best{0.57} \\
Glycine & 2.18 & 1.44 & \best{0.19} \\
Proline & 15.85 & 16.12 & \best{0.53} \\
RNA ($\mathbb T^7$) & 2.80 & 1.93 & \best{0.17} \\
\bottomrule
\end{tabular*}
\vspace{-1.5em}
\end{table}

We evaluate the statistical prior and posterior coupling across the
generators described in \cref{sec:setup}.
Comparisons with uniform-prior baselines assess generation quality
at the default training schedule and with additional sampling steps
or longer training.

\noindent\textbf{Performance across generators.}
Our method reduces the reported error in all 45 evaluated
generator--dataset pairs, with reductions exceeding three
noise-floor standard deviations in 40 comparisons
(\cref{tab:frameworks}).
The error reductions range from 3.4--125$\times$ on torsion-angle
data and 1.3--6$\times$ on Earth-science data.
Under the RMF kernel, the reported scores are within the empirical
noise-floor threshold on all evaluated protein and RNA datasets.

\noindent\textbf{Sampling steps and training budget.} On every dataset of \cref{tab:steps}, our one-step configuration
has 3 to 30 times lower error than the four-step uniform-prior baseline, and it has lower error in all 27 comparisons
with the formulations of \citet{woo2026riemannian} (Supp.~\sref{supp:frameworks}). On Proline, five times longer
training and four steps bring the uniform prior to 6.3, against 3.4 for our one-step configuration at the original
budget (Supp.~\sref{supp:res-q5}).

% \noindent\textbf{Sampling steps and training budget.}
% On the datasets reported in \cref{tab:steps}, our one-step configuration
% has 3 to 30 times lower error than the four-step uniform-prior
% baseline.
% It also outperforms the four-step uniform-prior baseline in all
% 27 comparisons using the formulations of \citet{woo2026riemannian}
% (Supp.~\sref{supp:frameworks}).
% On Proline, increasing uniform-prior training by a factor of five
% and using four sampling steps gives an error of 6.3, compared with
% 3.4 for our one-step configuration at the original training budget
% (Supp.~\sref{supp:res-q5}).
% AUTHOR CHECK: Verify the coverage of tab_steps_v2. Historical commented
% text reports 11 of 13 objective--dataset combinations and an alpha-Flow
% exception. The current active draft reports 3--30x gains on every
% dataset in this table; retain that claim only if the current table agrees.

\begin{figure*}[t]
\centering
\includegraphics[width=0.85\textwidth]{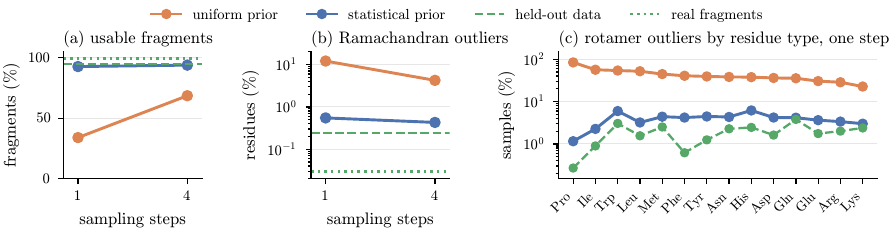}
%\vspace{-1.6em}
\caption{\textbf{Protein geometry diagnostics of one-step samples.}
(a, b) Usable-fragment and Ramachandran-outlier rates for backbone
fragments sampled for 5000 real sequences, as a function of the number
of sampling steps.
(c) Rotamer-outlier rates for 14 residue types.}
\label{fig:validation}
\vspace{-1.2em}
\end{figure*}

% \begin{figure*}[t]
% \centering
% \includegraphics[width=\textwidth]{figs/fig_geom_trends_v2.pdf}
% \vspace{-1.6em}
% \caption{\textbf{Protein geometry and effects of concentration and dimension.} (a, b) Usable-fragment and
% Ramachandran-outlier rates of backbone fragments sampled for 5000 real sequences, against the number of steps.
% (c) Rotamer-outlier rates for 14 residue types. (d) Log-likelihood lost against the concentration of von
% Mises--Fisher mixtures. (e) Loss per dimension on synthetic tori.}
% \label{fig:geomtrends}
% \vspace{-1.2em}
% \end{figure*}

\begin{figure}[t]
\centering
\includegraphics[width=0.89\linewidth]{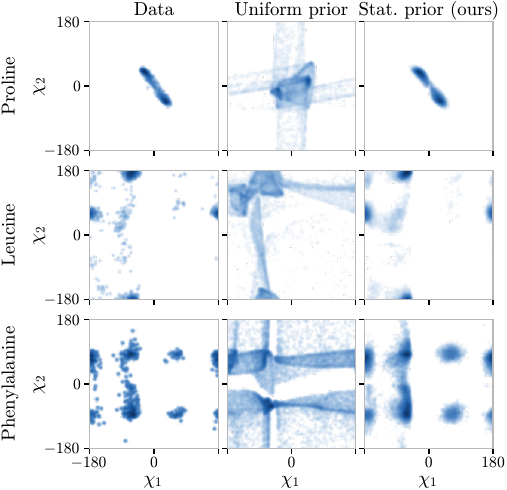}
% \caption{\textbf{Side-chain angles $(\chi_1,\chi_2)$} of proline,
% leucine, and phenylalanine.
% One-step samples initialized from the uniform prior form bands through
% regions with little observed density.
% Our samples concentrate more closely around the observed rotamer modes.}
%\vspace{-1.4em}
\caption{\textbf{Side-chain angles $(\chi_1,\chi_2)$} of proline, leucine and phenylalanine. From the uniform prior,
samples form bands through empty regions; ours stay on the rotamer modes.}
\label{fig:rotamers}
\vspace{-.3em}
\end{figure}

\noindent\textbf{Protein geometry diagnostics.} Usable fragments rise from 34\% with the uniform prior to 93\% with ours
(real data: 95\%), while four uniform-prior steps reach only 69\% (\cref{fig:validation}). Ramachandran outliers fall
from 12.1\% to 0.55\% (real: 0.24\%), and rotamer outliers from 43\% to 3.9\% (real: 1.9\%). From the uniform prior,
samples fill low-density regions of the Ramachandran plot (\cref{fig:teaser}a) and form bands between rotamer modes
(\cref{fig:rotamers}).

% \noindent\textbf{Protein geometry diagnostics.}
% The fraction of usable fragments under the criteria in
% \cref{sec:setup} increases from 34\% for the one-step uniform-prior
% baseline to 93\% with our method, compared with 95\% for real data
% and 69\% for the four-step uniform-prior baseline
% (\cref{fig:validation}).
% Ramachandran-outlier rates decrease from 12.1\% to 0.55\%
% (real data: 0.24\%), and rotamer-outlier rates decrease from 43\%
% to 3.9\% (real data: 1.9\%).
% These improvements are also visible in the angle distributions:
% uniform-prior samples populate low-density regions of the
% Ramachandran plot (\cref{fig:teaser}a) and form bands between
% observed rotamer modes (\cref{fig:rotamers}).

% \begin{figure}[t]
% \centering
% \includegraphics[width=\linewidth]{figs/fig_earth_density.pdf}
% \caption{\textbf{One-step samples on the four Earth datasets.}
% Uniform initialization places more mass away from the observed event
% concentrations, whereas our method more closely follows the spatial
% patterns in the catalogs.}
% \vspace{-1.4em}
% \label{fig:earthsamples}
% \end{figure}

\noindent\textbf{Earth samples stay where events occur.} On all four catalogs, samples from the statistical prior follow
the volcanic and seismic belts and the flood and fire regions, while samples from the uniform prior spread across the
open ocean (\cref{fig:teaser}b; Supp.~\sref{supp:res-q2}).

% \noindent\textbf{Spatial distributions of Earth-science samples.}
% Across the four catalogs, our samples more closely follow the
% observed volcanic and seismic belts and the regions of concentrated
% flood and fire activity.
% The uniform-prior baselines place more mass in regions with little
% support in the observed catalogs (\cref{fig:earthsamples}).

\begin{table}[t]
\centering\footnotesize\setlength{\tabcolsep}{2pt}
\caption{\textbf{One-step generation on $\SOthree$}: MMD against the baselines reported by \citet{zhong2026riemannian} (lower is
better; bold: lowest).}
\label{tab:so3}
\begin{tabular*}{\linewidth}{@{\extracolsep{\fill}}lcccc@{}}
\toprule
Method & Cone & Fisher & Line & Swiss \\
\midrule
RFM \citep{chen2024flow} & 0.311 & 0.154 & 0.095 & 0.346 \\
G-LSD \citep{davis2026generalised} & 0.044 & 0.073 & 0.037 & 0.032 \\
RMF \citep{zhong2026riemannian} & 0.057 & 0.076 & 0.038 & 0.080 \\
RMF-MT \citep{zhong2026riemannian} & 0.049 & 0.039 & 0.035 & 0.061 \\
\midrule
RMF + statistical prior (ours) & \best{0.012} & \best{0.025} & \best{0.025} & \best{0.017} \\
\bottomrule
\end{tabular*}
%\vspace{-1em}
\end{table}

\noindent\textbf{Rotation benchmarks.}
On all four $\SOthree$ benchmarks, our one-step configuration achieves
the lowest error among the compared methods.
Its error is lower by a factor of 1.4 to 3.7 than the best one-step
result reported by \citet{zhong2026riemannian}
(\cref{tab:so3}).

\subsection{Higher Dimensions and Structured Data}
\label{sec:q4}

\begin{table}[t]
%\vspace{-.5em}
\centering\footnotesize\setlength{\tabcolsep}{2pt}
% \caption{\textbf{Backbone fragments of 10 residues} ($\mathbb T^{20}$) rebuilt in 3D: clash rate (lower is better), helix
% and sheet fractions (closer to held-out is better), and $W_1$ distance between end-to-end distances (\AA; lower is
% better). Bold: closest to held-out fragments. Fragments of 20 residues are in Supp.~\sref{supp:res-q4}.}
\caption{\textbf{Backbone fragments of 10 residues} ($\mathbb T^{20}$) rebuilt in 3D: clash rate (lower is better), helix
and sheet fractions (closer to held-out is better), and $W_1$ distance between end-to-end distances (\AA; lower is
better). RMF rows are one-step samples of the generator trained with each prior. Bold: closest to held-out fragments.
Fragments of 20 residues are in Supp.~\sref{supp:res-q4}.}
%\vspace{-.5em}
\label{tab:highdim}
\begin{tabular*}{\linewidth}{@{\extracolsep{\fill}}lcccc@{}}
\toprule
 & clash & helix & sheet & $W_1$ \\
\midrule
Held-out fragments & 0.012 & 0.496 & 0.415 & -- \\
RMF + uniform prior & 0.164 & 0.417 & 0.225 & 2.24 \\
Kernel density estimate & 0.149 & 0.459 & 0.324 & 0.78 \\
RMF + mixture prior & 0.055 & 0.486 & 0.341 & \best{0.40} \\
RMF + hidden-Markov prior & \best{0.043} & \best{0.501} & \best{0.407} & 0.58 \\
\bottomrule
\end{tabular*}
\end{table}

We next test higher dimensions, longer protein fragments and conditional generation.
% We next evaluate higher-dimensional distributions, longer protein
% fragments, and generation conditioned on residue type.

\noindent\textbf{Higher-dimensional distributions.}
On synthetic tori from $d=2$ to $d=50$, where the fitted mixture is
misspecified, our one-step samples have a target log-density gap
of 0.08 to 0.23 nats per dimension, the lowest among the evaluated
methods at each dimension (\cref{fig:trends}b).
Kernel density estimation has gaps of up to 1.46 nats per dimension,
while the uniform-prior generator exhibits substantially larger
gaps from $d=5$ onward.
The method also improves performance on real protein windows.
For windows of 2 and 3 residues ($\mathbb T^4$, $\mathbb T^6$),
the reported errors are within the empirical noise-floor threshold.
For 5 residues ($\mathbb T^{10}$), the error is 9.2 noise-floor
standard deviations, compared with 305--540 for the uniform-prior
baselines (Supp.~\sref{supp:res-q4}).

\noindent\textbf{Protein fragment generation.} The hidden Markov prior has a parameter count independent of chain length
(\cref{sec:hmm}). On 10-residue fragments, it reaches a held-out log-likelihood of $-7.8$ per fragment, against $-17.5$
for a 32-component mixture with ten times more parameters. RMF trained with it has the lowest clash rate in
\cref{tab:highdim}, 4.3\% against 16.4\% for the uniform prior (real: 1.2\%), and helix and sheet fractions of 50.1\% and
40.7\% (real: 49.6\% and 41.5\%); 20-residue fragments agree (Supp.~\sref{supp:res-q4}). The prior captures local
structure but not global shape: on 20 residues, its end-to-end distances are the least accurate.

\noindent\textbf{Conditional generation.}
For residue-type conditioning, we fit one mixture per type and use
the condition to select the corresponding prior for a single
conditional generator.
The resulting model has RMF-kernel scores within the empirical
noise-floor threshold on all four residue types.
With uniform initialization, the shared conditional model has
higher error than separately trained models, with scores of 103
and 19 noise-floor standard deviations, respectively, on Proline
(Supp.~\sref{supp:res-q4}).
These comparisons support using condition-specific priors in the
evaluated conditional setting.

\subsection{Ablation and Sensitivity Analysis}
\label{sec:q5}\label{sec:analysis}

\begin{table}[t]
\centering\footnotesize\setlength{\tabcolsep}{2pt}
% \caption{\textbf{Prior and coupling} (noise-floor standard deviations; lower is better; $\le$1: within the estimated
% noise floor). Fitted prior: noise from the statistical prior, paired independently with the data; + coupling: paired by
% the posterior. Stress tests and cost: Supp.~\sref{supp:res-q5}.}
\caption{\textbf{The prior places the mass; the coupling recovers its shape} (noise-floor standard deviations; $\le$1:
within the noise floor). Fitted prior: paired independently with the data; + coupling: paired by the posterior.}
\label{tab:ablation}
\vspace{-.5em}
\begin{tabular*}{\linewidth}{@{\extracolsep{\fill}}l ccc ccc@{}}
\toprule
& \multicolumn{3}{c}{RMF kernel} & \multicolumn{3}{c}{finest scale} \\
\cmidrule(lr){2-4}\cmidrule(lr){5-7}
& uniform & fitted prior & + coupling & uniform & fitted prior & + coupling \\
\midrule
Cone & 14 & \best{$\le$1} & \best{$\le$1} & 158 & \best{$\le$1} & \best{$\le$1} \\
Pre-Pro & 83 & 2.8 & \best{$\le$1} & 58 & 52 & \best{6.5} \\
Fire & 1.6 & \best{$\le$1} & \best{$\le$1} & 110 & 123 & \best{20} \\
\bottomrule
\end{tabular*}
\vspace{-1.2em}
\end{table}

We ablate the prior, the coupling and the generator, then examine noise agreement, fitting accuracy and cost, and
sensitivity to the prior.

% We separate the contributions of the fitted prior, posterior coupling,
% and learned generator through component ablations.
% We then examine agreement between training and sampling noise,
% prior fitting accuracy and cost, and sensitivity to fitting quality
% and model choices.

\noindent\textbf{Prior and coupling.} \Cref{tab:ablation} compares three settings with the same network: uniform prior
with independent pairing, fitted prior with independent pairing, and fitted prior with posterior coupling. The RMF kernel
measures where mass lies; the finest scale measures the shape within modes. On Pre-Pro, the fitted prior alone lowers
the RMF-kernel error from 83 to 2.8 noise-floor standard deviations but barely changes the finest scale (58 to 52); on
Fire, the finest-scale error even rises from 110 to 123. Posterior coupling brings it to 6.5 and 20: the prior places
the mass, and the coupling recovers its shape.

% \noindent\textbf{Prior and coupling.}
% \Cref{tab:ablation} compares three settings with the same network:
% the uniform prior with independent pairing, the fitted prior with
% independent pairing, and the fitted prior with posterior coupling.
% The RMF kernel and finer-scale kernels probe distributional
% discrepancies at different spatial resolutions.
% On Pre-Pro, fitting the prior while retaining independent pairing
% reduces the RMF-kernel error from 83 to 2.8 noise-floor standard
% deviations.
% The corresponding improvement at the finest scale is much smaller
% (58 to 52), while the finest-scale error on Fire increases from 110
% to 123.
% Posterior coupling reduces these finest-scale errors to 6.5 on
% Pre-Pro and 20 on Fire.
% Thus, improving the marginal prior alone does not recover the
% fine-scale gains obtained with posterior coupling.

\noindent\textbf{Contribution of the generator.} Sampling the fitted prior directly, without a network, already gives
lower error than one-step samples from the uniform prior on all 13 real and $\SOthree$ datasets
(\cref{tab:frameworks}; Supp.~\sref{supp:frameworks}). On Fire, the error is 110 for the uniform prior, 35 for the
prior alone and 20 after the one-step generator. The generator thus reduces the prior-only error by 2.1--12$\times$ on
the torsion angles and 1.5--1.8$\times$ on Fire and Earthquake; on Flood, Fisher, Line and Swiss roll, where the prior
alone is already close to the data, it changes the error little. On the torsion angles, the memorization ratio is
1.0--1.2, close to the ideal of one, against 1.8--6.4 from the uniform prior (Supp.~\sref{supp:frameworks}).

% \noindent\textbf{Contribution of the generator.}
% Direct samples from the fitted prior, without a neural generator,
% have lower error than one-step uniform-prior samples on all 13
% evaluated real-data and $\SOthree$ datasets
% (Supp.~\sref{supp:frameworks}).
% In this comparison, the errors on Pre-Pro are 54 for the one-step
% uniform-prior baseline, 15 for the fitted prior alone, and 7.3 for
% the prior followed by the one-step generator.
% The corresponding values on Fire are 110, 35, and 20.
% The generator reduces the prior-only error by factors of
% 2.1--12$\times$ on torsion-angle data and 1.5--1.8$\times$ on Fire
% and Earthquake.
% Its additional effect depends on the dataset: on Fisher, Line,
% and Swiss roll, the network changes the prior-only error
% by less than 10\%.
% On the torsion-angle datasets, the memorization ratio is 1.0--1.2,
% compared with 1.8--6.4 for uniform-prior samples.
% These values are closer to the reference value of one under this
% nearest-neighbor diagnostic (Supp.~\sref{supp:frameworks}).
% % AUTHOR CHECK: The preceding coupling ablation reports Pre-Pro errors
% % 58 -> 52 -> 6.5, whereas this prior-only comparison reports
% % 54 -> 15 -> 7.3. Preserve the reported numbers, but reconcile the
% % evaluation settings, seeds/checkpoints, and normalization in the tables.

\noindent\textbf{Agreement between training noise and the prior.} By \cref{prop:marginal}, the gap between the fitted
weights and the mean posterior assignments bounds the total variation between training noise and prior; it is at most
0.030 over all standard fits, and the MMD$^2$ between coupled noise and prior samples is within 1.5 noise-floor standard
deviations (Supp.~\sref{supp:res-q5}). Assigning each point to its most likely component gives similar accuracy on
synthetic arcs and rings, and both are 3--6$\times$ more accurate than independent pairing; posterior sampling is the
scheme that also preserves the prior.

% \noindent\textbf{Agreement between training noise and the prior.}
% For the standard mixture-prior fits evaluated in
% Supp.~\sref{supp:res-q5}, the discrepancy between fitted mixture
% weights and averaged posterior assignments bounds the noise-marginal
% total variation by 0.030, using \cref{prop:marginal}.
% The absolute normalized MMD$^2$ between coupled noise and prior
% samples is at most 1.5 in these comparisons.
% Assigning each observation to its most likely component gives
% similar generation accuracy to posterior sampling on synthetic arcs
% and rings; both reduce the error by a factor of 3--6 relative to
% independent pairing.
% Posterior sampling additionally admits the marginal characterization
% and weight-consistency condition in \cref{prop:marginal}.

\noindent\textbf{Prior fitting accuracy and cost.}
On the synthetic datasets used to evaluate fitting, the procedure
recovers the component count and estimates component means within
$0.5^\circ$ of their known values.
Numerical integration of the fitted densities gives 1.000 on
$S^2$ and the torus and 1.003 on $\SOthree$
(Supp.~\sref{supp:estimation}).
A prior fit takes 11\,s to 8 minutes, and the added cost per
generator-training step is at most 23\% in the tested configurations
(Supp.~\sref{supp:cost}).

\noindent\textbf{Sensitivity to prior fitting and model choices.} A rough prior suffices: fitted on 500 points or with
EM stopped after three iterations, it still lowers one-step error by 2.4 to 9.3$\times$. So does a misspecified one: on
thin rings and long arcs, which tangent-space Gaussians cannot represent exactly, it lowers the finest-scale error by 3.7
to 19$\times$. The defaults ($s=2$, $K$ by BIC) are untuned. Validation tuning reduces Fisher error from 63 to 1.1 noise-floor standard deviations, while the number of HMM states has little effect on fragment statistics (Supp.~\sref{supp:res-q5}).

% The defaults ($s=2$, $K$ by BIC) are not tuned; choosing $s$ and $K$ on validation data lowers the error
% further, for example from 63 to 1.1 noise-floor standard deviations on Fisher. The number of hidden Markov states
% changes the fragment statistics little (Supp.~\sref{supp:res-q5}).

% \noindent\textbf{Sensitivity to prior fitting and model choices.}
% The method retains improvements over uniform initialization in
% the tested rough-fitting settings.
% When the prior is fitted on 500 points or fitting is stopped after
% three iterations, one-step errors remain lower by a factor of
% 2.4 to 9.3.
% On thin rings and long arcs, which the finite tangent-space Gaussian
% mixtures do not represent exactly, the method reduces the finest-scale
% error by a factor of 3.7 to 19.
% The default uses $s=2$ and selects $K$ by BIC.
% Selecting $s$ and $K$ on validation data can improve performance
% substantially, reducing the Fisher error from 63 to 1.1 noise-floor
% standard deviations.
% Thus, the method can benefit from imperfect priors while remaining
% sensitive to particular prior choices.
% Varying the number of hidden Markov states has a limited effect on
% the reported local fragment statistics over the tested range
% (Supp.~\sref{supp:res-q5}).

\section{Conclusion}
\label{sec:conclusion}

We combine statistical priors fitted once to training data with posterior coupling for one-step generation on manifolds. Our entropy analysis shows that, for a fixed target on a compact manifold, uniform initialization maximizes the net average log-volume contraction required by exact invertible transport. The method encourages local training pairs while retaining compatible generators' architectures, loss formulations, and one-step samplers. Across five generator configurations from three families, it reduces reported one-step error in all 45 evaluated generator--dataset pairs by factors of up to 125, and increases the proportion of protein fragments passing the specified geometric checks from 34\% to 93\%. Future work will address global geometry of longer protein chains, HMM marginal consistency, continuous conditioning
and equivariant generation.

\section*{LLM Usage Statement}
We used Claude (Anthropic) as an assistant throughout this project. It edited text in the main paper and supplementary material, and checked the consistency of numbers and terminology. It wrote the plotting code for the figures, which show our results. It also helped write and debug the code for experiments and baseline re-implementations. The research question, method and experimental design are ours. We verified all text, code, results and references, and take full responsibility for this submission.

% One-step generators on manifolds differ in how they learn but share where they start. Uniform noise demands the largest
% contraction of volume and, on data with sharp modes, produces samples that cannot exist. A prior fitted to the training
% data, with a coupling that preserves it, removes most of the problem in three one-step generators without changing their
% networks, objectives or samplers. Continuous conditions and equivariant settings are next.

% \input{sec/0_abstract}    
% \input{sec/1_intro}
% \input{sec/2_formatting}
% \input{sec/3_finalcopy}
{
    \small
    \bibliographystyle{ieeenat_fullname}
    \bibliography{refs}
}


\clearpage
\setcounter{page}{1}
\maketitlesupplementary

\setcounter{section}{0}
\setcounter{equation}{0}
\setcounter{figure}{0}
\setcounter{table}{0}
\counterwithin{proposition}{section}
\renewcommand{\thesection}{\Alph{section}}
\renewcommand{\theequation}{S\arabic{equation}}
\renewcommand{\thefigure}{S\arabic{figure}}
\renewcommand{\thetable}{S\arabic{table}}

% \setcounter{section}{0}
% \setcounter{equation}{0}
% \setcounter{figure}{0}
% \setcounter{table}{0}
% \counterwithin{proposition}{section}
% \renewcommand{\thesection}{\Alph{section}}
% \renewcommand{\theequation}{S\arabic{equation}}
% \renewcommand{\thefigure}{S\arabic{figure}}
% \renewcommand{\thetable}{S\arabic{table}}
% % \newcommand{\Exp}{\operatorname{Exp}}
% % \newcommand{\Log}{\operatorname{Log}}
% % \newcommand{\vol}{\operatorname{vol}}
% % \newcommand{\Lip}{\operatorname{Lip}}
% % \newcommand{\E}{\mathbb{E}}
% % \newcommand{\R}{\mathbb{R}}
% % \newcommand{\SOthree}{\mathrm{SO}(3)}
% % \newcommand{\best}[1]{{\boldmath\textbf{#1}}}
% % \newcommand{\pmcell}[2]{\begin{tabular}[c]{@{}c@{}}$#1$\\[-1pt]$\pm#2$\end{tabular}}
% \newcommand{\cellstack}[1]{\begin{tabular}[c]{@{}c@{}}#1\end{tabular}}
% % \newtheorem{proposition}{Proposition}[section]
\newtheorem{lemma}[proposition]{Lemma}
\newtheorem{corollary}[proposition]{Corollary}
\newtheorem{aremark}[proposition]{Remark}
\newtheorem{definition}[proposition]{Definition}

\section{Notation and background}\label{supp:notation}

\noindent\textbf{Notation.} Throughout, $(\M,g)$ is a Riemannian manifold of dimension $d$ with tangent spaces $T_x\M$,
geodesic distance $d_g$, Riemannian volume measure $\vol$, exponential map $\Exp_x:T_x\M\to\M$ and logarithm map
$\Log_x$, the inverse of $\Exp_x$ on its injectivity domain. All densities are taken with respect to $\vol$. For a
density $p$, $H(p)=-\int p\log p\,d\vol$ is its differential entropy; when $\M$ is compact, $\mathcal U=1/\vol(\M)$ is
the uniform density and $\mathrm{KL}(p\,\|\,\mathcal U)=\int p\log(p/\mathcal U)\,d\vol=\log\vol(\M)-H(p)$. The total
variation between two densities is $\mathrm{TV}(p,q)=\frac12\int|p-q|\,d\vol$. For a measurable map $T:\M\to\M$, the
push-forward $T_\#p$ is the distribution of $T(x)$ for $x\sim p$. Where $T$ is differentiable, $dT_x:T_x\M\to
T_{T(x)}\M$ is its differential, $\|dT_x\|$ its operator norm and $J_T(x)=|\det dT_x|$ its Jacobian with respect to
$\vol$, the determinant being taken in orthonormal bases of $T_x\M$ and $T_{T(x)}\M$;
$\Lip(T)=\sup_{x\neq y}d_g(T(x),T(y))/d_g(x,y)$ is its Lipschitz constant. We write $p_1$ for the density of the data
and $p_0$ for the \emph{prior}, the distribution of the noise from which generation starts.

\noindent\textbf{One-step generators on manifolds.} Flow matching on manifolds \citep{chen2024flow} joins noise $x_0\sim p_0$
and data $x_1\sim p_1$ by geodesics, $x_t=\Exp_{x_0}(t\Log_{x_0}x_1)$ for $t\in[0,1]$, and regresses the velocity of
this path with a neural network; sampling integrates the learned velocity field in many steps. One-step generators
learn the end point of this integration directly. MeanFlow \citep{geng2026mean} learns the \emph{average} velocity
$u(x_t,r,t)$ over an interval $[r,t]$ through an identity that relates it to the instantaneous velocity, so that a
single network evaluation gives a sample, $\hat x_1=\Exp_{x_0}\bigl(u_\theta(x_0,0,1)\bigr)$. The three generators of our experiments build on this idea in different ways:
\begin{itemize}[nosep,leftmargin=*]
\item the Riemannian MeanFlow of \citet{zhong2026riemannian} defines the average velocity on the manifold through parallel
transport and trains it with tangent-space Jacobian--vector products;
\item \citet{woo2026riemannian} learn the flow map with three objectives, derived from its Eulerian, Lagrangian and semigroup
characterizations;
\item generalized flow maps \citep{davis2026generalised} learn the two-time flow map with a Lagrangian self-distillation
objective (G-LSD).
\end{itemize}
All four pair the uniform noise with the data independently.

\noindent\textbf{Conventions for tables.} Unless stated otherwise, tables report the unbiased MMD$^2\times10^3$ of one-step
samples against the held-out test split and, in parentheses, its multiple of the standard deviation of the noise floor
($\le$1: within the estimated noise floor; \cref{supp:eval}). The \emph{finest scale} is the chordal Laplace kernel with
$\gamma=20$ on $\SOthree$, the joint Laplace kernel with $\gamma=10$ on the torus and the Laplace kernel with $\gamma=50$
on $S^2$. Values are means over seeds and over independent draws of $10^4$ samples, with the seeds and checkpoints of
\cref{supp:method}. Bold marks the best value.

\section{Proofs}\label{supp:proofs}
This section proves the two propositions of the main paper and gives three further results: a worst-case form of
\mref{prop:jacobian} (\cref{cor:worst}), its extension to maps that fold (\cref{lem:noninj}), and the densities of
tangent-space Gaussians used in \mref{eq:density}. \Cref{supp:circle} checks \mref{prop:jacobian} exactly on the circle.

Throughout this appendix, $\M$ is a compact connected Riemannian manifold of dimension $d$ without boundary, and the
notation is that of \cref{supp:notation}. Two standard results are used repeatedly.

\begin{lemma}[Differential bounds]\label{lem:diff}
Let $T:\M\to\M$ be differentiable at $x$. Then
\begin{enumerate}[label=(\roman*),nosep]
\item $\|dT_x\|\le\Lip(T)$ if $T$ is Lipschitz;
\item $J_T(x)\le\|dT_x\|^d$.
\end{enumerate}
\end{lemma}
\begin{proof}
(i) Let $u\in T_x\M$ be a unit vector and $c(t)=\Exp_x(tu)$. For small $t>0$, $d_g(c(t),x)=t$, and therefore
\[
\begin{aligned}
\|dT_xu\|&=\lim_{t\downarrow0}\frac{d_g\bigl(T(c(t)),T(x)\bigr)}{t}\\
&\le\lim_{t\downarrow0}\frac{\Lip(T)\,d_g(c(t),x)}{t}=\Lip(T).
\end{aligned}
\]
(ii) In orthonormal bases, $J_T(x)=|\det dT_x|=\prod_{i=1}^d\sigma_i(dT_x)\le\sigma_{\max}(dT_x)^d=\|dT_x\|^d$, where
$\sigma_1,\dots,\sigma_d$ are the singular values of $dT_x$.
\end{proof}

\begin{lemma}[Change of variables; e.g.\ \citealp{lee2018introduction}]\label{lem:cov}
Let $T$ be a $C^1$ diffeomorphism of $\M$. For every non-negative or integrable measurable $f$,
$\int_\M f(T(x))\,J_T(x)\,d\vol(x)=\int_\M f(y)\,d\vol(y)$.
\end{lemma}

\subsection{Proof of \texorpdfstring{\mref{prop:jacobian}}{Proposition 1}}
The main text states the identity of \mref{prop:jacobian} in short form. In full, it holds under the following
conditions and comes with a lower bound on the Lipschitz constant of the inverse map.
\vspace{1em}

\begin{assumption}\label{ass:regularity}
$\M$ is a compact connected Riemannian manifold of dimension $d$ without boundary; $p_0$ and $p_1$ are densities with
finite entropy; and $T$ is a $C^1$ diffeomorphism with $T(x)\sim p_1$ for $x\sim p_0$.
\end{assumption}

\noindent Under Assumption~\ref{ass:regularity},
\begin{equation}
\begin{gathered}
  \E_{x\sim p_0}\bigl[\log J_T(x)\bigr]=H(p_1)-H(p_0),\\[2pt]
  \Lip(T^{-1})\ge\exp\!\bigl((H(p_0)-H(p_1))/d\bigr).
\end{gathered}
\label{eq:lip}
\end{equation}
For $p_0=\mathcal U$, the average contraction equals $\mathrm{KL}(p_1\,\|\,\mathcal U)$, and for fixed $p_1$ it is
maximal if and only if $p_0=\mathcal U$. Steps 1 and 2 prove the identity and Step 3 the bound.

\emph{Step 1: density relation.} For a measurable $A\subseteq\M$, apply \cref{lem:cov} to $f=p_1\mathbf 1_{T(A)}$ and use
$T_\#p_0=p_1$:
\[
\begin{aligned}
&\int_Ap_1(T(x))\,J_T(x)\,d\vol(x)=\int_{T(A)}p_1\,d\vol\\
&\qquad=\Pr_{x\sim p_0}\bigl(T(x)\in T(A)\bigr)=\int_Ap_0\,d\vol.
\end{aligned}
\]
Since $A$ is arbitrary, $p_0=(p_1\circ T)\,J_T$ almost everywhere. Both factors are positive on $\{p_0>0\}$, a set of
full $p_0$-measure, so that $p_0$-almost everywhere
\begin{equation}\label{eq:logjac}
  \log J_T=\log p_0-\log p_1\circ T.
\end{equation}

\emph{Step 2: identity.} Because $T(x)\sim p_1$ for $x\sim p_0$, both terms in \eqref{eq:logjac} are integrable, with
$\E_{p_0}|\log p_0|=\int p_0|\log p_0|\,d\vol<\infty$ and $\E_{p_0}|\log p_1(T(x))|=\int p_1|\log p_1|\,d\vol<\infty$
by Assumption~\ref{ass:regularity}. Hence $\log J_T\in L^1(p_0)$ and
\[
\begin{aligned}
\E_{p_0}[\log J_T]
&\overset{\text{(a)}}{=}\int p_0\log p_0\,d\vol-\E_{x\sim p_0}\bigl[\log p_1(T(x))\bigr]\\
&\overset{\text{(b)}}{=}-H(p_0)-\int p_1\log p_1\,d\vol\\
&=H(p_1)-H(p_0),
\end{aligned}
\]
where (a) is \eqref{eq:logjac} and (b) uses $T_\#p_0=p_1$.

\emph{Step 3: Lipschitz bound.} If $\Lip(T^{-1})=\infty$, there is nothing to prove. Otherwise,
\[
\begin{aligned}
H(p_0)-H(p_1)&=\E_{p_0}\bigl[-\log J_T(x)\bigr]\\
&\overset{\text{(c)}}{=}\E_{p_0}\bigl[\log J_{T^{-1}}(T(x))\bigr]\\
&\overset{\text{(d)}}{\le}d\,\E_{p_0}\bigl[\log\|dT^{-1}_{T(x)}\|\bigr]\\
&\overset{\text{(e)}}{\le}d\log\Lip(T^{-1}),
\end{aligned}
\]
where (c) follows from $dT^{-1}_{T(x)}=(dT_x)^{-1}$, (d) from \cref{lem:diff}(ii) and (e) from \cref{lem:diff}(i).

\emph{Step 4: uniform prior.} Since $H(\mathcal U)=\log\vol(\M)$,
\[
\begin{aligned}
H(\mathcal U)-H(p_1)&=\int p_1\log\bigl(p_1\vol(\M)\bigr)\,d\vol\\
&=\mathrm{KL}(p_1\,\|\,\mathcal U),
\end{aligned}
\]
which is the second statement of \mref{prop:jacobian}. Finally, for every density $p_0$ with finite entropy,
\[
  H(p_0)=H(\mathcal U)-\mathrm{KL}(p_0\,\|\,\mathcal U)\le H(\mathcal U),
\]
with equality if and only if $p_0=\mathcal U$ almost everywhere by Gibbs' inequality, so $H(p_0)-H(p_1)$ is maximal if
and only if $p_0=\mathcal U$ almost everywhere. \hfill$\square$

\subsection{A worst-case consequence}
\begin{corollary}[Worst-case distortion]\label{cor:worst}
Under Assumption~\ref{ass:regularity}, let $p_0$ and $p_1$ be continuous and positive, and let $r(x)=p_0(x)/p_1(T(x))$. Then
$\Lip(T)\,\Lip(T^{-1})\ge(\max_x r(x)/\min_x r(x))^{1/d}$. For the uniform prior, the right-hand side equals
$(\max p_1/\min p_1)^{1/d}$.
\end{corollary}
\begin{proof}
If $p_0$ and $p_1$ are continuous and positive, both sides of $p_0=(p_1\circ T)J_T$ are continuous, so the relation of
Step 1 holds everywhere and $J_T=r$. Since $\M$ is compact, $r$ attains its maximum at some $x^\ast$ and its minimum at
some $x_\ast$, and
\[
\begin{aligned}
\frac{\max r}{\min r}&=\frac{J_T(x^\ast)}{J_T(x_\ast)}
\overset{\text{(c)}}{=}J_T(x^\ast)\,J_{T^{-1}}(T(x_\ast))\\
&\overset{\text{(d)}}{\le}\|dT_{x^\ast}\|^d\,\|dT^{-1}_{T(x_\ast)}\|^d\\
&\overset{\text{(e)}}{\le}\bigl(\Lip(T)\,\Lip(T^{-1})\bigr)^d,
\end{aligned}
\]
with (c)--(e) as in Step 3. For $p_0=\mathcal U$, $r(x)=1/\bigl(\vol(\M)\,p_1(T(x))\bigr)$, and since $T$ is a bijection,
$\{p_1(T(x)):x\in\M\}=\{p_1(y):y\in\M\}$, so that $\max r/\min r=\max p_1/\min p_1$.
\end{proof}

\subsection{Maps that fold}
\begin{lemma}\label{lem:noninj}
Let $T:\M\to\M$ be Lipschitz, and let $p_0$ and $p_1=T_\#p_0$ be densities with
$\int p_i|\log p_i|\,d\vol<\infty$. If $J_T>0$ $p_0$-almost everywhere and $\log J_T\in L^1(p_0)$, then
\[
  H(p_1)\le H(p_0)+\E_{p_0}[\log J_T],
\]
with equality if $T$ is injective on a set of full $p_0$-measure.
\end{lemma}
\begin{proof}
By Rademacher's theorem, $T$ is differentiable almost everywhere, and the area formula for Lipschitz maps
\citep[Sec.~3.2]{federer2014geometric} gives, for almost every $y$,
\[
  p_1(y)=\sum_{x\in T^{-1}(y),\;J_T(x)>0}\frac{p_0(x)}{J_T(x)}.
\]
All summands are non-negative, so $p_1(T(x))\ge p_0(x)/J_T(x)$ for $p_0$-almost every $x$. Hence
\[
\begin{aligned}
H(p_1)&\overset{\text{(f)}}{=}-\E_{x\sim p_0}\bigl[\log p_1(T(x))\bigr]\\
&\overset{\text{(g)}}{\le}-\E_{x\sim p_0}\Bigl[\log\frac{p_0(x)}{J_T(x)}\Bigr]
=H(p_0)+\E_{p_0}[\log J_T],
\end{aligned}
\]
where (f) uses $T_\#p_0=p_1$ and (g) the lower bound on $p_1(T(x))$. If $T$ is injective on a set of full
$p_0$-measure, the sum has one term for $p_0$-almost every $x$, and (g) is an equality.
\end{proof}
\noindent A trained one-step map $x\mapsto\Exp_x(u_\theta(x,0,1))$ is Lipschitz on the compact manifold when the network
is. Folding, i.e.\ several preimages of the same point, makes (g) strict, which is how such a map can concentrate mass
with less average contraction than \mref{prop:jacobian} requires of an injective map; \mref{tab:geometry} of the
main paper measures how often trained maps do so.

\subsection{Proof of \texorpdfstring{\mref{prop:marginal}}{Proposition 2}}
Write $\hat q_k=q(\cdot;\hat\mu_k,s\hat\Sigma_k)$. Under the posterior coupling, draw $i$ uniformly from
$\{1,\dots,n\}$, then $k\sim\hat\gamma(\cdot\mid x_i)$ and $\varepsilon\sim\hat q_k$. Then
$\Pr(k)=\frac1n\sum_i\hat\gamma(k\mid x_i)=\bar w_k$, and since $\varepsilon$ depends on $i$ only through $k$, its
density is $\bar p_0=\sum_k\bar w_k\hat q_k$. With $p_0=\sum_k\hat w_k\hat q_k$,
\[
\begin{aligned}
\mathrm{TV}(\bar p_0,p_0)&=\frac12\int\Bigl|\sum_{k=1}^K(\bar w_k-\hat w_k)\,\hat q_k\Bigr|\,d\vol\\
&\overset{\text{(h)}}{\le}\frac12\sum_{k=1}^K|\bar w_k-\hat w_k|\int\hat q_k\,d\vol\\
&=\frac12\sum_{k=1}^K|\bar w_k-\hat w_k|,
\end{aligned}
\]
where (h) is the triangle inequality. At a fixed point of the EM weight update (\cref{eq:em}), $\hat w_k$ equals the
average of the responsibilities computed with the fitted parameters, i.e.\ $\hat w=\bar w$, and then $\bar p_0=p_0$.
For the hierarchical OT coupling, let $G_k$ be the indices of a minibatch assigned to component $k$. The noise
$\{\varepsilon_b\}_{b\in G_k}$ is drawn independently from $\hat q_k$ and then permuted, so the multiset
$\{\varepsilon_b\}_{b\in G_k}$, and with it the empirical distribution of the noise of every minibatch, is unchanged.
\hfill$\square$

\subsection{Densities of tangent-space Gaussians}
\label{supp:densities}
Fix $\mu\in\M$ and an orthonormal basis $B_\mu$ of $T_\mu\M$, and write $\phi(v)=\Exp_\mu(B_\mu v)$ for $v\in\R^d$. Let
$\mathcal D\subset\R^d$ be the injectivity domain of $\phi$, and let $j(v)=|\det d\phi_v|$, computed in the standard basis
of $\R^d$ and an orthonormal basis of $T_{\phi(v)}\M$.

\begin{lemma}\label{lem:density}
If $\phi$ maps $\mathcal D$ diffeomorphically onto an open set of full volume and $v$ has a probability density $f$ on
$\mathcal D$, then $x=\phi(v)$ has the density $f(\phi^{-1}(x))/j(\phi^{-1}(x))$ with respect to $\vol$.
\end{lemma}
\begin{proof}
For a measurable $A\subseteq\M$,
\[
\begin{aligned}
\Pr(x\in A)&=\int_{\phi^{-1}(A)}f(v)\,dv\\
&\overset{\text{(a)}}{=}\int_{A\cap\phi(\mathcal D)}\frac{f(\phi^{-1}(x))}{j(\phi^{-1}(x))}\,d\vol(x)\\
&\overset{\text{(b)}}{=}\int_A\frac{f(\phi^{-1}(x))}{j(\phi^{-1}(x))}\,d\vol(x),
\end{aligned}
\]
where (a) is the change-of-variables formula for the diffeomorphism $\phi:\mathcal D\to\phi(\mathcal D)$
\citep{lee2018introduction}, and (b) holds because $\M\setminus\phi(\mathcal D)$ has volume zero.
\end{proof}
\noindent With $f$ the density of $\mathcal N(0,\Sigma)$ truncated to $\mathcal D$, the lemma gives \mref{eq:density}.
On a complete manifold, $\phi(\mathcal D)$ is the complement of the cut locus of $\mu$, which has volume zero, so the
lemma applies to the three manifolds of this paper, for which $\mathcal D$ and $j$ are derived next.

\emph{Sphere $S^2$.} Write $v=r(\cos\vartheta,\sin\vartheta)$ in polar coordinates. Then $\phi(v)$ is the point at
geodesic distance $r$ from $\mu$ in direction $\vartheta$, and the area element of $S^2$ in these geodesic polar
coordinates is $\sin r\,dr\,d\vartheta$, whereas the Lebesgue measure on $\R^2$ is $r\,dr\,d\vartheta$. Hence
\[
  j(v)=\frac{\sin r}{r},\qquad \mathcal D=\{v:\|v\|<\pi\},
\]
and the cut locus of $\mu$ is the antipode $-\mu$.

\emph{Rotation group $\SOthree$.} For $v\in\R^3$, let $[v]_\times$ be the skew-symmetric matrix with
$[v]_\times w=v\times w$. We use the bi-invariant metric under which $R\mapsto R\exp([v]_\times)$ is a geodesic of length
$r=\|v\|$, the rotation angle. By the formula for the differential of the exponential map of a Lie group
\citep{varadarajan2013lie}, $d\exp_v$ equals left translation composed with $(I-e^{-\mathrm{ad}_v})/\mathrm{ad}_v$, and
$\mathrm{ad}_v$ has the eigenvalues $0$ and $\pm ir$ on $\mathfrak{so}(3)\cong\R^3$. Therefore
\[
\begin{aligned}
j(v)&=\Bigl|\det\frac{I-e^{-\mathrm{ad}_v}}{\mathrm{ad}_v}\Bigr|
=\prod_{\lambda\in\{0,\,ir,\,-ir\}}\Bigl|\frac{1-e^{-\lambda}}{\lambda}\Bigr|\\
&=\frac{|1-e^{-ir}|^2}{r^2}=\frac{2(1-\cos r)}{r^2},
\end{aligned}
\]
where the factor for $\lambda=0$ is its limit $1$. Here $\mathcal D=\{v:\|v\|<\pi\}$, the cut locus of $R$ consists of
the rotations $R\exp([v]_\times)$ with $\|v\|=\pi$, a set of dimension two, and the total volume is
\[
\begin{aligned}
\vol(\SOthree)&=\int_{\|v\|<\pi}j(v)\,dv=\int_0^\pi4\pi r^2\,\frac{2(1-\cos r)}{r^2}\,dr\\
&=8\pi\int_0^\pi(1-\cos r)\,dr=8\pi^2.
\end{aligned}
\]

\emph{Flat torus $\mathbb T^d=[0,2\pi)^d$.} Here $\phi(v)=(\mu+v)\bmod2\pi$ maps $\mathcal D=(-\pi,\pi)^d$ isometrically
onto the complement of a finite union of coordinate hyperplanes, so that $j\equiv1$.

\emph{Truncation constant.} $Z(\Sigma)=\Pr_{v\sim\mathcal N(0,\Sigma)}(v\in\mathcal D)$ is estimated by Monte Carlo
with $N=2\times10^5$ samples. The relative standard error of the estimate $\hat Z$ is
$\sqrt{(1-Z)/(NZ)}$, which is below $0.3\%$ for $Z\ge0.1$. Sampling, and hence \mref{prop:marginal}, does not depend on
$Z$.

\subsection{An exact example on the circle}\label{supp:circle}
\begin{figure}[t]
\centering
\includegraphics[width=\linewidth]{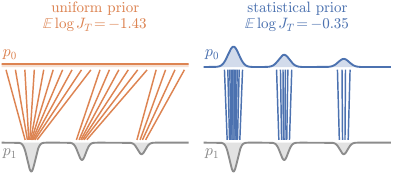}
\caption{\textbf{Exact transport on the circle} from the uniform prior (left) and from a statistical prior (right) onto
three narrow von Mises modes; each line joins a quantile of $p_0$ to its image under $T=F_1^{-1}\circ F_0$.}
\label{fig:circle}
\end{figure}
The data density on the circle $S^1=[0,2\pi)$ is a mixture of three von Mises densities with means $1.0$, $2.7$ and
$4.7$, concentration $60$ and weights $0.5$, $0.3$ and $0.2$. The statistical prior has the same means and weights and
concentration $30$, so that it is wider than the data. On the circle, the increasing transport map from $p_0$ to $p_1$
that fixes the point $0$ is $T=F_1^{-1}\circ F_0$, where $F_i$ is the distribution function of $p_i$ on $[0,2\pi)$; it
satisfies $T_\#p_0=p_1$ and is a diffeomorphism because both densities are positive. Each line of \cref{fig:circle} joins one of 20 equally spaced quantiles of $p_0$ to its image. We evaluate $F_0$, $F_1$, $T$ and its derivative on a grid of
$2\times10^5$ points. The average log-Jacobian is $-1.43$ from the uniform prior and $-0.35$ from the statistical prior,
equal to $H(p_1)-H(p_0)$ to two decimals in both cases, as \mref{prop:jacobian} states.

\section{Method and implementation}\label{supp:method}
This section gives what is needed to reproduce the method: how the prior is fitted and sampled, how the models are
trained in each framework, and what the method costs.
\subsection{Estimation and model selection}\label{supp:estimation}
Let $x_1,\dots,x_n$ be the training set. We fit \mref{eq:prior} with $s=1$ by EM on the manifold. Given current parameters $(w_k,\mu_k,\Sigma_k)$, the E-step computes the responsibilities and the M-step updates the parameters:
\begin{equation}
\begin{gathered}
  \gamma(k\mid x_i)=\frac{w_k\,q(x_i;\mu_k,\Sigma_k)}{\sum_{l=1}^Kw_l\,q(x_i;\mu_l,\Sigma_l)},\\[2pt]
  w_k\leftarrow\frac1n\sum_{i=1}^n\gamma(k\mid x_i),\\[2pt]
  \mu_k\leftarrow\Exp_{\mu_k}\Bigl(\frac{\sum_i\gamma(k\mid x_i)\Log_{\mu_k}x_i}{\sum_i\gamma(k\mid x_i)}\Bigr),
\end{gathered}
\label{eq:em}
\end{equation}
The mean update is one Gauss--Newton step toward the weighted Fr\'echet mean, repeated a few times, and $\Sigma_k$ is set to the weighted covariance of the normal coordinates $B_{\mu_k}^\top\Log_{\mu_k}x_i$. We write
$(\hat w_k,\hat\mu_k,\hat\Sigma_k)$ for the fitted parameters and $\hat\gamma(k\mid x)$ for the corresponding
responsibilities. The E-step and the weight update are exact; the updates of $\mu_k$ and $\Sigma_k$ match weighted moments in the tangent space. Initialization and stopping rule are given in \cref{supp:impl}. The number of components is selected by BIC, $-2\log\hat L+[K(d+d(d+1)/2)+K-1]\log n$, where $\hat L$ is the
likelihood of the fitted mixture computed exactly from \mref{eq:density}, or by held-out likelihood. \Cref{tab:fit} reports the accuracy of the fitted densities. The mixture is fitted once, before training.

\subsection{Structured priors on products of tori}\label{supp:hmm}
The number of parameters of the mixture grows quadratically with the dimension, which matters for products of tori. The
torsion angles of $L$ consecutive residues lie on $(\mathbb T^2)^L$, and a mixture on the full $2L$-dimensional torus has
$O(KL^2)$ covariance parameters and is poorly estimated for $L\ge10$, although the dependence between angles is
concentrated between neighboring residues. The hidden-Markov prior is defined over sites. Each residue has a state $z_j\in\{1,\dots,S\}$, and the states follow a Markov chain with transition matrix $A$. Each state emits a truncated wrapped Gaussian on $\mathbb T^2$, whose covariance the prior inflates by $s$. The model is fitted by Baum--Welch with $S$ selected by BIC, has $O(S^2)$ parameters
independently of $L$, and admits exact sampling and density evaluation. The posterior coupling draws a state path
$z\sim p(z\mid x)$ by forward filtering--backward sampling and emits the noise from the inflated emissions. For this model, the analogue of \mref{prop:marginal} holds for the first state and the pooled transition frequencies: at a fixed point of Baum--Welch, both match the prior. On fragments of 10
residues, the model improves the held-out log-likelihood per fragment from $-17.5$ for a 32-component mixture to
$-7.8$, with about one tenth of the parameters. As a prior, it reproduces the local statistics of real fragments, though not their end-to-end distances (\mref{tab:highdim}).

\noindent\textbf{Forward filtering and backward sampling.} For a chain $\theta=(\theta_1,\dots,\theta_L)$ with initial
distribution $\nu$, transition matrix $A$ and emissions $q(\cdot;\mu_z,\Sigma_z)$, the coupling draws the state path from
its posterior by the recursion
\begin{equation}
\begin{gathered}
  \alpha_1(z)=\nu_z\,q(\theta_1;\mu_z,\Sigma_z),\\[2pt]
  \alpha_j(z)=q(\theta_j;\mu_z,\Sigma_z)\textstyle\sum_{z'}\alpha_{j-1}(z')A_{z'z}\quad(j=2,\dots,L),\\[2pt]
  z_L\sim\alpha_L,\qquad z_j\mid z_{j+1}\sim\alpha_j(z_j)\,A_{z_jz_{j+1}}\quad(j<L),
\end{gathered}
\label{eq:ffbs}
\end{equation}
with each line normalized, in $O(LS^2)$ time, and then draws $\varepsilon_j\sim q(\cdot;\mu_{z_j},s\Sigma_{z_j})$. The
forward pass with $s\Sigma_z$ in place of $\Sigma_z$ also evaluates the density of the prior.

\subsection{Implementation of the prior}\label{supp:impl}
\noindent\textbf{EM on the manifold.} The algorithm is initialized by $k$-means$++$ with the geodesic distance, followed by
20 Lloyd iterations (with spherical means on $S^2$, chordal means projected onto $\SOthree$ by a singular value
decomposition, and circular means on $\mathbb T^d$). The update of each mean takes three Gauss--Newton steps, and $10^{-6}I$ is added to each covariance. The iterations stop after 100 steps or
when the mean log-likelihood changes by less than $10^{-7}$. The constant $Z(\Sigma_k)$ is estimated with $2\times10^4$
Monte Carlo samples during EM and with $2\times10^5$ samples for the final model.

\noindent\textbf{Model selection.} The candidate numbers of components are $K\in\{1,2,3,4,6,8,12,16,24,32\}$, extended to 48,
64, 96 and 128 where stated, and BIC is computed with $K(d+d(d+1)/2)+K-1$ free parameters. The prior is fitted once,
before training, on at most $10^4$ training points, using a random subsample for larger datasets.

\noindent\textbf{Sampling and coupling.} Samples from a component are drawn by rejecting $v\notin\mathcal D$; the acceptance
rate is close to one for concentrated components. The responsibilities of the coupling use the fitted covariances ($s=1$); the noise comes from the inflated components ($s\,\hat\Sigma_k$). For hierarchical OT, the data points of a minibatch that share a component are shuffled and cut into blocks of at most 128 points. The Hungarian algorithm then matches each block to the noise of that component, with cost $\|\Log_{\varepsilon}(x)\|^2$.

\begin{figure}[t]
\centering
\includegraphics[width=\linewidth]{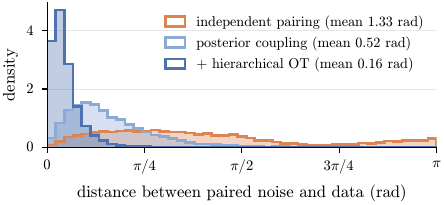}
\caption{\textbf{The posterior coupling keeps moves short.} Distance between paired noise and data on Pre-Pro.}
\label{fig:coupling-dist}
\end{figure}

\noindent\textbf{Component-wise optimal matching.} Within a minibatch, the data points assigned to the same component are
cut into blocks $G$ of at most 128, and the Hungarian algorithm solves
\begin{equation}
  \min_{\sigma}\;\sum_{i\in G}\bigl\|\Log_{\varepsilon_{\sigma(i)}}x_i\bigr\|^2\label{eq:hot}
\end{equation}
over the permutations $\sigma$ of the noise drawn for that block. It only permutes noise that has already been drawn,
so the distribution of the training noise is unchanged. \Cref{tab:pairings} compares the pairing schemes, and
\cref{fig:coupling-dist} shows the distances they produce on Pre-Pro; all main results use the posterior coupling
alone.

\begin{table}[t]
\centering\footnotesize\setlength{\tabcolsep}{3pt}
\caption{\textbf{Pairing schemes and their properties.} Local: the noise is drawn near its data point. Prior kept: the
training noise follows the prior $p_0$ used for sampling (\mref{prop:marginal}); component-wise matching only permutes
noise that has been drawn.}
\label{tab:pairings}
\begin{tabular*}{\linewidth}{@{\extracolsep{\fill}}lll@{}}
\toprule
Pairing & local & prior kept \\
\midrule
Independent & no & yes \\
Most likely component & yes & not in general \\
Posterior component (ours) & yes & at EM fixed points \\
\quad + component-wise matching & shorter & as above \\
\bottomrule
\end{tabular*}
\end{table}

\noindent\textbf{Validation of the estimator.} On synthetic mixtures with known parameters, BIC recovers the true number of
components on $S^2$, $\SOthree$ and $\mathbb T^2$, including a component that crosses the $0/2\pi$ boundary and one with a
$\phi$--$\psi$ correlation of $0.7$. The weights are recovered to within $0.013$, the means to within $0.5^\circ$, and the
eigenvalues of the covariances to within a few percent. The fitted densities integrate to $1.0000$ on $S^2$ and
$\mathbb T^2$ and to $1.003\pm0.009$ on $\SOthree$ (Monte Carlo), and the samples agree with the densities in the
expectations of test functions.

\noindent\textbf{Accuracy of the fitted priors.} On the synthetic data, the statistical prior comes within 0.02 nats of the
true held-out log-likelihood except on thin rings (0.22 nats), and it gains 1.8--17 nats over the uniform density on the
real data (\cref{tab:fit}).

\begin{table}[t]
\centering\footnotesize\setlength{\tabcolsep}{3pt}
\caption{\textbf{Accuracy of the fitted priors}: mean held-out log-likelihood (priors fitted on 8000 points, evaluated on
$2\times10^4$ independent samples). In parentheses: $K$ selected by BIC.}
\label{tab:fit}
\textbf{(a)} Synthetic data on $S^2$, true density known\\[2pt]
\begin{tabular*}{\linewidth}{@{\extracolsep{\fill}}lcccc@{}}
\toprule
Data & true density & statistical & vMF, $K=8$ & uniform \\
\midrule
Ring, 100 mrad & $-1.324$ & $-1.337$ (12) & $-1.610$ & $-2.531$ \\
Ring, 30 mrad & $-0.385$ & $-0.608$ (16) & $-1.478$ & $-2.531$ \\
Arc, 30 mrad & $0.150$ & $0.144$ (4) & $-0.895$ & $-2.531$ \\
\bottomrule
\end{tabular*}\\[6pt]
\textbf{(b)} Real data\\[2pt]
\begin{tabular*}{\linewidth}{@{\extracolsep{\fill}}lcc@{}}
\toprule
Data & statistical & uniform \\
\midrule
Cone ($\SOthree$) & $4.586$ (2) & $-4.369$ \\
Glycine ($\mathbb T^2$) & $-1.915$ (12) & $-3.676$ \\
Proline ($\mathbb T^2$) & $-0.167$ (12) & $-3.676$ \\
RNA ($\mathbb T^7$) & $4.495$ (24) & $-12.865$ \\
\bottomrule
\end{tabular*}
\end{table}

\subsection{Training details}\label{supp:training}
The architectures follow \citet{zhong2026riemannian}. On the Earth data we use a TMLP with 7--12 layers of width 2048, on the synthetic sphere data 6 layers of width 512 and on $\SOthree$ 4 layers of width 512 (8 of width 1024 for the Swiss roll). On the torus we keep the configurations of that work. We train with AdamW, a cosine schedule and, by default, the MeanFlow objective, without early stopping. Training lasts 200 epochs on the synthetic sphere data and $\SOthree$, 100 on the Earth data and 1000 on the torus. Two additional runs use 400 epochs on Fire and 5000 epochs with the uniform prior on Proline. Final
checkpoints are evaluated, except on Proline, Pre-Pro and RNA, where both methods use the best checkpoint on validation
data, which favors the uniform prior. By default the prior uses $s=2$, the posterior coupling and at most 32
components. We use three seeds on the torus (one on General), two on the synthetic, $\SOthree$ and Earth data, and one or
two for the ablations. Training on one NVIDIA A40 GPU takes 10--50 minutes per model, and about 2.5 hours on General.

\subsection{Integration into the other frameworks}\label{supp:frameworks-impl}
For every framework, only the prior and the pairing change. Both priors share the code, architecture, objective, optimizer, training budget and seeds. The statistical prior is the one used with the Riemannian MeanFlow of \citet{zhong2026riemannian}: at most 32 components, $s=2$ and the posterior coupling.

\noindent\textbf{Riemannian MeanFlow of \citet{woo2026riemannian}.} We use the official code with the authors' configurations for
the Eulerian, Lagrangian and semigroup formulations. The Earth data are trained for 500 epochs, the authors' default,
and the torsion angles for the authors' 10\,000 epochs on $\mathbb T^2$ and their default budget on RNA. Each model is
trained with two seeds and sampled with 1, 2 and 4 steps.

\noindent\textbf{Generalized flow maps \citep{davis2026generalised}.} We implement the Lagrangian self-distillation objective (G-LSD), with a flow-matching term of weight one, in the code of \citet{zhong2026riemannian}. It is trained with the budgets of that work, 100 epochs on the Earth data and 1000 on the torus, and two seeds.

\subsection{Computational cost}\label{supp:cost}
\Cref{tab:stress}b gives the cost of the posterior coupling; it also leaves the memory unchanged. Hierarchical
OT solves the matching on the CPU, at a cost that does not depend on the network. Its overhead is therefore 31\% on Fire
and a factor of 1.8--3.1 for the small networks of Cone and Pre-Pro (\cref{tab:cost}a). Blocks of 64--128 points
minimize this cost; we use 128 (\cref{tab:cost}b). Times are means over 200 steps after 20 warm-up steps on one NVIDIA
L40S GPU.

\begin{table*}[t]
\footnotesize\setlength{\tabcolsep}{3pt}
\centering
\caption{\textbf{Cost of hierarchical OT} on one NVIDIA L40S GPU. (a) Time per training step (ratio to the uniform prior)
and peak memory. (b) Time (ms) to pair one minibatch, by block size.}
\label{tab:cost}
\begin{minipage}[t]{0.46\textwidth}\centering
\textbf{(a)} Time per step and memory\\[2pt]
\begin{tabular*}{\linewidth}{@{\extracolsep{\fill}}l c cc c}
\toprule
 & & \multicolumn{2}{c}{time per step (ms)} & \\
\cmidrule(lr){3-4}
Dataset & batch & uniform & hierarchical OT & memory \\
\midrule
Fire ($S^2$) & 8192 & 202 & 264 (1.3) & 11.7\,GB \\
Cone ($\SOthree$) & 1024 & 14.7 & 26.6 (1.8) & 0.3\,GB \\
Pre-Pro ($\mathbb T^2$) & 4096 & 11.2 & 34.3 (3.1) & 0.6\,GB \\
\bottomrule
\end{tabular*}
\end{minipage}\hfill
\begin{minipage}[t]{0.52\textwidth}\centering
\textbf{(b)} Time to pair one minibatch (ms)\\[2pt]
\begin{tabular*}{\linewidth}{@{\extracolsep{\fill}}l c c ccccc}
\toprule
& & & \multicolumn{5}{c}{hierarchical OT, block size} \\
\cmidrule(lr){4-8}
Dataset & batch, $K$ & posterior & 32 & 64 & 128 & 256 & 512 \\
\midrule
Fire & 8192, 32 & 1.3 & 89.3 & \best{63.3} & 66.5 & 107 & 210 \\
Cone & 1024, 3 & 1.8 & 23.8 & 15.6 & \best{12.9} & 17.5 & 26.5 \\
Pre-Pro & 4096, 8 & 1.2 & 17.9 & \best{16.5} & 23.9 & 51.5 & 126 \\
\bottomrule
\end{tabular*}
\end{minipage}
\end{table*}

\section{Evaluation protocol}\label{supp:eval}
This section details the evaluation summarized in \mref{sec:experiments}. We first describe the standard protocol and
three problems that we found with it (\cref{supp:protocol}), then our evaluation (\cref{supp:ourevaluation}), the proof that its kernels are positive definite and characteristic
(\cref{supp:kernels}) and further details (\cref{supp:evaldetails}).

\subsection{The standard protocol and its limitations}\label{supp:protocol}
Results on the Earth and torus benchmarks \citep{zhong2026riemannian,davis2026generalised} are compared by the maximum mean discrepancy \citep{gretton2012kernel} between generated and held-out samples. For a
kernel $k$ and distributions $P$ and $Q$,
\[
  \mathrm{MMD}_k^2(P,Q)=\E\,k(x,x')+\E\,k(y,y')-2\,\E\,k(x,y)
\]
for independent $x,x'\sim P$ and $y,y'\sim Q$,
and from samples it is estimated either without bias, averaging $k$ over pairs of distinct points (a U-statistic), or
by averaging over all pairs (a V-statistic). Following \citet{zhong2026riemannian}, the kernel is the geodesic Gaussian kernel $k(x,y)=\exp(-d_g(x,y)^2)$, averaged over the angles on the torus; we call it the \emph{RMF kernel}. The reported value is the square root of a V-statistic, the \emph{RMF estimator}. Together, they form the \emph{RMF protocol}. Since our conclusions depend on how reliable this protocol is, we examined
it using real data only (\cref{tab:audit}; details in \cref{supp:evaldetails}) and found three problems.

\begin{table*}[t]
\footnotesize\setlength{\tabcolsep}{3pt}
\centering
\caption{\textbf{Examination of the RMF protocol using real data only.} Top: \emph{Floor} is the MMD between two disjoint
real subsets of test-set size (20 repetitions), the score of a perfect model; \emph{Within floor} counts the baselines
below floor mean $+2$ s.d.; \emph{Neg.\ eig.}\ counts negative eigenvalues of the Gram matrix on 1500 points. Bottom:
real data with $\psi$ and higher angles permuted, scored with the RMF kernel and with a joint kernel.}
\label{tab:audit}
\begin{tabular*}{\linewidth}{@{\extracolsep{\fill}}lrcccr}
\toprule
Dataset & $n_{\text{test}}$ & Floor & Lowest baseline & Within floor & Neg.\ eig. \\
\midrule
Volcano    & 82    & $0.121\pm0.035$ & 0.092 & 3/4 & 36 \\
Earthquake & 612   & $0.044\pm0.015$ & 0.032 & 3/4 & 156 \\
Flood      & 487   & $0.047\pm0.011$ & 0.048 & 3/4 & 179 \\
Fire       & 1280  & $0.029\pm0.010$ & 0.027 & 3/4 & 63 \\
\midrule
General    & 13820 & $0.006\pm0.002$ & 0.020 & 0/4 & 400 \\
Glycine    & 1328  & $0.025\pm0.007$ & 0.030 & 2/4 & 630 \\
Proline    & 763   & $0.023\pm0.010$ & 0.040 & 2/4 & 461 \\
Pre-Pro    & 691   & $0.027\pm0.009$ & 0.050 & 0/4 & 152 \\
RNA        & 947   & $0.019\pm0.003$ & 0.070 & 0/4 & 198 \\
\midrule
Cone       & 2000  & $0.022\pm0.006$ & 0.044 & 0/4 & 797 \\
Fisher     & 4000  & $0.022\pm0.003$ & 0.039 & 0/4 & 650 \\
Line       & 4000  & $0.021\pm0.003$ & 0.035 & 0/4 & 507 \\
\bottomrule
\end{tabular*}

\vspace{0.8em}
\begin{tabular*}{\linewidth}{@{\extracolsep{\fill}}lcccc}
\toprule
& \multicolumn{2}{c}{RMF kernel (additive)} & \multicolumn{2}{c}{Joint kernel} \\
\cmidrule(lr){2-3}\cmidrule(lr){4-5}
Dataset & Trivial model & Lowest baseline & Trivial model & Real data \\
\midrule
General & \best{0.007} & 0.020 & 0.201 & 0.009 \\
Glycine & \best{0.029} & 0.030 & 0.182 & 0.033 \\
Proline & \best{0.023} & 0.040 & 0.056 & 0.031 \\
Pre-Pro & \best{0.032} & 0.050 & 0.105 & 0.039 \\
RNA     & \best{0.020} & 0.070 & 0.279 & 0.034 \\
\bottomrule
\end{tabular*}
\end{table*}

\noindent\textbf{Noise floor.} The first problem concerns the scale of the scores. A perfect model would obtain the MMD
between two disjoint real subsets of test-set size, and on the Earth datasets this value is as large as the scores of the three best baselines. On Volcano, with 82 test points, it is $0.121\pm0.035$, compared with $0.092$--$0.115$ for the
baselines, and on every Earth dataset three of the four baselines lie within two standard deviations of it. On
these datasets, the protocol therefore cannot distinguish the baselines from a perfect model.

\noindent\textbf{The torus kernel ignores dependence.} The second problem is specific to the torus, where the RMF kernel
averages one-dimensional kernels over the angles and therefore compares only their marginal distributions. To
illustrate this, we permute $\psi$ independently of $\phi$ in the real data, which keeps the marginals but destroys the
joint structure. This trivial model \emph{matches or improves on the best baseline on all five torus datasets} (RNA: $0.020$ against $0.07$). A joint kernel detects it at once ($0.279$ against $0.034$ for real data).

\noindent\textbf{The kernel is not positive definite.} The third problem is more fundamental. The geodesic Gaussian kernel is
positive definite for all bandwidths only on flat spaces \citep{feragen2015geodesic}, and its Gram matrices have
hundreds of negative eigenvalues on every benchmark (797 on Cone), so that the resulting MMD is not a metric. On
$\SOthree$, in contrast, the noise floor ($\approx0.022$) is well below the baseline values ($\ge0.035$), so that
comparisons there remain meaningful despite this issue.

\Cref{tab:rmf} gives the one-step error of both priors under the RMF kernel, on which the main text reports its
comparison with the original protocol.

\begin{table}[t]
\centering\footnotesize\setlength{\tabcolsep}{2pt}
\caption{\textbf{One-step error under the RMF kernel} of \citet{zhong2026riemannian}, in noise-floor standard deviations
(lower is better; $\le$1: within the estimated noise floor).}
\label{tab:rmf}
\begin{tabular*}{\linewidth}{@{\extracolsep{\fill}}lcc lcc@{}}
\toprule
Dataset & uniform & statistical & Dataset & uniform & statistical \\
\midrule
Fire & 1.6 & $\le$1 & General & 176 & $\le$1 \\
Earthquake & $\le$1 & $\le$1 & Glycine & 1.3 & $\le$1 \\
Flood & 2.9 & 2.7 & Proline & 31 & $\le$1 \\
Volcano & $\le$1 & $\le$1 & Pre-Pro & 49 & $\le$1 \\
 & &  & RNA ($\mathbb T^7$) & 146 & $\le$1 \\
\bottomrule
\end{tabular*}
\end{table}

\subsection{Our evaluation}\label{supp:ourevaluation}
Our evaluation rests on three choices. First, we use kernels that are positive definite and characteristic, so that the MMD is a metric on distributions. On $S^2$, $\SOthree$ and $\mathbb T^d$, these are the Laplace kernels
\begin{equation}\label{eq:kernels}
\begin{aligned}
k_{S^2}(x,y)&=e^{-\gamma\arccos\langle x,y\rangle},\\
k_{\SOthree}(R,R')&=e^{-\gamma\|R-R'\|_F/\sqrt2},\\
k_{\mathbb T^d}(x,y)&=\textstyle\prod_{i=1}^de^{-\gamma|\mathrm{wrap}(x_i-y_i)|},
\end{aligned}
\end{equation}
where $x,y\in S^2\subset\R^3$ are unit vectors with inner product $\langle x,y\rangle$, $\|\cdot\|_F$ is the Frobenius
norm of rotation matrices $R,R'\in\SOthree$, and $\mathrm{wrap}(a)\in(-\pi,\pi]$ is $a$ reduced modulo $2\pi$. We prove
both properties in \cref{supp:kernels}, because sampled Gram matrices cannot reveal a failure at high frequencies.
Second, we evaluate at several scales, $\gamma\in\{1,10,50\}$ on $S^2$, $\{1,5,20\}$ on $\SOthree$ and $\{1,3,10\}$ on
$\mathbb T^d$, and call the largest the \emph{finest scale}; our conclusions hold at every $\gamma$ (\cref{fig:gamma}).
Third, we calibrate every score against a \emph{noise floor} $\sigma_0$, the standard deviation of MMD$^2$ between a
test-sized subset of the real data and the remaining real data. Tables report MMD$^2\times10^3$ with its ratio to
$\sigma_0$ in parentheses; a value is \emph{within the floor} if this ratio is at most one, and two methods are
\emph{separated} if their values differ by more than $3\sigma_0$ . We also report
two log-likelihood proxies (\cref{tab:ll}), the true log-likelihood on synthetic data, and a memorization ratio on real
data (\cref{supp:evaldetails}).

\subsection{Further details}\label{supp:evaldetails}

\noindent\textbf{Estimator and kernels.} We use the unbiased estimator of MMD$^2$ with the kernels \eqref{eq:kernels} and, for
comparison, the RMF kernel. On 1500 points of Cone, the Gram matrix of the geodesic Gaussian kernel has 731 negative
eigenvalues; \cref{supp:kernels} proves that the kernels \eqref{eq:kernels} have none.

\begin{figure*}[t]
\centering
\vspace{-1em}
\includegraphics[width=\textwidth]{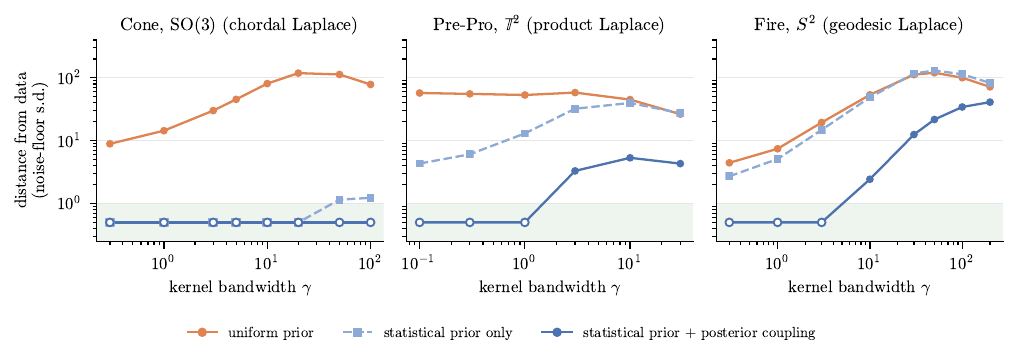}
\vspace{-2em}
\caption{\textbf{Sensitivity to the kernel bandwidth.} One-step MMD$^2$ in multiples of the noise-floor standard
deviation as a function of $\gamma$; values within the floor are drawn at the dashed line. MeanFlow, 2--3 seeds, two draws of 5000 samples.}
\label{fig:gamma}
\vspace{-1em}
\end{figure*}

\noindent\textbf{Supporting results.} Two results support these choices: the two log-likelihood proxies (\cref{tab:ll}) and the
robustness of the conclusions to the kernel bandwidth (\cref{fig:gamma}).

\begin{table*}[t]
\footnotesize\setlength{\tabcolsep}{3pt}
\centering
\caption{\textbf{Log-likelihood proxies of one-step samples.} In nats per point relative to a perfect model (0: as good as
real data; negative: nats lost). MeanFlow; mean over 2--3 seeds $\times$ 2 draws.}
\label{tab:ll}
\begin{tabular*}{\linewidth}{@{\extracolsep{\fill}}l cc cc cc}
\toprule
& \multicolumn{2}{c}{Pre-Pro ($\mathbb T^2$)} & \multicolumn{2}{c}{Fire ($S^2$)} & \multicolumn{2}{c}{Cone ($\SOthree$)} \\
\cmidrule(lr){2-3}\cmidrule(lr){4-5}\cmidrule(lr){6-7}
Prior & precision & coverage & precision & coverage & precision & coverage \\
\midrule
uniform                           & $-2.96$ & $-0.67$ & $-4.18$ & $-1.09$ & $-117.3$ & $-2.92$ \\
statistical, independent coupling & $-0.42$ & $-0.30$ & $-2.43$ & $-1.15$ & $-0.03$ & $-0.02$ \\
statistical, posterior coupling   & \best{$-0.07$} & \best{$\phantom{-}0.04$} & \best{$-0.23$} & \best{$-0.19$} & \best{$-0.07$} & \best{$-0.03$} \\
\bottomrule
\end{tabular*}
\end{table*}

\subsection{Positive definiteness of the evaluation kernels}\label{supp:kernels}
A kernel $k$ on $\M$ is \emph{positive definite} if $\sum_{a,b}c_ac_bk(x_a,x_b)\ge0$ for all finite sets of points
$x_a\in\M$ and coefficients $c_a\in\R$, and \emph{characteristic} if $\mathrm{MMD}_k(P,Q)=0$ implies $P=Q$ for all Borel
probability measures $P,Q$ on $\M$. The following lemma reduces both properties to the signs of expansion
coefficients.

\begin{lemma}\label{lem:char}
Let $\M$ be compact with a probability measure $\nu$ of full support, and let
$L^2(\M,\nu)=\bigoplus_{n\in\mathcal I}H_n$ be an orthogonal decomposition over a countable index set into
finite-dimensional subspaces of continuous functions, with $H_{n_0}$ the constants and $\bigoplus_nH_n$ dense in
$C(\M)$. For an orthonormal basis $(e_{n,j})_j$ of $H_n$, let $K_n(x,y)=\sum_je_{n,j}(x)e_{n,j}(y)$, and let
$k=\sum_nb_nK_n$ with $b_n\ge0$ and $\sum_nb_n\sup_xK_n(x,x)<\infty$. Then $k$ is continuous and positive definite, and
if $b_n>0$ for all $n\neq n_0$, it is characteristic.
\end{lemma}
\begin{proof}
By the Cauchy--Schwarz inequality, $|K_n(x,y)|\le\sup_xK_n(x,x)$, so the series defining $k$ converges absolutely and
uniformly, and $k$ is continuous. For a finite signed Borel measure $\mu$ on $\M$,
\begin{equation}\label{eq:kmu}
\begin{aligned}
  \iint k\,d\mu\,d\mu&\overset{\text{(a)}}{=}\sum_nb_n\iint K_n\,d\mu\,d\mu\\
  &\overset{\text{(b)}}{=}\sum_nb_n\sum_j\Bigl(\int e_{n,j}\,d\mu\Bigr)^2\ge0,
\end{aligned}
\end{equation}
where (a) exchanges sum and integrals by uniform convergence and (b) is the definition of $K_n$. With
$\mu=\sum_ac_a\delta_{x_a}$, \eqref{eq:kmu} is positive definiteness. With $\mu=P-Q$, the left-hand side of
\eqref{eq:kmu} equals $\mathrm{MMD}_k^2(P,Q)$. If it vanishes and $b_n>0$ for $n\neq n_0$, then
$\int e_{n,j}\,d\mu=0$ for all $n\neq n_0$ and all $j$, and also for the constants, since $\mu(\M)=0$. Hence
$\int f\,d\mu=0$ for every $f\in\bigoplus_nH_n$ and, by density, for every $f\in C(\M)$, so that $\mu=0$ by the Riesz
representation theorem (cf.\ \citealp{sriperumbudur2010hilbert,fukumizu2008characteristic}).
\end{proof}

\begin{proposition}\label{prop:kernels}
For every $\gamma>0$, the kernels \eqref{eq:kernels} on $\mathbb T^d$, $S^2$ and $\SOthree$ are positive definite and
characteristic.
\end{proposition}
\begin{proof}
\emph{Torus.} On the circle with the normalized Lebesgue measure, let $H_0$ be the constants and $H_n$ the span of
$\sqrt2\cos n\theta$ and $\sqrt2\sin n\theta$ for $n\ge1$; their union is dense in $C(\mathbb T)$ by the
Stone--Weierstrass theorem, and $K_n(\theta,\theta')=2\cos n(\theta-\theta')$ for $n\ge1$. The Fourier cosine
coefficients of $\theta\mapsto e^{-\gamma|\theta|}$ on $[-\pi,\pi]$ are
\[
\begin{aligned}
  b_n&=\frac1\pi\int_{-\pi}^{\pi}e^{-\gamma|\theta|}\cos n\theta\,d\theta
  =\frac2\pi\int_0^\pi e^{-\gamma\theta}\cos n\theta\,d\theta\\
  &=\frac{2\gamma}{\pi}\,\frac{1-(-1)^ne^{-\gamma\pi}}{\gamma^2+n^2}>0,\qquad n\ge0,
\end{aligned}
\]
so that $e^{-\gamma|\mathrm{wrap}(\theta-\theta')|}=\frac{b_0}{2}+\sum_{n\ge1}\frac{b_n}{2}K_n(\theta,\theta')$ with
positive and summable coefficients, $b_n=O(n^{-2})$. On $\mathbb T^d$, the product kernel expands in the products of these
one-dimensional bases, whose span is dense by the Stone--Weierstrass theorem, and its coefficients are products of the
positive one-dimensional coefficients. \Cref{lem:char} applies.

\emph{Sphere.} Let $t=\langle x,y\rangle\in[-1,1]$, let $P_n$ be the Legendre polynomial of degree $n$, and let $K_n$
be the reproducing kernel of the spherical harmonics of degree $n$ with respect to the normalized area measure. Then
\begin{align*}
  e^{-\gamma d_g(x,y)}&\overset{\text{(c)}}{=}e^{-\gamma\pi/2}e^{\gamma\arcsin t}
  \overset{\text{(d)}}{=}e^{-\gamma\pi/2}\sum_{m\ge0}a_mt^m\\
  &\overset{\text{(e)}}{=}e^{-\gamma\pi/2}\sum_{m\ge0}a_m\!\!\sum_{\substack{n\le m\\ n\equiv m\ (2)}}\!\!c_{m,n}P_n(t)\\
  &\overset{\text{(f)}}{=}\sum_{n\ge0}b_nP_n(t)
  \overset{\text{(g)}}{=}\sum_{n\ge0}\frac{b_n}{2n+1}K_n(x,y),
\end{align*}
where (c) uses $d_g(x,y)=\arccos t=\pi/2-\arcsin t$; (d) composes the Taylor series of $\arcsin$, whose coefficients are
positive, with that of $\exp$, so that $a_m>0$ for all $m$ and $\sum_ma_m=e^{\gamma\pi/2}<\infty$; (e) expands
$t^m$ in Legendre polynomials with positive coefficients $c_{m,n}$ \citep{atkinson2012spherical}; (f) exchanges the
sums, which is allowed because all terms are non-negative, and defines
$b_n=e^{-\gamma\pi/2}\sum_{m\ge n,\,m\equiv n\,(2)}a_mc_{m,n}>0$; and (g) is the addition theorem
$P_n(\langle x,y\rangle)=K_n(x,y)/(2n+1)$ \citep{atkinson2012spherical}. Since $K_n(x,x)=2n+1$, the summability condition
of \cref{lem:char} reads $\sum_nb_n=\sum_nb_nP_n(1)=e^{-\gamma d_g(x,x)}=1$, and \cref{lem:char} applies, with the
spherical harmonics dense in $C(S^2)$ by the Stone--Weierstrass theorem.

\emph{Rotation group.} The chordal Laplace kernel is the restriction to $\SOthree\subset\R^{3\times3}\cong\R^9$ of the
translation-invariant kernel $e^{-\gamma\|z-z'\|/\sqrt2}$ on $\R^9$. Its Fourier transform is positive everywhere, so by
Bochner's theorem it is positive definite, and it is characteristic for Borel probability measures on $\R^9$
\citep[Thm.~9]{sriperumbudur2010hilbert}. Probability measures on $\SOthree$ are probability measures on $\R^9$, so its
restriction to $\SOthree$ is positive definite and characteristic as well.
\end{proof}

\begin{aremark}[The geodesic Laplace kernel on $\SOthree$ is not positive definite]\label{rem:geolaplace}
Let $\theta(R,R')\in[0,\pi]$ be the rotation angle of $R^\top R'$ and $f(\theta)=e^{-\gamma\theta}$. As a class function,
$f$ has the Peter--Weyl expansion $\sum_\ell(2\ell+1)\hat f_\ell\chi_\ell$ in the characters
$\chi_\ell(\theta)=\sin((\ell+\frac12)\theta)/\sin(\theta/2)$ of the irreducible representations of dimension
$2\ell+1$, and the kernel $f(\theta(R,R'))$ is positive definite if and only if $\hat f_\ell\ge0$ for all $\ell$. By the
Weyl integration formula \citep{varadarajan2013lie},
\[
\begin{aligned}
  \hat f_\ell&\propto\int_0^\pi f(\theta)\,\chi_\ell(\theta)\,(1-\cos\theta)\,d\theta\\
  &\overset{\text{(h)}}{=}\int_0^\pi e^{-\gamma\theta}\bigl(\cos\ell\theta-\cos(\ell+1)\theta\bigr)\,d\theta
  =I_\ell-I_{\ell+1},
\end{aligned}
\]
with $I_m=\int_0^\pi e^{-\gamma\theta}\cos m\theta\,d\theta=\gamma\bigl(1-(-1)^me^{-\gamma\pi}\bigr)/(\gamma^2+m^2)$,
where (h) uses $\chi_\ell(\theta)(1-\cos\theta)=2\sin((\ell+\frac12)\theta)\sin(\theta/2)=\cos\ell\theta-\cos(\ell+1)\theta$.
For even $\ell$,
\begin{align*}
  I_\ell-I_{\ell+1}&=\gamma\Bigl[\frac{1}{\gamma^2+\ell^2}-\frac{1}{\gamma^2+(\ell+1)^2}\Bigr]\\
  &\quad-\gamma e^{-\gamma\pi}\Bigl[\frac{1}{\gamma^2+\ell^2}+\frac{1}{\gamma^2+(\ell+1)^2}\Bigr]\\
  &=O(\ell^{-3})-\frac{2\gamma e^{-\gamma\pi}}{\ell^2}\bigl(1+O(\ell^{-1})\bigr),
\end{align*}
which is negative for all sufficiently large even $\ell$, so the kernel is not positive definite for any $\gamma>0$.
For $\gamma=1$, the first negative coefficient occurs at $\ell=24$, and the threshold grows roughly like
$e^{\gamma\pi}$; sampled Gram matrices do not reveal these high-frequency violations. The chordal kernel agrees with the
geodesic Laplace kernel for small angles, since $\|R-R'\|_F/\sqrt2=2\sin(\theta/2)=\theta+O(\theta^3)$.
\end{aremark}

\noindent\textbf{Noise floor and repeated draws.} For synthetic data, we compare with $10^4$ independent samples from the
known density, and the noise floor is the standard deviation of MMD$^2$ between two independent samples of the same
size from the true distribution (five repetitions). For real data, we use the exact held-out split. The noise floor pools all real points (training, validation and test) and draws 20 random subsets of the size of the test split. For each subset, we compute the unbiased MMD$^2$ to up to $10^4$ of the remaining points; $\sigma_0$ is the standard deviation of these 20 values. The subsets are drawn
with a seed derived from the name of the dataset, so that each noise floor depends only on its dataset and is the same
in every evaluation. The Riemannian MeanFlow of \citet{zhong2026riemannian} is evaluated on 3--5 independent draws of
generated samples and the other frameworks on the samples saved by each run; we report means over seeds and draws. A perfect model would obtain values of mean zero and standard deviation about $\sigma_0$. Most differences we report exceed the separation threshold $3\sigma_0$ by one to two orders of magnitude.

\noindent\textbf{Log-likelihood proxies and memorization.} Following precision--recall analyzes of generative models \citep{sajjadi2018assessing,kynkaanniemi2019improved},
\emph{precision} is the mean log-density of generated samples under a kernel density estimate of the test split. \emph{Coverage} is the mean log-density of test points under a kernel density estimate of the generated samples. Both are relative to a perfect model. One-step samples from the uniform prior
lose 3--4 nats per point in precision on Pre-Pro and Fire and 117 nats on Cone. From the statistical prior, they lose 0.07--0.23 nats (\cref{tab:ll}). The memorization ratio is the mean geodesic distance from each generated point to its nearest training point,
divided by the same quantity for test points; the training set is subsampled to $5\times10^4$ points. Values near one are ideal. Values well below one indicate copying, and values well above one indicate samples outside the support. All distances are computed from coordinate differences in double
precision, $d(x,y)=2\arcsin(\|x-y\|/2)$ on $S^2$ and $\theta(R,R')=2\arcsin(\|R-R'\|_F/(2\sqrt2))$ on $\SOthree$, which
remain accurate for nearby points.

\noindent\textbf{Examination of the RMF protocol.} The noise floors in \cref{tab:audit} are computed with the RMF estimator
(the square root of a V-statistic with the geodesic Gaussian kernel, $\gamma=1$) on 20 random disjoint splits of the
real data. The lowest baseline values are the minima over the four baselines RFM, G-LSD, RMF and RMF-MT. On the torus, the joint kernel in the shuffling control uses the geodesic distance on
$\mathbb T^d$. \Cref{tab:audit} summarizes this examination.

\noindent\textbf{Geodesics on $\SOthree$.} For all $\SOthree$ models, including the uniform baseline, geodesics and
their time derivatives are computed in closed form, $x_t=x_0\Exp(t\Log(x_0^\top x_1))$ and
$\dot x_t=x_t\Log(x_0^\top x_1)$, with the matrix exponential and logarithm evaluated by Rodrigues' formula.

\section{Datasets}\label{supp:data}
The datasets are public measurements or are generated from closed-form densities. \Cref{tab:datasets} lists them with their use; their construction follows.

\begin{table*}[t]
\footnotesize\setlength{\tabcolsep}{3pt}
\centering
\caption{\textbf{Datasets.} Real data are measurements; synthetic data have known densities and serve controlled analyzes.
Details below.}
\label{tab:datasets}
\begin{tabular*}{\linewidth}{@{\extracolsep{\fill}}l l l r l}
\toprule
Data & Manifold & Type & Size & Use \\
\midrule
Volcano, Earthquake, Flood, Fire & $S^2$ & real & 827--12{,}809 & main comparison, evaluation \\
General, Glycine, Proline, Pre-Pro & $\mathbb T^2$ & real & 6910--138{,}208 & main comparison, evaluation \\
RNA backbone & $\mathbb T^7$ & real & 9478 & main comparison \\
Torsion windows ($k=2,3,5$) & $\mathbb T^{4}$--$\mathbb T^{10}$ & real & 32{,}756--82{,}711 & dimension \\
Backbone fragments ($k=10,20$) & $\mathbb T^{20}$, $\mathbb T^{40}$ & real & 7795--16{,}111 & dimension, physical statistics \\
Side chains $(\chi_1,\chi_2)$, 14 residue types & $\mathbb T^2$ & real & 14 types & conditional, rotamer validation \\
\midrule
Cone, Fisher, Line, Swiss roll & $\SOthree$ & synthetic & $4\times10^4$ & standard benchmarks \\
vMF mixtures, rings, arcs & $S^2$ & synthetic & $10^4$ & concentration, shape, ablations \\
Banana mixtures & $\mathbb T^{2}$--$\mathbb T^{50}$ & synthetic & $2\times10^4$ & dimension \\
\bottomrule
\end{tabular*}
\end{table*}

\noindent\textbf{Synthetic sphere data.} The concentration experiments use mixtures of five von Mises--Fisher distributions
with Dirichlet(2) weights and a common concentration $\kappa\in\{5,25,100,250\}$. The rings consist of three small
circles with opening angles of $0.5$, $0.9$ and $1.3$ rad and a Gaussian width of $30$ or $100$ mrad, and the arcs consist
of four bands along great circles with along-arc standard deviations of $0.35$--$0.6$ rad and a width of $30$ or $100$
mrad. Each dataset contains $10^4$ points, and all densities are available in closed form and integrate to one.

\noindent\textbf{Side chains.} The PDB entries of the Top500 chains were downloaded from RCSB. We computed $(\chi_1,\chi_2)$ with the standard atom definitions for the 14 residue types that have both angles, using only atoms without alternate locations (or the first) with full occupancy and B-factors below $40$\,\AA$^2$. Our implementation agrees with Biopython to three decimals. Each type is split 80/10/10 with seed 42, and smaller types are repeated in training, up
to 30 times, toward balance; models are trained for 200 epochs. Validation contours are computed from the training
split on a $2^\circ$ grid smoothed with a Gaussian of $3^\circ$.

\noindent\textbf{Earth data.} The Earth datasets comprise volcanic eruptions (827 events), earthquakes (6120), floods (4875)
and wildfires (12{,}809), as in \citet{mathieu2020riemannian}, and are split 80/10/10 with a fixed seed.

\noindent\textbf{Synthetic torus data of increasing dimension.} For $d\in\{2,5,10,20,50\}$, each dataset is a mixture of
eight components with Dirichlet(2) weights and centers $c_k$ drawn uniformly on $\mathbb T^d$. Component $k$ draws
$v\sim\mathcal N(0,\Sigma_k)$ with $\Sigma_k=Q_k\,\mathrm{diag}(\sigma_{k,1}^2,\dots,\sigma_{k,d}^2)\,Q_k^\top$, where $Q_k$ is a uniformly random rotation and $\sigma_{k,i}$ is uniform on $[0.15,0.45]$ rad. It then bends $v$ into
\[
  y_1=v_1,\qquad y_i=v_i+0.6\sin v_{i-1}\quad(i=2,\dots,d).
\]
The Jacobian matrix of $v\mapsto y$ is lower triangular with unit diagonal, so the bending preserves volume and the
density of $y$ is that of $v$ evaluated at the inverse map. Samples with $|y_i|\ge\pi$ for some $i$ are rejected, and
the point $(c_k+y)\bmod2\pi$ is returned. The density is therefore exact up to the truncation constant, which is
estimated by Monte Carlo; it integrates to one ($d=2,3$) and agrees with the samples in the expectations of test
functions. Each dataset contains $2\times10^4$ points and is split 80/10/10.

\noindent\textbf{Torsion windows and fragments.} From the backbone torsion angles of the Top500 set, we extract runs of consecutive residue numbers within a chain and cut them into non-overlapping windows of $k$ residues. No residue appears in two windows, so no residue is shared between the random training and test splits. This procedure yields 82\,711, 54\,962, 32\,756, 16\,111 and 7\,795 windows for $k=2,3,5,10,20$, respectively,
on $\mathbb T^{2k}$, split 80/10/10. For $k=10$ and $k=20$, we reconstruct backbones from $(\phi,\psi)$ with the NeRF
algorithm, ideal bond lengths and angles, and $\omega=180^\circ$. The reconstruction reproduces the input torsion angles
exactly and yields the standard geometry of ideal $\alpha$-helices and $\beta$-strands. Real and generated fragments are
reconstructed in the same way.

\noindent\textbf{Torus data.} The protein datasets contain the backbone dihedral angles of the Top500 proteins
\citep{lovell2003structure}, grouped by residue type: General (138{,}208 residues), Glycine (13{,}283), Proline (7634)
and Pre-Pro (6910). The RNA dataset contains backbone torsion angles (9478 points with seven angles each;
\citealp{murray2003rna}). All datasets are split 80/10/10 with a fixed seed.

\noindent\textbf{$\SOthree$ data.} We use Cone, Fisher and Line as released by \citet{zhong2026riemannian}, together with the Swiss
roll, which is generated as the minimal rotations that map $e_y$ to points on a noisy spiral (noise $0.01$). We use
$4\times10^4$ samples of each dataset, shuffled with the random seed of the experiment and split 80/10/10.

\section{Detailed results}\label{supp:results}
% Results follow the five questions of \mref{sec:experiments}; the conventions are those of \cref{supp:notation}.

Results follow \mref{sec:experiments}; the conventions are those of \cref{supp:notation}.

% \subsection{Uniform priors impose harder one-step maps}\label{supp:res-q1}

% \subsection{Why uniform priors fail: a controlled study}\label{supp:res-q1}
\subsection{A Controlled Study of the Uniform Prior}\label{supp:res-q1}

\mref{fig:trends}a reports the log-likelihood lost; \cref{tab:conc} gives the MMD at two bandwidths. At $\kappa=100$, the MMD$^2$ at $\gamma=10$ is 27 times
smaller from the statistical prior, while the RMF kernel on 1000 test points stays at its noise floor on all four datasets
and misses the difference. \Cref{tab:shapes} gives the values for thin rings and long arcs.

\begin{figure*}[t]
\centering
\includegraphics[width=\textwidth]{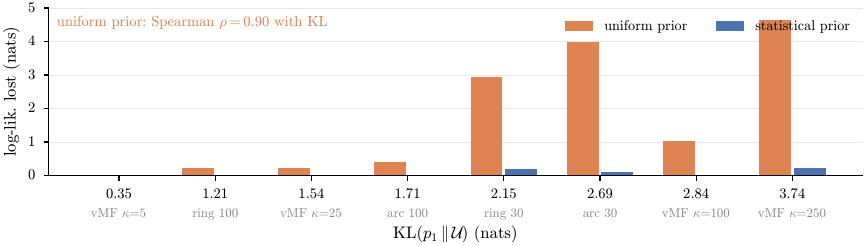}
\caption{\textbf{Divergence and harm} on the eight synthetic datasets on $S^2$, ordered by $\mathrm{KL}(p_1\,\|\,\mathcal U)$:
log-likelihood lost by one-step samples. Ring and arc widths in mrad.}
\label{fig:klcorr}
\end{figure*}

\noindent\textbf{The divergence predicts the harm.} \Cref{fig:klcorr} shows, for the eight synthetic datasets on $S^2$, the log-likelihood lost by one-step samples against the divergence of the data from uniformity. The loss of the uniform prior
increases with the divergence (Spearman $\rho=0.90$, $p=0.002$), as \mref{prop:jacobian} predicts; the loss of the
statistical prior stays below 0.24 nats and is not correlated with it ($\rho=0.45$, $p=0.26$).

\begin{table*}[t]
\footnotesize\setlength{\tabcolsep}{3pt}
\centering
\caption{\textbf{Controlled experiments on $S^2$.} (a) Data concentration (vMF prior); LL: mean true log-likelihood of
the samples, with the target in the last column. (b) Rings and arcs; LL gap: true minus sample log-likelihood. Floors:
0.005--0.024 ($\gamma=10$) and 0.001--0.006 ($\gamma=50$).}
\label{tab:conc}
\label{tab:shapes}
\noindent{\textbf{(a)} Data concentration}\par\smallskip
{\begin{tabular*}{\linewidth}{@{\extracolsep{\fill}}lccccccc}
\toprule
& \multicolumn{2}{c}{Laplace $\gamma=10$} & \multicolumn{2}{c}{Laplace $\gamma=50$} & \multicolumn{2}{c}{LL} & \\
\cmidrule(lr){2-3}\cmidrule(lr){4-5}\cmidrule(lr){6-7}
$\kappa$ & Uniform & vMF & Uniform & vMF & Uniform & vMF & LL (true) \\
\midrule
5   & \best{0.112} & 0.531 & \best{0.008} & 0.093 & $-2.187$ & \best{$-2.180$} & $-2.165$ \\
25  & 1.323 & \best{0.286} & 0.340 & \best{0.130} & $-1.160$ & \best{$-0.998$} & $-0.999$ \\
100 & 4.825 & \best{0.178} & 2.691 & \best{0.092} & $-0.742$ & \best{0.312} & 0.301 \\
250 & 13.07 & \best{1.017} & 7.278 & \best{0.359} & $-3.397$ & \best{0.984} & 1.208 \\
\bottomrule
\end{tabular*}}
\par\medskip\noindent{\textbf{(b)} Rings and arcs}\par\smallskip
{\begin{tabular*}{\linewidth}{@{\extracolsep{\fill}}lccccccccc}
\toprule
& \multicolumn{3}{c}{Laplace $\gamma=10$} & \multicolumn{3}{c}{Laplace $\gamma=50$} & \multicolumn{3}{c}{LL gap} \\
\cmidrule(lr){2-4}\cmidrule(lr){5-7}\cmidrule(lr){8-10}
Data & Unif. & vMF & Stat. & Unif. & vMF & Stat. & Unif. & vMF & Stat. \\
\midrule
Ring, 100 mrad & \pmcell{0.46}{0.04} & \pmcell{0.26}{0.02} & \best{\pmcell{0.12}{0.03}} & \pmcell{0.146}{0.060} & \pmcell{0.086}{0.005} & \best{\pmcell{0.027}{0.007}} & 0.196 & $-0.021$ & \best{0.008} \\
Ring, 30 mrad & \pmcell{1.29}{0.17} & \pmcell{0.88}{0.28} & \best{\pmcell{0.27}{0.04}} & \pmcell{0.671}{0.073} & \pmcell{0.596}{0.126} & \best{\pmcell{0.181}{0.021}} & 3.123 & 0.682 & \best{0.204} \\
Arc, 100 mrad & \pmcell{1.10}{0.12} & \pmcell{0.27}{0.10} & \best{\pmcell{0.10}{0.03}} & \pmcell{0.375}{0.063} & \pmcell{0.113}{0.032} & \best{\pmcell{0.020}{0.007}} & 0.402 & 0.006 & \best{0.000} \\
Arc, 30 mrad & \pmcell{4.49}{0.80} & \pmcell{0.69}{0.06} & \best{\pmcell{0.36}{0.05}} & \pmcell{1.753}{0.236} & \pmcell{0.624}{0.057} & \best{\pmcell{0.272}{0.023}} & 4.066 & 0.526 & \best{0.138} \\
\bottomrule
\end{tabular*}}
\end{table*}

\begin{figure*}[t]
\vspace{-1em}
\centering
\includegraphics[width=\textwidth]{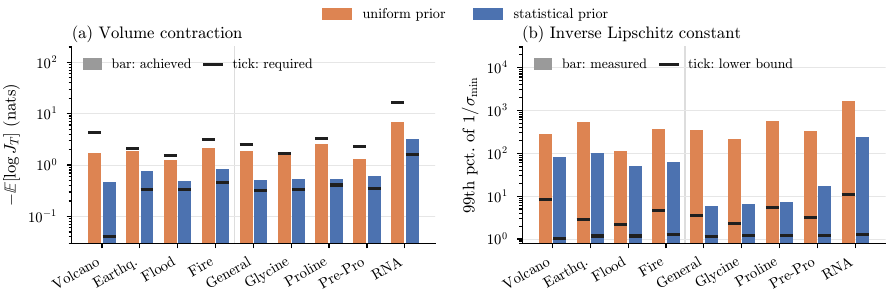}
\caption{\textbf{Geometry of the learned one-step maps} on 1000 prior samples. (a) Achieved contraction (bars) and the
contraction an injective map requires (ticks). (b) Inverse stretch (bars) and the lower bound on $\Lip(T^{-1})$ (ticks).}
\label{fig:geometry}
\vspace{-1em}
\end{figure*}

\noindent\textbf{Geometry of the learned maps.} \Cref{fig:geometry} compares four quantities, and \cref{tab:geometry-full} reports two more on all nine datasets:
\begin{itemize}[nosep,leftmargin=*]
\item \emph{achieved contraction}, $-\E_{p_0}[\log J_T]$;
\item \emph{required contraction}, $H(p_0)-\hat H$ with $\hat H=-\E_{\mathrm{test}}[\log q]$ for the fitted prior density
$q$, an estimate of the contraction that an injective map needs (\mref{prop:jacobian});
\item \emph{inverse stretch}, the 99th percentile of $1/\sigma_{\min}(dT_x)$;
\item \emph{lower bound}, $\exp(\text{required}/d)$, the matching estimate of the lower bound on $\Lip(T^{-1})$;
\item \emph{folded}, the percentage of samples at which $dT_x$ reverses orientation;
\item \emph{displacement}, the mean geodesic distance between $z$ and $T(z)$, in radians.
\end{itemize} Besides the folded
fraction, the achieved contraction increases with the required contraction across the nine datasets (Spearman
$\rho=0.77$, exact permutation $p=0.021$), and so does the inverse stretch, without reaching significance ($\rho=0.65$,
$p=0.067$). From the uniform prior, the achieved contraction lies below the bound for injective maps on eight of nine
datasets; from the statistical prior, the maps still fold on 8--29\% of the samples on the Earth data. The Lipschitz bound holds in all cases.

\begin{table}[t]
\centering\footnotesize\setlength{\tabcolsep}{3.2pt}
\caption{\textbf{All nine datasets. How trained maps meet the demand.} Required contraction (nats), folded prior samples (\%) and
displacement (rad). U, S: uniform, statistical prior.}
\label{tab:geometry-full}
\begin{tabular*}{\linewidth}{@{\extracolsep{\fill}}l cc cc cc@{}}
\toprule
& \multicolumn{2}{c}{required} & \multicolumn{2}{c}{folded (\%)} & \multicolumn{2}{c}{displacement} \\
\cmidrule(lr){2-3}\cmidrule(lr){4-5}\cmidrule(lr){6-7}
Dataset & U & S & U & S & U & S \\
\midrule
Volcano & 4.25 & 0.04 & 33.6 & 28.9 & 0.66 & 0.05 \\
Earthquake & 2.10 & 0.33 & 18.4 & 14.2 & 0.64 & 0.06 \\
Flood & 1.55 & 0.33 & 15.5 & 8.6 & 0.57 & 0.06 \\
Fire & 3.10 & 0.46 & 22.3 & 8.2 & 0.76 & 0.03 \\
\midrule
General & 2.54 & 0.32 & 28.9 & 0.2 & 1.38 & 0.09 \\
Glycine & 1.66 & 0.34 & 18.1 & 0.5 & 0.79 & 0.13 \\
Proline & 3.35 & 0.41 & 44.9 & 0.3 & 1.76 & 0.09 \\
Pre-Pro & 2.34 & 0.35 & 37.8 & 1.2 & 1.53 & 0.10 \\
RNA ($\mathbb T^7$) & 16.84 & 1.63 & 51.0 & 13.0 & 3.92 & 0.30 \\
\bottomrule
\end{tabular*}
\end{table}

% \subsection{Samples stay where the data occur}\label{supp:res-q2}
% \subsection{Structural biology and Earth science: where samples land}\label{supp:res-q2}

\subsection{Structural Biology and Earth Science}\label{supp:res-q2}

\begin{figure}[t]
\centering
\includegraphics[width=\linewidth]{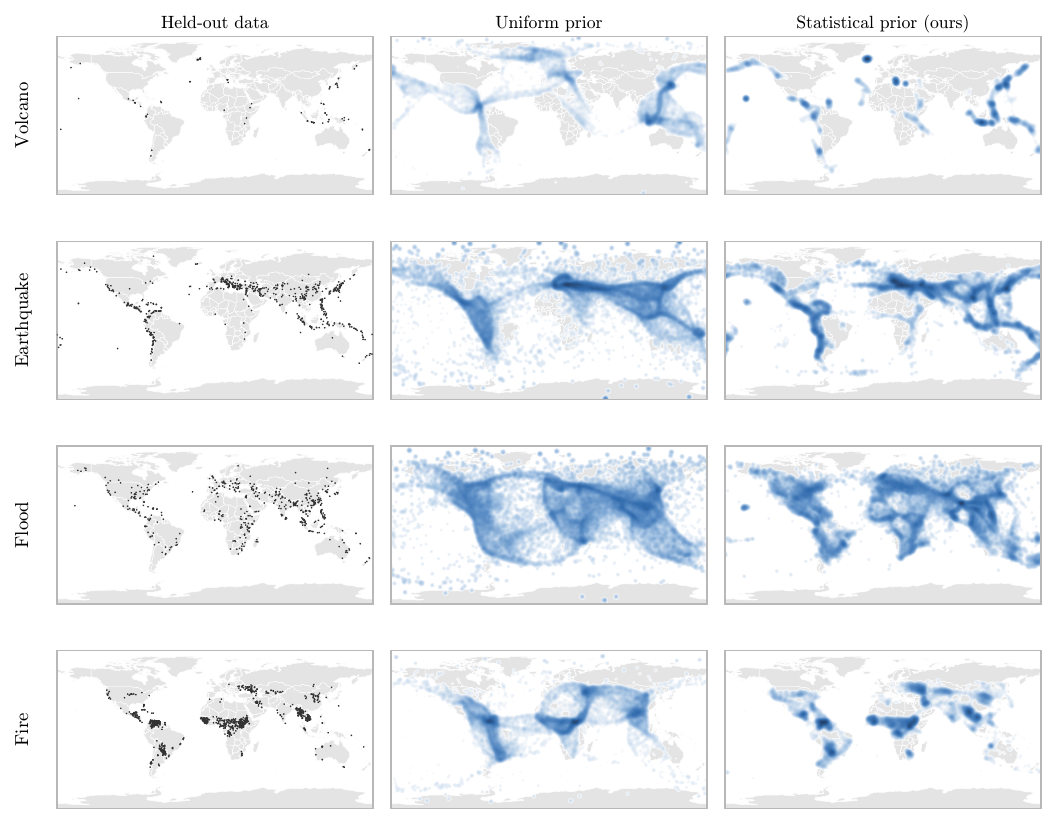}
\caption{\textbf{One-step samples on the four Earth datasets.} From the uniform prior, samples spread away from the seismic
belts across the open ocean; from the statistical prior, they follow the data.}
\label{fig:earth}
\vspace{-1em}
\end{figure}

\begin{figure}[t]
\centering
\includegraphics[width=0.82\linewidth, trim=0 144 0 0, clip]{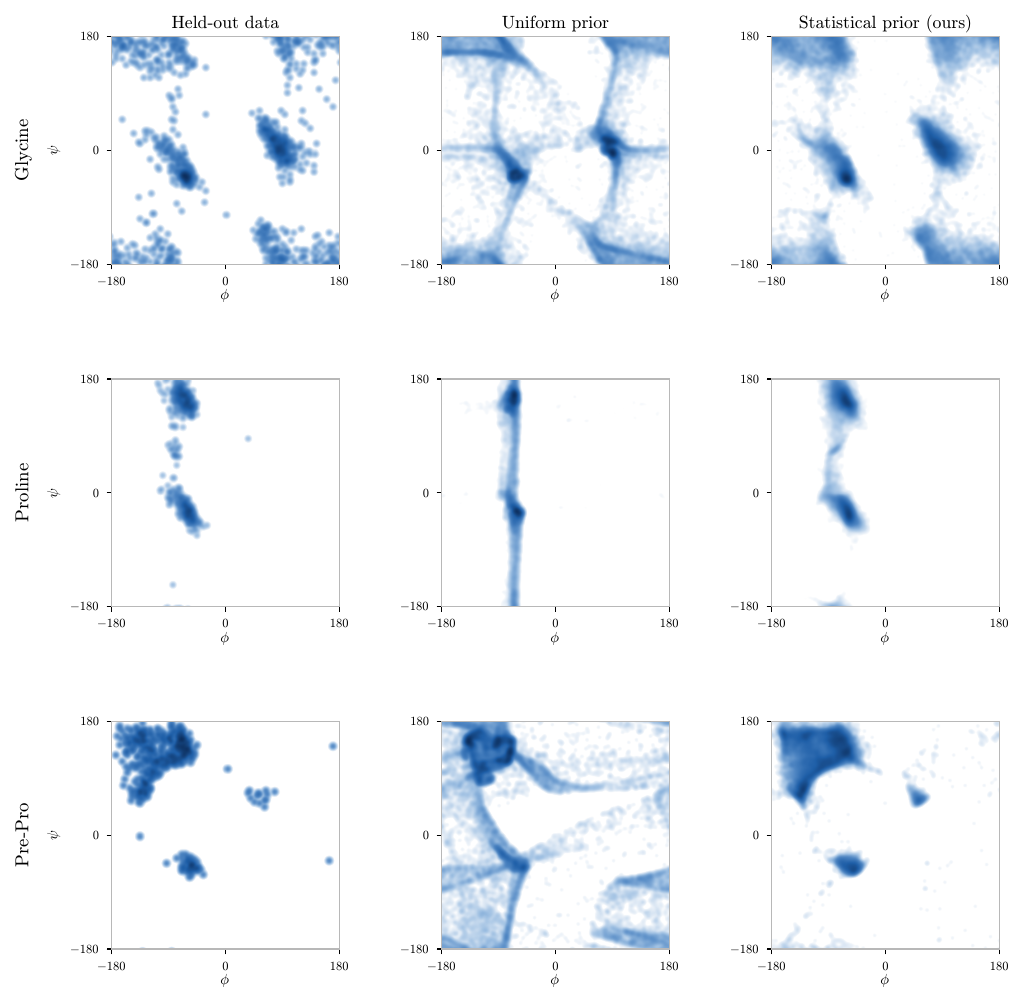}
\vspace{-1.4em}
\caption{\textbf{Backbone angles of Glycine and Proline} (Pre-Pro in \mref{fig:teaser}). Samples from the uniform
prior cross forbidden regions; ours match the data.}
\label{fig:rama}
\vspace{-1em}
\end{figure}

\Cref{fig:earth} shows one-step samples on the four Earth datasets, and \cref{fig:rama} the backbone angles of all
residue types.

\noindent\textbf{Downstream application: sequence-conditioned fragment sampling.}
Fragment libraries, which are used in loop modeling and structure prediction, contain backbone conformations sampled
for given sequence segments. To test the conditional model in this setting, we take 5000 segments of eight residues from the Top500 chains and keep only their residue types. We sample $(\phi,\psi)$ at each position from the conditional model of \cref{sec:cond} and rebuild the backbone with ideal geometry. We follow MolProbity-style validation. A residue is a Ramachandran outlier if it lies outside the 99.95\% density contour of the training data of its type, and favored if it lies inside the 98\% contour. A fragment is usable if it has neither an outlier nor a severe steric clash. Independent draws of held-out angles for the same residue types serve as a reference, since they
bound what any per-residue sampler can achieve.

\Cref{tab:downstream} adds the favored residues, clashes and secondary structure to \mref{fig:validation}. The statistical
prior reaches 98\% of the usable fraction of independent held-out draws, their clash rate, and the helix and sheet
fractions of real segments. All per-residue samplers have more clashes than real segments: they ignore the dependence between neighboring residues, which the hidden-Markov prior of \cref{supp:hmm} models.

\noindent\textbf{Side-chain rotamers.}
Rotamer libraries supply the side-chain dihedral angles from which
full-atom protein models are built. Continuous statistical models of these angles exist \citep{harder2010beyond}, yet neural generators of side chains usually start from the uniform distribution on the torus \citep{zhang2024generalized}. We computed $(\chi_1,\chi_2)$ for the 14
residue types that have both angles in the Top500 chains (\cref{supp:data}) and trained one model conditioned on the
residue type, as in \cref{sec:cond}. Side chains are an instructive test case because they are strongly concentrated:
each type occupies a few rotamer wells, and proline, whose side chain closes a ring with the backbone, occupies a single
narrow band. As before, we count rotamer outliers, i.e.\ samples outside the 99.7\% density contour of the training data
of their type, and samples in the favored region (98\% contour), with held-out data of the same types as reference.

The favored region of each residue type is in \cref{tab:sidechain}. Under the RMF kernel, samples from the statistical prior are within the noise
floor on average (0.35 noise-floor standard deviations); those from the uniform prior are 93 standard deviations away. The densities show why (\mref{fig:rotamers}). The uniform prior produces bands across the torus. The statistical prior reproduces the rotamer wells, including the six wells of phenylalanine, whose $\chi_2$ is
symmetric under rotation by $180^\circ$.

\begin{table*}[t]
\footnotesize\setlength{\tabcolsep}{3pt}
\centering
\caption{\textbf{Protein applications}: what \mref{fig:validation} does not show (percentages, two seeds). (a) Fragments of
eight residues for 5000 real sequences: favored residues, clashes per fragment, helix and sheet. (b) Side chains of 14
residue types: samples in the favored region (98\% contour). n/a: not applicable.}
\label{tab:downstream}
\label{tab:sidechain}
\noindent{\textbf{(a)} Sequence-conditioned fragment sampling}\par\smallskip
{\begin{tabular*}{\linewidth}{@{\extracolsep{\fill}}l c cccc}
\toprule
Method & steps & favored & clashes & helix & sheet  \\
\midrule
Uniform prior     & 1 & 74.5 & 5.1 & 45.2 & 35.1  \\
& 4 & 89.6 & 3.8 & 49.3 & 36.3  \\
Statistical prior & 1 & 96.7 & 3.0 & 50.1 & 40.7  \\
& 4 & \best{97.4} & \best{2.7} & 50.4 & 40.9  \\
\midrule
\emph{Independent held-out draws} & n/a & \emph{97.7} & \emph{3.1} & \emph{49.9} & \emph{41.3}  \\
\emph{Real segments}          & n/a & \emph{98.4} & \emph{0.5} & \emph{50.4} & \emph{41.1}  \\
\bottomrule
\end{tabular*}}
\par\medskip\noindent{\textbf{(b)} Side-chain rotamers in the favored region}\par\smallskip
{\begin{tabular*}{\linewidth}{@{\extracolsep{\fill}}l ccc}
\toprule
Residue & uniform prior & statistical prior & \emph{held-out data} \\
\midrule
All 14 types & 46.9 & \best{93.1} & \emph{96.3}  \\
\midrule
Proline       & 10.5 & \best{96.7} & \emph{98.4}  \\
Isoleucine    & 31.4 & \best{95.8} & \emph{97.1}  \\
Leucine       & 32.4 & \best{94.1} & \emph{97.1}  \\
Phenylalanine & 52.2 & \best{92.4} & \emph{97.8}  \\
Tryptophan    & 38.5 & \best{90.0} & \emph{94.8}  \\
Histidine     & 54.6 & \best{90.3} & \emph{96.5}  \\
\bottomrule
\end{tabular*}}
\end{table*}

\begin{figure}[t]
\centering
\includegraphics[width=\linewidth]{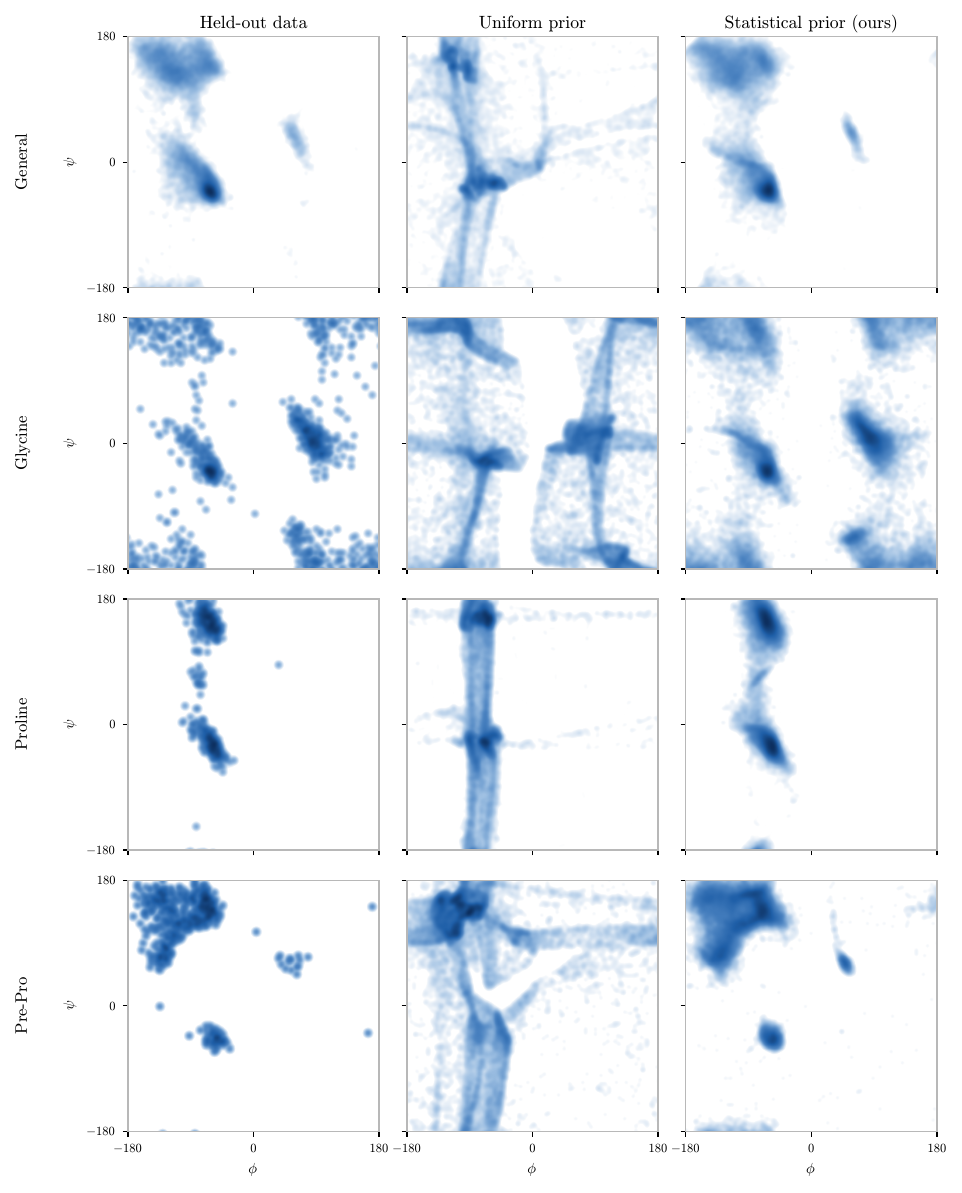}
\caption{\textbf{Ramachandran densities from a single model conditioned on the residue type} (held-out data and one-step
samples, seed 0).}
\label{fig:cond}
\vspace{-1em}
\end{figure}

% \begin{figure*}[t]
% \centering
% \begin{minipage}[b]{0.558\textwidth}\centering
% \includegraphics[width=\linewidth]{figs/fig_sidechain.pdf}\\[1pt]
% {\small (a) Side-chain rotamers $(\chi_1,\chi_2)$}
% \end{minipage}\hfill
% \begin{minipage}[b]{0.422\textwidth}\centering
% \includegraphics[width=\linewidth]{figs/fig_cond.pdf}\\[1pt]
% {\small (b) Backbone angles, one conditional model}
% \end{minipage}
% \caption{\textbf{Densities of held-out data and of one-step samples} (seed 0). (a) Side-chain rotamers of proline, leucine and
% phenylalanine. (b) Ramachandran densities from a single model conditioned on the residue type.}
% \label{fig:densities}\label{fig:sidechain}\label{fig:cond}
% \end{figure*}

% \subsection{Across one-step generators and sampling budgets}
% \subsection{Across generators and sampling budgets}
% \subsection{One prior for every one-step generator}

\subsection{Comparison with Existing Generators}
\label{supp:frameworks}
\begin{figure*}[t]
\centering
\includegraphics[width=\textwidth]{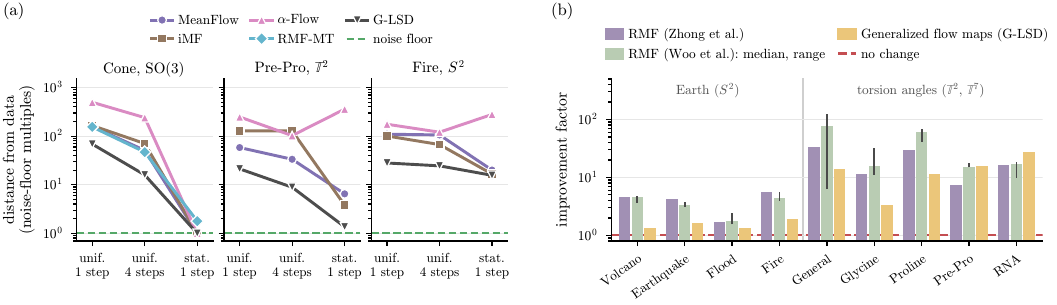}
\caption{\textbf{Switching the prior helps across objectives and generators}, at the finest scale. (a) Four MeanFlow
objectives and G-LSD, from one and four steps of the uniform prior to one step of ours. (b) Improvement factor (error
from the uniform prior over error from ours) of three one-step generators on nine real datasets; for
\citet{woo2026riemannian}, median and range over three formulations.}
\label{fig:objectives}\label{fig:generality}
\end{figure*}

\begin{figure*}[t]
\centering
\includegraphics[width=\textwidth]{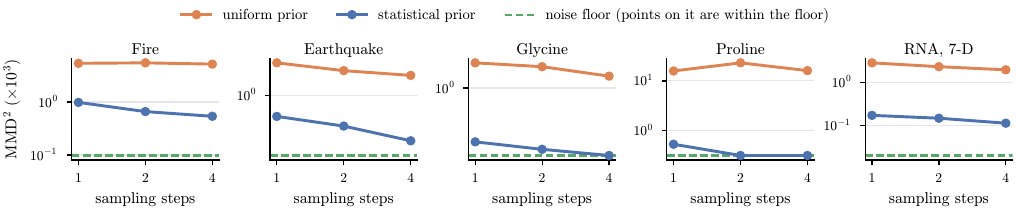}
\caption{\textbf{One step from the statistical prior beats four steps from the uniform prior} on five real datasets
(Riemannian MeanFlow of \citet{zhong2026riemannian}; the three $\SOthree$ benchmarks in \cref{fig:steps-so3}).}
\label{fig:steps}
\end{figure*}

\begin{figure*}[t]
\centering
\includegraphics[width=0.72\textwidth]{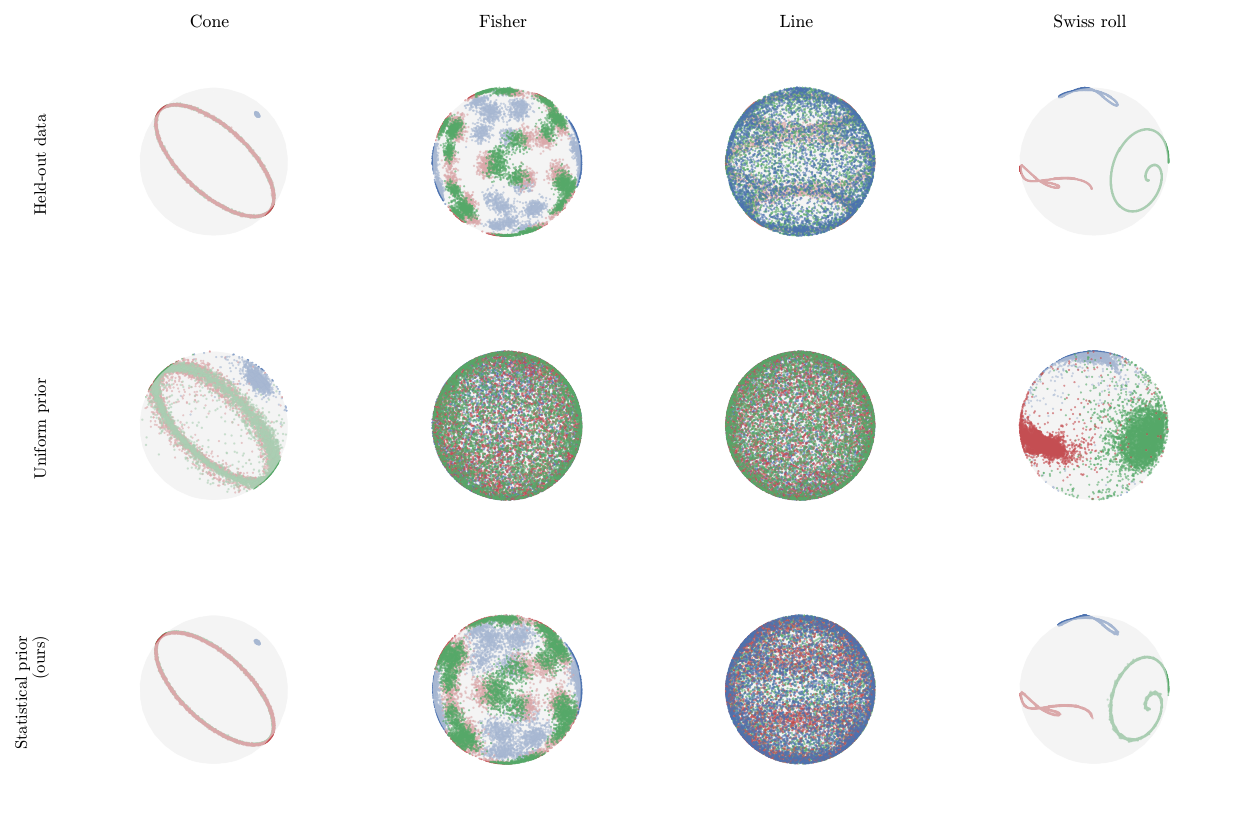}
\caption{\textbf{One-step samples on $\SOthree$}, each rotation drawn as its three axes. Samples from the uniform prior
spread or collapse; ours reproduce the data.}
\label{fig:so3}
\end{figure*}

\Cref{fig:objectives} compares training objectives and generators, \cref{fig:steps} the number of sampling steps on
five real datasets, and \cref{fig:so3} one-step samples on the $\SOthree$ benchmarks.

\Cref{tab:objectives} adds the RMF kernel to \cref{fig:objectives}. That figure divides all objectives by the same noise
floor per dataset, whereas \mref{tab:frameworks} and \cref{tab:frameworks-full} use the noise floors of the final
evaluation, drawn with a fixed seed per dataset (\cref{supp:evaldetails}). For G-LSD on Cone, Pre-Pro and Fire, the raw
MMD$^2$ of the two evaluations differ by at most 8\%, through independent draws of samples; the noise-floor multiples
differ only through the normalization. For $\alpha$-Flow (\cref{tab:aflow}), the
candidate settings (training epochs, curriculum length) are (200, 100) and (200, 200) on Cone; (1000, 100), (1000, 500)
and (1000, 1000) on Pre-Pro; and (100, 50), (100, 100), (400, 100), (400, 200) and (400, 400) on Fire. The curriculum anneals $\alpha$ from 1 to 0 along a sigmoid centered at half of its length. Each prior selects its setting by the finest-scale MMD$^2$ of one-step samples on the validation split. Settings that coincide with the default use two seeds, the others one.

\begin{table*}[t]
\footnotesize\setlength{\tabcolsep}{3pt}
\centering
\caption{\textbf{Other one-step objectives}: what \cref{fig:objectives} does not show. (a) The RMF kernel (noise-floor
multiples in parentheses); $^\ast$no one-step map was learned. (b) $\alpha$-Flow tuned to the same extent for both priors.}
\label{tab:objectives}
\label{tab:aflow}
\noindent{\textbf{(a)} RMF kernel, one step}\par\smallskip
{\begin{tabular*}{\linewidth}{@{\extracolsep{\fill}}ll ccc}
\toprule
Objective & Prior & Cone ($\SOthree$) & Fire ($S^2$) & Pre-Pro ($\mathbb T^2$) \\
\midrule
MeanFlow & uniform & 2.17 (14) & 0.659 (1.6) & 13.0 (83)  \\
& statistical & \best{$-0.043$ ($\le$1)} & \best{$-0.221$ ($\le$1)} & \best{$-0.079$ ($\le$1)}  \\
\midrule
iMF & uniform & 3.68 (24) & 0.492 (1.2) & 318 (2027)$^\ast$  \\
& statistical & \best{$-0.109$ ($\le$1)} & \best{$-0.279$ ($\le$1)} & \best{$-0.031$ ($\le$1)}  \\
\midrule
$\alpha$-Flow & uniform & 240 (1539)$^\ast$ & 131 (324) & 26.6 (169)  \\
& statistical & \best{$-0.064$ ($\le$1)} & \best{$-0.177$ ($\le$1)} & \best{11.6 (74)}  \\
\midrule
G-LSD & uniform & 2.19 (14) & \best{$-0.236$ ($\le$1)} & 1.55 (9.9)  \\
& statistical & \best{$-0.015$ ($\le$1)} & \best{$-0.223$ ($\le$1)} & \best{$-0.127$ ($\le$1)}  \\
\bottomrule
\end{tabular*}}
\par\medskip\noindent{\textbf{(b)} $\alpha$-Flow tuned to the same extent for both priors}\par\smallskip
{\begin{tabular*}{\linewidth}{@{\extracolsep{\fill}}ll c ccc}
\toprule
& & & RMF kernel & \multicolumn{2}{c}{finest scale} \\
\cmidrule(lr){4-4}\cmidrule(lr){5-6}
Dataset & Prior & Selected setting & 1 step & 1 step & 4 steps \\
\midrule
Cone    & uniform     & (200, 100)   & 240 (1539)        & 14.2 (497)        & 6.85 (240) \\
        & statistical & (200, 100)   & \best{$-0.064$ ($\le$1)} & \best{0.002 ($\le$1)} & \best{$-0.004$ ($\le$1)} \\
\midrule
Pre-Pro & uniform     & (1000, 1000) & 26.2 (167)        & \best{25.3 (241)} & 10.7 (102) \\
        & statistical & (1000, 1000) & \best{11.5 (73)}  & 33.7 (321)        & \best{1.13 (11)} \\
\midrule
Fire    & uniform     & (100, 50)    & 154 (383)         & \best{8.55 (174)} & 6.34 (129) \\
        & statistical & (100, 100)   & \best{$-0.126$ ($\le$1)} & 13.6 (277) & \best{1.60 (33)} \\
\bottomrule
\end{tabular*}}
\end{table*}

\begin{figure*}[t]
\centering
\includegraphics[width=\textwidth]{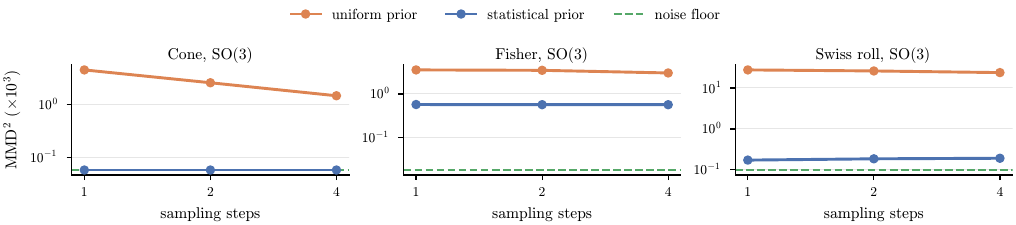}
\caption{\textbf{Number of sampling steps on the $\SOthree$ benchmarks}, which \cref{fig:steps} does not show (Riemannian
MeanFlow of \citet{zhong2026riemannian}, unbiased MMD$^2$ at the finest scale; values within the noise floor drawn at the floor).}
\label{fig:steps-so3}
\end{figure*}

\noindent\textbf{Number of sampling steps.} The three $\SOthree$ benchmarks (\cref{fig:steps-so3}) behave like the real datasets of \cref{fig:steps}: one step
from the statistical prior is closer to the data than four steps from the uniform prior.

\noindent\textbf{Other frameworks.} \Cref{tab:frameworks-full} adds the RMF kernel and two and four steps to \mref{tab:frameworks}. For all three
formulations of \citet{woo2026riemannian}, the statistical prior stays closer to the data with two and four steps, on all nine
datasets.  For generalized flow maps (G-LSD), the statistical prior also has the lower error with four steps on all nine datasets;
under the RMF kernel, the two priors are separated on General, Pre-Pro and RNA.

\begin{table*}[t]
\centering\footnotesize\setlength{\tabcolsep}{3pt}\renewcommand{\arraystretch}{0.96}
\caption{\textbf{The other one-step frameworks}: the RMF kernel and two and four steps, which \mref{tab:frameworks} does not show (noise-floor multiples, two seeds). G-LSD was run with one and four steps. U: uniform; S: statistical prior.}
\label{tab:frameworks-full}\label{tab:woo-full}
\begin{tabular*}{\linewidth}{@{\extracolsep{\fill}}ll cc cc cc}
\toprule
 & & \multicolumn{2}{c}{RMF kernel, 1 step} & \multicolumn{2}{c}{finest, 2 steps} & \multicolumn{2}{c}{finest, 4 steps} \\
\cmidrule(lr){3-4}\cmidrule(lr){5-6}\cmidrule(lr){7-8}
Dataset & Generator & U & S & U & S & U & S \\
\midrule
Volcano & Woo et al., Eulerian & $\le$1 & $\le$1 & 4.4 & 1.0 & 3.9 & $\le$1 \\
& Woo et al., Lagrangian & $\le$1 & $\le$1 & 2.5 & 1.1 & 1.8 & 1.4 \\
& Woo et al., semigroup & $\le$1 & $\le$1 & 4.3 & $\le$1 & 3.9 & $\le$1 \\
 & G-LSD & $\le$1 & $\le$1 &  &  & $\le$1 & $\le$1 \\
\midrule
Earthquake & Woo et al., Eulerian & $\le$1 & $\le$1 & 22.3 & 7.3 & 17.3 & 5.5 \\
& Woo et al., Lagrangian & $\le$1 & $\le$1 & 10.8 & 6.1 & 7.7 & 5.7 \\
& Woo et al., semigroup & $\le$1 & $\le$1 & 25.2 & 6.8 & 25.1 & 6.3 \\
 & G-LSD & $\le$1 & $\le$1 &  &  & 6.1 & 3.8 \\
\midrule
Flood & Woo et al., Eulerian & $\le$1 & $\le$1 & 10.2 & 5.3 & 9.0 & 4.3 \\
& Woo et al., Lagrangian & $\le$1 & 1.1 & 6.1 & 4.4 & 5.4 & 4.1 \\
& Woo et al., semigroup & $\le$1 & 1.0 & 11.8 & 5.0 & 11.6 & 3.9 \\
 & G-LSD & 1.1 & $\le$1 &  &  & 6.7 & 5.6 \\
\midrule
Fire & Woo et al., Eulerian & 1.9 & $\le$1 & 45.5 & 12.2 & 47.8 & 11.0 \\
& Woo et al., Lagrangian & 1.3 & $\le$1 & 29.6 & 11.7 & 19.0 & 11.1 \\
& Woo et al., semigroup & $\le$1 & $\le$1 & 70.7 & 12.7 & 70.7 & 12.2 \\
 & G-LSD & $\le$1 & $\le$1 &  &  & 14.0 & 7.0 \\
\midrule
General & Woo et al., Eulerian & 304 & 10.9 & 332 & 45.1 & 413 & 39.0 \\
& Woo et al., Lagrangian & 296 & 4.6 & 266 & 5.6 & 208 & 6.4 \\
& Woo et al., semigroup & 9.9 & 4.6 & 546 & 4.2 & 651 & 5.3 \\
 & G-LSD & 4.4 & $\le$1 &  &  & 25.9 & 1.0 \\
\midrule
Glycine & Woo et al., Eulerian & 4.2 & $\le$1 & 50.5 & 2.6 & 37.0 & 2.3 \\
& Woo et al., Lagrangian & 104 & $\le$1 & 42.0 & 3.0 & 14.7 & 2.3 \\
& Woo et al., semigroup & 2.7 & $\le$1 & 37.1 & 4.3 & 32.8 & 3.4 \\
 & G-LSD & $\le$1 & $\le$1 &  &  & 2.4 & 1.1 \\
\midrule
Proline & Woo et al., Eulerian & 44.8 & $\le$1 & 151 & 2.0 & 129 & 1.9 \\
& Woo et al., Lagrangian & 179 & $\le$1 & 54.6 & 1.8 & 24.0 & 1.5 \\
& Woo et al., semigroup & 2.7 & $\le$1 & 169 & 3.3 & 174 & 2.9 \\
 & G-LSD & $\le$1 & $\le$1 &  &  & 4.6 & $\le$1 \\
\midrule
Pre-Pro & Woo et al., Eulerian & 62.4 & $\le$1 & 24.8 & 1.8 & 17.8 & 1.4 \\
& Woo et al., Lagrangian & 49.3 & $\le$1 & 12.7 & 2.5 & 7.9 & 1.8 \\
& Woo et al., semigroup & 9.1 & $\le$1 & 17.7 & 1.6 & 16.3 & 1.1 \\
 & G-LSD & 5.8 & $\le$1 &  &  & 6.6 & $\le$1 \\
\midrule
RNA ($\mathbb T^7$) & Woo et al., Eulerian & 155 & $\le$1 & 126 & 15.0 & 227 & 11.8 \\
& Woo et al., Lagrangian & 203 & $\le$1 & 207 & 14.8 & 153 & 12.7 \\
& Woo et al., semigroup & 32.8 & $\le$1 & 222 & 8.1 & 599 & 7.9 \\
 & G-LSD & 16.8 & $\le$1 &  &  & 18.2 & 1.9 \\
\bottomrule
\end{tabular*}
\end{table*}

\noindent\textbf{All datasets.} \Cref{tab:zhong-full} gives the results of the Riemannian MeanFlow of \citet{zhong2026riemannian} on
every dataset, with all kernels, the prior alone and the memorization ratio. The RMF kernel separates the two priors on 15
of 21 datasets, always in favor of the statistical prior: the four $\SOthree$ benchmarks, seven of the eight synthetic
sphere datasets, and General, Proline, Pre-Pro and RNA (\cref{tab:rmf}). It does not separate them on the Earth data
and on Glycine, where finer kernels do. For every dataset, we report the checkpoint type that favors the uniform prior.
A geodesic kernel density estimate with validated bandwidth is competitive below four dimensions and falls behind from
four dimensions on.

\begin{table*}[p]
\footnotesize\setlength{\tabcolsep}{3pt}\renewcommand{\arraystretch}{0.9}
\centering
\caption{\textbf{Riemannian MeanFlow of \citet{zhong2026riemannian} on every dataset}, MMD$^2\times10^3$ (mean $\pm$ s.d.\ over seeds
and draws); floor: noise-floor s.d.\ of the kernels in column order; NN: memorization ratio (\cref{supp:evaldetails}). RMF est.: the RMF estimator of
\citet{zhong2026riemannian}. Bold: best in each column of a dataset; n/a: not applicable.}
\label{tab:zhong-full}\label{tab:fire}\label{tab:earth}\label{tab:torus}\label{tab:so3-rigorous}\label{tab:pcgrad}\label{tab:rmf-kernel}
\textbf{(a)} Earth data ($S^2$)\\[2pt]
\begin{tabular*}{\linewidth}{@{\extracolsep{\fill}}ll cccc}
\toprule
Data & Method & RMF kernel & $\gamma=10$ & $\gamma=50$ & NN \\
\midrule
Fire, 100 epochs & uniform prior, 1 step & $0.659\pm0.266$ & $9.517\pm0.659$ & $5.373\pm0.329$ & 18.32 \\
\emph{floor: 0.403 / 0.147 / 0.049} & uniform prior, 4 steps & $-0.177\pm0.131$ & $4.400\pm0.256$ & $5.187\pm0.714$ & 9.82 \\
 & vMF prior, 1 step & $-0.162\pm0.126$ & $0.419\pm0.051$ & $1.009\pm0.061$ & 4.54 \\
 & statistical prior alone & $-0.124\pm0.132$ & $1.962\pm0.185$ & $1.737\pm0.074$ & 9.3 \\
 & statistical prior, 1 step & $-0.221\pm0.105$ & $0.395\pm0.062$ & $0.983\pm0.070$ & 4.74 \\
 & statistical prior + hier.\ OT, 1 step & \best{$-0.304\pm0.061$} & \best{$0.190\pm0.038$} & \best{$0.621\pm0.060$} & \best{4.26} \\
\midrule
Fire, 400 epochs & uniform prior, 1 step & $0.527\pm0.091$ & $6.156\pm0.042$ & $4.382\pm0.036$ & 12.39 \\
 & uniform prior, 4 steps & $-0.246\pm0.053$ & $0.921\pm0.129$ & $1.529\pm0.032$ & 6.45 \\
 & vMF prior, 1 step & \best{$-0.271\pm0.092$} & \best{$0.133\pm0.022$} & \best{$0.431\pm0.041$} & \best{3.15} \\
 & statistical prior, 1 step & $-0.245\pm0.127$ & $0.145\pm0.080$ & $0.451\pm0.002$ & 3.44 \\
\midrule
Earthquake & uniform prior, 1 step & $-0.31$ & $4.45\pm0.09$ & $2.37\pm0.18$ & 4.46 \\
\emph{floor ($\gamma=10$/50): 0.234 / 0.100} & statistical prior alone &  & 1.09 & 0.86 & 3.08 \\
 & statistical prior, 1 step & \best{$-0.66$} & \best{$0.23\pm0.02$} & \best{$0.57\pm0.02$} & \best{1.78} \\
\midrule
Flood & uniform prior, 1 step & 1.77 & $2.67\pm0.39$ & $1.46\pm0.13$ & 4.38 \\
\emph{floor ($\gamma=10$/50): 0.372 / 0.099} & statistical prior alone &  & 1.30 & \best{0.80} & 3.22 \\
 & statistical prior, 1 step & \best{1.69} & \best{$0.70\pm0.26$} & $0.89\pm0.20$ & \best{2.23} \\
\midrule
Volcano & uniform prior, 1 step & $-1.17$ & $13.31\pm1.12$ & $11.83\pm0.15$ & 32.56 \\
\emph{floor ($\gamma=10$/50): 3.448 / 2.589} & statistical prior alone &  & $-1.31$ & 3.12 & 9.77 \\
 & statistical prior, 1 step & \best{$-2.78$} & \best{$-1.63\pm0.35$} & \best{$2.69\pm0.43$} & \best{5.94} \\
\bottomrule
\end{tabular*}\\[3pt]
\textbf{(b)} Torsion angles ($\mathbb T^2$; RNA: $\mathbb T^7$), one step; the RMF kernel is additive over the angles\\[2pt]
\begin{tabular*}{\linewidth}{@{\extracolsep{\fill}}ll ccccc}
\toprule
Data & Method & RMF kernel & joint $\gamma=1$ & joint $\gamma=3$ & joint $\gamma=10$ & NN \\
\midrule
Glycine & uniform prior & 0.594 & 1.670 & 2.738 & 2.176 & 1.78 \\
 & statistical prior alone & 1.568 & 1.369 & 1.409 & 0.678 & 1.65 \\
 & statistical prior & \best{$-0.239$} & \best{0.088} & \best{0.163} & \best{0.189} & \best{1.06} \\
 & \emph{floor} & \emph{0.458} & \emph{0.286} & \emph{0.150} & \emph{0.062} &  \\
\midrule
Proline & uniform prior & 5.320 & 13.53 & 21.07 & 15.85 & 5.20 \\
 & statistical prior alone & 0.989 & 3.880 & 5.857 & 2.831 & 1.73 \\
 & statistical prior & \best{$-0.224$} & \best{$-0.153$} & \best{0.201} & \best{0.532} & \best{1.02} \\
 & \emph{floor} & \emph{0.170} & \emph{0.234} & \emph{0.188} & \emph{0.158} &  \\
\midrule
Pre-Pro & uniform prior & 7.641 & 14.89 & 14.50 & 5.710 & 6.35 \\
 & statistical prior alone & 1.659 & 3.097 & 3.748 & 1.602 & 2.17 \\
 & statistical prior & \best{$-0.079$} & \best{0.344} & \best{0.962} & \best{0.764} & \best{1.13} \\
 & \emph{floor} & \emph{0.157} & \emph{0.184} & \emph{0.164} & \emph{0.105} &  \\
\midrule
General & uniform prior & 4.752 & 7.806 & 10.05 & 5.304 & 5.99 \\
 & statistical prior alone & 1.169 & 1.757 & 2.424 & 1.792 & 1.76 \\
 & statistical prior & \best{$-0.032$} & \best{0.017} & \best{0.093} & \best{0.155} & \best{1.04} \\
 & \emph{floor} & \emph{0.027} & \emph{0.029} & \emph{0.020} & \emph{0.012} &  \\
\midrule
RNA & uniform prior & 23.92 & 123 & 49.40 & 2.798 & 5.39 \\
 & statistical prior alone & 1.034 & 7.858 & 6.938 & 0.949 & 1.83 \\
 & statistical prior & \best{$-0.027$} & \best{0.457} & \best{0.601} & \best{0.170} & \best{1.24} \\
 & \emph{floor} & \emph{0.164} & \emph{0.124} & \emph{0.062} & \emph{0.010} &  \\
\bottomrule
\end{tabular*}\\[3pt]
\textbf{(c)} $\SOthree$ benchmarks, Laplace kernels (five draws of $2\times10^4$ samples); floors: range over the test splits of two seeds\\[2pt]
\begin{tabular*}{\linewidth}{@{\extracolsep{\fill}}ll ccccc}
\toprule
Data & Method & RMF est. & $\gamma=5$ & $\gamma=20$ & NN & RMF-MT, $\gamma=20$ \\
\midrule
Cone & uniform prior, 1 step & 0.051 & $4.01\pm0.33$ & $4.44\pm0.26$ & 16.3 & $4.40\pm0.09$ \\
\emph{floor: 0.059--0.066 / 0.023--0.034} & statistical prior alone &  & $-0.035\pm0.032$ & $0.044\pm0.020$ & 1.36 & n/a \\
 & statistical prior, 1 step & \best{0.012} & \best{$-0.031\pm0.030$} & \best{$-0.005\pm0.015$} & \best{1.03} & \best{$0.051\pm0.015$} \\
\midrule
Fisher & uniform prior, 1 step & 0.077 & $17.38\pm0.09$ & $3.544\pm0.004$ & 28.5 & $3.563\pm0.003$ \\
\emph{floor: 0.033--0.034 / 0.007--0.011} & statistical prior alone &  & \best{$0.91\pm0.04$} & \best{$0.57\pm0.02$} & 1.78 & n/a \\
 & statistical prior, 1 step & \best{0.025} & \best{$0.91\pm0.04$} & \best{$0.57\pm0.02$} & \best{1.77} & \best{$0.53\pm0.01$} \\
\midrule
Line & uniform prior, 1 step & 0.043 & $0.723\pm0.018$ & $0.084\pm0.003$ & 3.45 &  \\
\emph{floor: 0.009--0.011 / 0.001--0.002} & statistical prior alone &  & $0.237\pm0.037$ & $0.046\pm0.006$ & 2.31 & n/a \\
 & statistical prior, 1 step & \best{0.025} & \best{$0.222\pm0.024$} & \best{$0.044\pm0.005$} & \best{2.29} &  \\
\midrule
Swiss roll & uniform prior, 1 step & 0.073 & $45.9\pm3.4$ & $27.6\pm1.1$ & 558 & $29.7\pm0.1$ \\
\emph{floor: 0.085--0.095 / 0.045--0.054} & statistical prior alone &  & $0.109\pm0.055$ & $0.189\pm0.044$ & 10.9 & n/a \\
 & statistical prior, 1 step & \best{0.017} & \best{$0.081\pm0.043$} & \best{$0.173\pm0.027$} & \best{10.7} & \best{$0.172\pm0.016$} \\
\bottomrule
\end{tabular*}
\end{table*}

% \subsection{Higher-dimensional and conditional generation}\label{supp:res-q4}
\subsection{Higher Dimensions and Structured Data}\label{supp:res-q4}

\begin{figure*}[t]
\centering
\includegraphics[width=\textwidth]{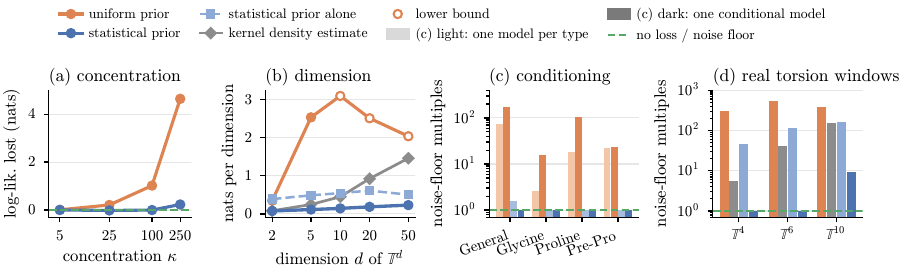}
\caption{\textbf{How the gap scales}, all panels. (a, b) As in \mref{fig:trends}. (c) One conditional model against one
model per residue type. (d) Real torsion windows of $k$ residues on $\mathbb T^{2k}$.}
\label{fig:trends-full}
\end{figure*}

\Cref{fig:trends-full} gives all four panels of \mref{fig:trends}, and \cref{tab:highdim20} the fragments of 20
residues.

\begin{table*}[t]
\vspace{-1em}
\centering\footnotesize\setlength{\tabcolsep}{2pt}
\caption{\textbf{Backbone fragments of 20 residues} ($\mathbb T^{40}$) rebuilt in 3D, as in \mref{tab:highdim}: clash
rate (lower is better), helix and sheet fractions (closer to held-out is better), and $W_1$ distance between end-to-end
distances (\AA; lower is better). Bold: closest to held-out fragments.}
\label{tab:highdim20}
\begin{tabular*}{0.75\textwidth}{@{\extracolsep{\fill}}lcccc@{}}
\toprule
 & clash & helix & sheet & $W_1$ \\
\midrule
Held-out fragments & 0.071 & 0.516 & 0.398 & -- \\
RMF + uniform prior & 0.405 & 0.392 & 0.233 & \best{0.40} \\
Kernel density estimate & 0.418 & 0.437 & 0.286 & 1.47 \\
RMF + mixture prior & 0.264 & 0.469 & 0.259 & 2.99 \\
RMF + hidden-Markov prior & \best{0.160} & \best{0.504} & \best{0.403} & 5.83 \\
\bottomrule
\end{tabular*}
\end{table*}

\noindent\textbf{Synthetic tori of increasing dimension.} To study how the methods scale with dimension, we use synthetic mixtures of eight
banana-shaped components on $\mathbb T^d$, $d\in\{2,5,10,20,50\}$, whose densities are known exactly and whose divergence from uniformity per dimension is roughly constant (\cref{supp:data}). The tangent-space mixture is misspecified for these data. We report the log-likelihood lost per dimension by $10^4$ one-step samples, with the same architecture
and training budget for every $d$. \Cref{tab:higherdim}a gives the values behind \mref{fig:trends}b. For $d\ge10$, most samples from the uniform prior fall
where the true density is below $e^{-20}$ times the uniform density, so its loss is only a lower bound. Throughout, the network reduces the loss of the prior by a factor of 2.2--5.

\noindent\textbf{Real data in higher dimensions.} To check that this behavior carries over to real data, we take
non-overlapping windows of $k$ consecutive residues, which give real data on $\mathbb T^{2k}$ (\cref{supp:data}). For $k=10$ and $k=20$, MMD is no longer informative, so we rebuild the fragments in 3D. Helical runs, at least four
consecutive residues in the $\alpha$ basin, are not in \mref{tab:highdim}: for 10 and 20 residues, held-out fragments have
0.463 and 0.679, the uniform prior 0.157 and 0.237, a kernel density estimate 0.405 and 0.576, and the hidden-Markov
prior 0.473 and 0.690. The uniform prior also has 5--14 times the clash rate of real fragments, and the hidden-Markov
prior two to four times. \Cref{tab:higherdim}b adds a residue-shuffled control and the priors without the network.

\begin{table*}[t]
\footnotesize\setlength{\tabcolsep}{3pt}
\centering
\caption{\textbf{Higher dimensions.} (a) Synthetic tori: log-likelihood lost per dimension (nats); in parentheses, samples
with density below $e^{-20}$ times uniform (lower bounds). (b) Fragments rebuilt in 3D: a residue-shuffled control and the
priors without the network; clashes, helical runs ($\ge$4 residues), helix, sheet, $W_1$ of end-to-end distances (\AA). n/a: not applicable.}
\label{tab:higherdim}\label{tab:dim}\label{tab:fragments}
\textbf{(a)} Synthetic tori $\mathbb T^d$\\[2pt]
\begin{tabular*}{\linewidth}{@{\extracolsep{\fill}}cccccccc}
\toprule
$d$ & KL/$d$ & RMF, uniform & KDE (no network) & Prior alone & RMF + stat.\ prior & + stat.\ prior, 4 steps & $K$ (BIC) \\
\midrule
2  & 0.71 & 0.349 (0\%)            & 0.084 (0\%)   & 0.389 & {$0.077\pm0.012$} & 0.024 & 12 \\
5  & 1.31 & 2.529 (17\%)           & 0.247 (0\%)   & 0.487 & {$0.115\pm0.031$} & $-0.016$ & 12 \\
10 & 1.50 & $\ge3.090$ (72\%)      & 0.450 (0\%)   & 0.546 & {$0.145\pm0.003$} & 0.037 & 8 \\
20 & 1.55 & $\ge2.506$ (94\%)      & 0.923 (1\%)   & 0.611 & {$0.186\pm0.017$} & 0.106 & 8 \\
50 & 1.63 & $\ge2.034$ (100\%)     & 1.459 (14\%)  & 0.504 & {$0.231\pm0.008$} & 0.164 & 8 \\
\bottomrule
\end{tabular*}\\[6pt]
\textbf{(b)} Backbone fragments of 10 and 20 residues (bold: closest to held-out data)\\[2pt]
\begin{tabular*}{\linewidth}{@{\extracolsep{\fill}}l ccccc ccccc}
\toprule
& \multicolumn{5}{c}{10 residues ($\mathbb T^{20}$)} & \multicolumn{5}{c}{20 residues ($\mathbb T^{40}$)} \\
\cmidrule(lr){2-6}\cmidrule(lr){7-11}
& clash & helical run & helix & sheet & $W_1$ & clash & helical run & helix & sheet & $W_1$ \\
\midrule
Held-out fragments & 0.012 & 0.463 & 0.496 & 0.415 & n/a & 0.071 & 0.679 & 0.516 & 0.398 & n/a \\
Residue-shuffled control & 0.074 & 0.252 & 0.504 & 0.409 & 1.90 & 0.312 & 0.486 & 0.503 & 0.410 & 0.48 \\
\midrule
Flat mixture alone & 0.064 & 0.410 & 0.466 & 0.244 & 0.59 & 0.270 & 0.555 & 0.430 & 0.203 & 2.32 \\
Hidden-Markov prior alone & 0.045 & 0.470 & 0.499 & 0.404 & 0.63 & 0.163 & 0.694 & 0.508 & 0.393 & 5.70 \\
\bottomrule
\end{tabular*}
\end{table*}

\noindent\textbf{Residue-type-conditioned generation.}\label{sec:cond}
The conditional model of \cref{fig:trends-full}c takes the residue type as a one-hot input, and its data splits match those of
the per-type models. At the finest scale (\cref{tab:cond}), the conditional model with the statistical prior is close to the per-type models
on Glycine, Proline and Pre-Pro and less accurate on General (35 against 10). \Cref{fig:cond} shows the densities. Pooled training visits General about three times less often than the per-type model does.

\begin{table*}[t]
\footnotesize\setlength{\tabcolsep}{3pt}
\centering
\caption{\textbf{Residue-type-conditioned generation} at the finest scale (noise-floor multiples; \cref{fig:trends-full}c shows the RMF
kernel): one conditional model for four residue types (300 epochs, two seeds) against the per-type models.}
\label{tab:cond}
\begin{tabular*}{\linewidth}{@{\extracolsep{\fill}}lll cc}
\toprule
& & & \multicolumn{2}{c}{finest scale} \\
\cmidrule(lr){4-5}
Residue & Model & Prior & 1 step & 4 steps  \\
\midrule
\multirow{4}{*}{General} & per-type models & uniform & 246 & 186  \\
& per-type models & statistical & \best{10} & \best{5.1}  \\
& conditional model & uniform & 880 & 351  \\
& conditional model & statistical & 35 & 8.1  \\
\midrule
\multirow{4}{*}{Glycine} & per-type models & uniform & 22 & 14  \\
& per-type models & statistical & \best{1.7} & 1.3  \\
& conditional model & uniform & 73 & 27  \\
& conditional model & statistical & 2.5 & \best{$\le$1}  \\
\midrule
\multirow{4}{*}{Proline} & per-type models & uniform & 118 & 121  \\
& per-type models & statistical & \best{4.1} & \best{$\le$1}  \\
& conditional model & uniform & 113 & 69  \\
& conditional model & statistical & 4.6 & 2.4  \\
\midrule
\multirow{4}{*}{Pre-Pro} & per-type models & uniform & 55 & 35  \\
& per-type models & statistical & \best{5.5} & \best{1.7}  \\
& conditional model & uniform & 97 & 33  \\
& conditional model & statistical & 8.0 & 2.1  \\
\bottomrule
\end{tabular*}
\end{table*}

% \subsection{Roles of the prior, coupling and generator}\label{supp:res-q5}

\subsection{Ablation and Sensitivity Analysis}\label{supp:res-q5}
\Cref{tab:stress} gives the stress tests and the cost summarized in the main text.

\begin{table*}[t]
\centering\footnotesize\setlength{\tabcolsep}{2pt}
\caption{\textbf{Stress tests and cost.} (a) Finest scale, uniform $\to$ statistical prior (noise-floor standard
deviations). (b) One-time cost of fitting the prior and time per training step on one NVIDIA L40S GPU; sampling cost is
unchanged.}
\label{tab:stress}
\begin{minipage}[t]{0.53\textwidth}\centering
\textbf{(a)} Stress tests\\[2pt]
\begin{tabular*}{\linewidth}{@{\extracolsep{\fill}}ll cc@{}}
\toprule
Condition & Data & unif.\,$\to$\,stat. & factor \\
\midrule
Fitted on 500 points & Pre-Pro & 54 $\to$ 12 & 4.5$\times$ \\
 & Fire & 110 $\to$ 23 & 4.8$\times$ \\
EM, three iterations & Pre-Pro & 54 $\to$ 5.8 & 9.3$\times$ \\
 & Fire & 110 $\to$ 45 & 2.4$\times$ \\
Misspecified shapes & rings, arcs & \cref{supp:res-q1} & 3.7--19$\times$ \\
Uniform: 5$\times$ training, 4 steps & Proline & 6.3 $\to$ 3.4 & 1.9$\times$ \\
\bottomrule
\end{tabular*}
\end{minipage}\hfill
\begin{minipage}[t]{0.43\textwidth}\centering
\textbf{(b)} Cost\\[2pt]
\begin{tabular*}{\linewidth}{@{\extracolsep{\fill}}l c cc@{}}
\toprule
& fitting & \multicolumn{2}{c}{time per training step (ms)} \\
\cmidrule(lr){3-4}
Data & once & uniform & ours \\
\midrule
Fire ($S^2$) & 49\,s & 202 & 200 (1.0$\times$) \\
Cone ($\SOthree$) & 168\,s & 14.7 & 15.8 (1.1$\times$) \\
Pre-Pro ($\mathbb T^2$) & 11\,s & 11.2 & 12.6 (1.1$\times$) \\
Fragments ($\mathbb T^{20}$, HMM) & 451\,s & 11.4 & 14.0 (1.2$\times$) \\
\bottomrule
\end{tabular*}
\end{minipage}
\end{table*}

\begin{figure*}[t]
\centering
\includegraphics[width=\textwidth]{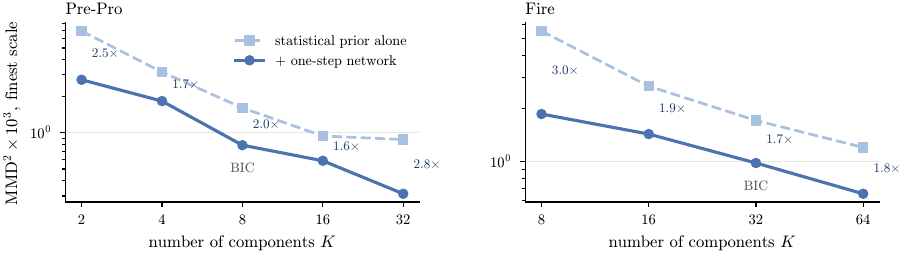}
% \vspace{-1.8em}
\caption{\textbf{Number of components} $K$ (MeanFlow, one step, finest scale): the fitted prior alone and one-step samples
trained with it; numbers: improvement factor. BIC: the selected $K$; other values from one seed.}
\label{fig:components}
\end{figure*}

\begin{figure*}[t]
\centering
\includegraphics[width=\textwidth]{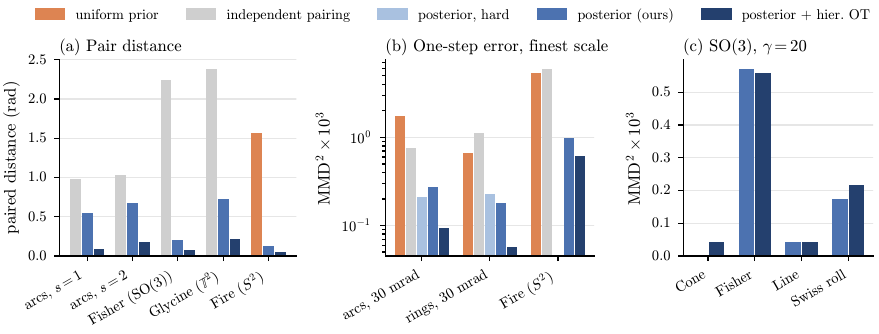}
\caption{\textbf{Couplings} with the statistical prior fixed ($s=2$ unless stated). (a) Mean geodesic distance between
paired noise and data (Fire, independent: uniform prior). (b) One-step error at the finest scale (arcs and rings: 30\,mrad;
no hard coupling on Fire). (c) $\SOthree$ at $\gamma=20$.}
\label{fig:coupling}
\end{figure*}

\noindent\textbf{Design choices.} The design choices matter less than the prior itself. Choosing $s$ and $K$ on validation data brings all four
$\SOthree$ benchmarks within or close to the noise floor, with errors of 0.010, 0.015 and 0.024 on Fisher, Line and Swiss
roll (\cref{tab:config}a), and more components help (\cref{fig:components}).  Drawing noise independently of the data point is 3--9 times less accurate than the posterior coupling (\cref{fig:coupling}b). Hierarchical OT shortens the paired distance by a further factor of 2.7--6.3 (\cref{fig:coupling}a) and lowers the error on Fire from 20 to 13 noise-floor standard deviations (\cref{fig:coupling}b). Under the posterior coupling, the noise follows the prior, as \mref{prop:marginal} states (\cref{tab:config}c). The main tables use the posterior coupling and so understate the full method.

\noindent\textbf{What the network adds.}
\label{sec:division}

\begin{figure*}[t]
\centering
\includegraphics[width=\textwidth]{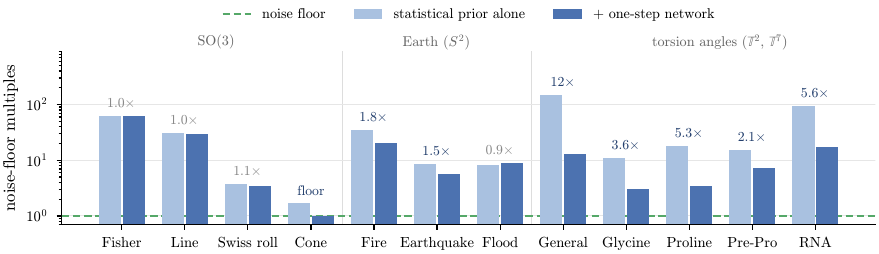}
\caption{\textbf{What the network adds}: distance from held-out data of the fitted prior alone ($s=2$) and of one-step
samples trained with it; numbers: improvement factor. Kernels: Laplace $\gamma=20$ ($\SOthree$) and $\gamma=10$ (Earth),
joint Laplace $\gamma=3$ (torus). On synthetic arcs and rings, the network lowers the log-likelihood lost from 0.87 to
0.14 and from 0.72 to 0.20 nats.}
\label{fig:network}
\end{figure*}

Since the fitted prior alone already performs well on several benchmarks, we ask what the network adds
(\cref{fig:network}). On data that are themselves mixtures of simple components, such as Fisher, Line and Swiss roll,
it changes the result by less than 10\%. On curved components, it corrects the shape. On synthetic arcs, the log-likelihood lost falls from 0.87 to 0.14 nats. At the finest scale, the distance falls by a factor of 2.1--12 on the real torus data and 1.5--1.8 on Fire and Earthquake. The prior decides \emph{where} the mass goes; the network decides \emph{its shape}.

\noindent\textbf{Prior complexity.} A more accurate prior shifts work from the network to the prior (\cref{fig:components}). As
$K$ grows, the prior alone approaches the data, and the network improves on it by a factor of 1.6--3.0 at every $K$.  The number of hidden-Markov states changes the fragment statistics little (\cref{tab:config}b).

\begin{table*}[t]
\footnotesize\setlength{\tabcolsep}{3pt}
\centering
\caption{\textbf{Configuring and checking the prior.} (a) Width $s$ and number of components $K$ chosen on validation
data ($K\le128$, one seed) against the default ($s=2$, $K\le32$). (b) Hidden-Markov states $S$, fragments of 10 residues
($^\ast$BIC). (c) \mref{prop:marginal}: TV of weights and mean responsibilities; MMD$^2$ of coupled noise to the prior. n/a: not applicable.}
\label{tab:config}\label{tab:sselect}\label{tab:hmmstates}\label{tab:couplingcheck}
\textbf{(a)} Width and number of components chosen on validation data\\[2pt]
\begin{tabular*}{\linewidth}{@{\extracolsep{\fill}}l c cc cc}
\toprule
& & \multicolumn{2}{c}{RMF kernel} & \multicolumn{2}{c}{finest scale} \\
\cmidrule(lr){3-4}\cmidrule(lr){5-6}
Dataset & selected & default & selected & default & selected \\
\midrule
Cone & $s=1$, $K=3$ & $-0.043$ ($\le$1) & $-0.037$ ($\le$1) & 0.003 ($\le$1) & 0.024 ($\le$1) \\
Fisher & $s=1$, $K=32$ & 0.147 (2.3) & 0.168 (2.6) & 0.571 (63) & \best{0.010 (1.1)} \\
Line & $s=1$, $K=96$ & 0.088 (1.9) & 0.056 (1.2) & 0.044 (29) & \best{0.015 (10)} \\
Swiss roll & $s=2$, $K=64$ & 0.050 ($\le$1) & 0.075 ($\le$1) & 0.173 (3.5) & \best{0.024 ($\le$1)} \\
Pre-Pro & $s=1$, $K=8$ & $-0.079$ ($\le$1) & $-0.083$ ($\le$1) & 0.683 (6.5) & \best{0.640 (6.1)} \\
Flood & $s=1$, $K=48$ & 1.69 (2.7) & 1.37 (2.2) & 0.890 (9.0) & \best{0.643 (6.5)} \\
\bottomrule
\end{tabular*}\\[6pt]
\textbf{(b)} Number of hidden-Markov states\\[2pt]
\begin{tabular*}{\linewidth}{@{\extracolsep{\fill}}c ccccc@{}}
\toprule
$S$ & clash & helical run & helix & sheet & $W_1$ \\
\midrule
8 & 0.044 & 0.461 & 0.499 & 0.396 & 0.44 \\
16 & 0.043 & 0.464 & 0.502 & 0.405 & 0.51 \\
32$^\ast$ & 0.043 & 0.473 & 0.502 & 0.407 & 0.58 \\
48 & 0.041 & 0.467 & 0.496 & 0.413 & 0.60 \\
\midrule
\emph{held-out} & \emph{0.012} & \emph{0.463} & \emph{0.496} & \emph{0.415} & n/a \\
\bottomrule
\end{tabular*}\\[6pt]
\textbf{(c)} Preservation of the prior (largest over seeds; MMD$^2$ in noise-floor s.d.)\\[2pt]
\begin{tabular*}{\linewidth}{@{\extracolsep{\fill}}l c c c}
\toprule
Model & $K$ & total variation & $|$MMD$^2|$ (noise-floor s.d.) \\
\midrule
Cone, Fisher, Line, Swiss roll & 3--32 & $\le0.030$ & $\le1.2$ \\
Volcano, Earthquake, Flood, Fire & 32 & $\le0.012$ & $\le1.4$ \\
General, Glycine, Proline, Pre-Pro, RNA & 8--24 & $\le0.006$ & $\le1.5$ \\
\midrule
Pre-Pro, EM stopped after three iterations & 8 & 0.006 & 0.85 \\
Pre-Pro, fitted on 500 points & 6 & 0.041 & 1.8 \\
Fire, EM stopped after three iterations & 24 & 0.065 & 2.0 \\
Fire, fitted on 500 points & 16 & 0.048 & 1.7 \\
\bottomrule
\end{tabular*}
\end{table*}

\section{Further Related Work}
\label{supp:related}

\noindent\textbf{Multi-step generative models on manifolds.} Diffusion and flow-matching models transport noise to data by
integrating a differential equation \citep{ho2020denoising,song2020score,lipman2022flow,liu2022flow,albergo2022building}.
Their Riemannian versions \citep{de2022riemannian,huang2022riemannian,lou2023scaling,chen2024flow} and normalizing flows on
manifolds \citep{mathieu2020riemannian,rezende2020normalizing,rozen2021moser,ben2022matching} model data on spheres, tori and
rotation groups accurately, but sampling typically takes tens to hundreds of network evaluations. Distillation
\citep{salimans2022progressive,yin2024one} and consistency models \citep{song2024improved,kim2024consistency,lu2025simplifying}
reduce this cost in Euclidean space.

\noindent\textbf{Statistical models of molecular structure.} Directional statistics models protein angles with mixtures of
distributions on the torus \citep{mardia2009directional,mardia2007protein}, and TorusDBN models backbone chains with a hidden
Markov model \citep{hamelryck2006sampling,boomsma2008generative}. Neural generators of protein structure
\citep{jing2022torsional,yim2023se,bose2024se,wu2024protein,watson2023novo} start from data-independent noise. We use these
classical statistical models as priors for neural one-step generators.

\noindent\textbf{Evaluating generated structures.} Structure validation checks backbone and side-chain angles against
high-quality experimental structures \citep{lovell2003structure,chen2010molprobity,shapovalov2011smoothed}. We combine
such checks with the MMD \citep{gretton2012kernel} under characteristic kernels \citep{sriperumbudur2011universality}, so
that samples are judged both as distributions and as structures.

\end{document}